\documentclass[sn-mathphys]{sn-jnl}

\usepackage[utf8]{inputenc}
\usepackage{mathtools}
\usepackage{amsthm}
\usepackage{bm}
\usepackage{booktabs}
\usepackage{graphbox}

\usepackage{csquotes}

\usepackage{dirtree}

\usepackage{setspace}

\DTsetlength{0.2em}{0.5em}{0.2em}{0.4pt}{1.6pt}

\newcommand{\argmax}{\mathop{\mathrm{argmax}}}
\newcommand{\softmax}{\mathop{\mathrm{softmax}}}
\newcommand{\expect}{\mathop{\mathbb{E}}}

\algnewcommand{\IfThenElse}[3]{
  \State \algorithmicif\ #1\ \algorithmicthen\ #2\ \algorithmicelse\ #3}

\algnewcommand{\IfThen}[2]{
  \State \algorithmicif\ #1\ \algorithmicthen\ #2}

\let\citet\cite

\usepackage{color}

\jyear{2021}%

\theoremstyle{thmstyleone}%
\theoremstyle{thmstyletwo}%

\theoremstyle{thmstylethree}%

\begin{document}

\title{Classification with Costly Features in Hierarchical Deep Sets}

\author*{\fnm{Jaromír} \sur{Janisch}}\email{jaromir.janisch@fel.cvut.cz}

\author{\fnm{Tomáš} \sur{Pevný}}\email{tomas.pevny@fel.cvut.cz}

\author{\fnm{Viliam} \sur{Lisý}}\email{viliam.lisy@fel.cvut.cz}

\affil{\orgdiv{Artificial Intelligence Center, Department of Computer Science}, \orgname{Faculty of Electrical Engineering, Czech Technical University in Prague}, \orgaddress{\country{Czech Republic}}}

\abstract{
Classification with Costly Features (CwCF) is a classification problem that includes the cost of features in the optimization criteria. Individually for each sample, its features are sequentially acquired to maximize accuracy while minimizing the acquired features' cost. However, existing approaches can only process data that can be expressed as vectors of fixed length. In real life, the data often possesses rich and complex structure, which can be more precisely described with formats such as XML or JSON. The data is hierarchical and often contains nested lists of objects. In this work, we extend an existing deep reinforcement learning-based algorithm with hierarchical deep sets and hierarchical softmax, so that it can directly process this data. The extended method has greater control over which features it can acquire and, in experiments with seven datasets, we show that this leads to superior performance. To showcase the real usage of the new method, we apply it to a real-life problem of classifying malicious web domains, using an online service.}


\maketitle

\section{Introduction} \label{sec:introduction}

The online world around us is composed of structured relational data. For example, users of a social network can be described by a set of their friends, posts they published or commented on, likes they received and from whom. This data is often not available as a whole, but rather provided on request by a paid service. Application Programming Interfaces (APIs) are specific examples. Google search, maps, Youtube, social networks such as Facebook or Twitter, and more provide rich information that may be free in low volumes but is charged as soon as you consider using it commercially. Even if the complete data is available, one can still save substantial resources by using only its fraction, e.g., when analyzing a large number of users.  Recently, we see that sustainability and ecology have started to play an increasingly larger role and the interest could lie in lowering electricity consumption or CO\textsubscript{2} production.

In the social network example, the use of the data may be targeted advertising. As another example, let us consider the field of computer security. One may be interested in whether a particular web domain is legitimate or malicious. Specialized services provide rich sets of features about the requested domain, such as known malware binaries communicating with the domain, WHOIS information, DNS resolutions, subdomains, associated email addresses, and, in some cases, a flag that the domain is known to be malicious. The user can further probe any detail, e.g., after acquiring a list of subdomains, the user can focus on one of them and request more information about it. Again, access to the service may be charged, therefore there is a natural pressure to limit the number of requests.

The problem at hand has multiple names -- \emph{classification with costly features} (CwCF) \citep{janisch2019sequential}, \emph{active feature acquisition and classification} \citep{shim2018joint} or \emph{datum-wise classification} \citep{dulac2012sequential}. In essence, the problem is to sequentially gather features, in a unique order for each sample, and stop optimally when ready to classify. Optimality is usually defined as one of the two: (1) a trade-off between the total cost of the features and the classification accuracy or (2) maximal accuracy with the condition that the total per-sample cost cannot exceed a specified budget. We emphasize that a potentially different feature subset acquired in different order is retrieved for each sample. For example, with some samples, the classification may be made after a single feature is acquired. Other samples may require multiple or all features, and the decision which is made sequentially, based on the values revealed so far. Note that the number of possible ways to process a sample is exponential in its size.

Over the years, many different algorithms have been developed for this problem. Some employ decision trees \citep{xu2012greedy,kusner2014feature,xu2013cost,xu2014classifier,nan2015feature,nan2016pruning,nan2017adaptive}, recurrent neural networks \citep{contardo2016recurrent}, linear programming \citep{wang2014lp,wang2014model} or partially observable Markov decision processes \citep{ji2007cost}. There are multiple reinforcement learning (RL) methods based on \citet{dulac2012sequential}, e.g., by \citet{janisch2019classification,janisch2019sequential,shim2018joint}. The problem itself, or its variations, appears across multiple fields: medicine \citep{peng2018refuel,lee2020interactive,song2018deep,vivar2020peri,lee2020co,shpakova2021probabilistic,zhu2020learning,goldstein2020target,erion2022cost}, meteorology \citep{banerjee2020deep}, data analysis \citep{ali2020reinforcement}, surveillance \citep{xu2021crowd,liu2018dependency} or network security \citep{badr2022enabling}.

\begin{figure}[t]
  \centering
  \includegraphics[width=1.0\linewidth]{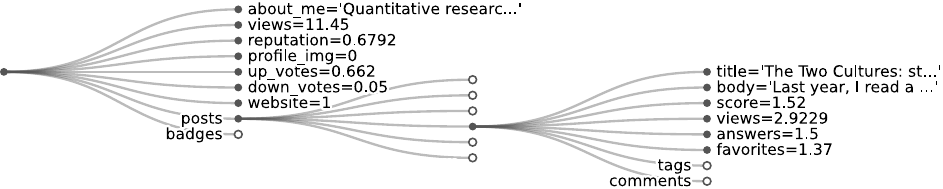}
  \caption{A pruned data sample from our \emph{stats} dataset, which is extracted from Stats StackExchange online service. The variable number of badges, posts, and their tags and comments means that each sample contains a different number of features. Application of existing techniques (e.g., original CwCF) would require alteration of the data. As a better alternative, we present a modified method that naturally works with the structured data and can select individual features in the hierarchy.}
  \label{fig:introsample}
\end{figure}

Despite the clear spread of the problem and its applications, we identified a substantial lack on the side of available algorithms. As we have shown in the introductory examples, a data sample is often provided in a complex structure, not a fixed-length vector. Formats such as XML or JSON, to which newly acquired information is sequentially added, are better suited. These formats commonly contain lists of elements with a priori undefined lengths and nested objects. For example, imagine a list of a user's posts (see Figure~\ref{fig:introsample}). However, the common requirement of the available algorithms, which we surveyed above, is a flat structure of the samples. In other words, it is assumed that the samples can be described as fixed-size vectors, with their slices allocated to predefined features. 

If we want to apply the existing algorithms to this structured data, we need to process the samples so they can be described as fixed-size vectors. However, as we show in this article, this approach leads to sub-optimal results. It is much better to provide a means for the algorithm to select individual features anywhere in the structure. Eventually, this is what we expect -- the algorithm that can request a few relevant features for one of the user's posts can be much more efficient than an algorithm that uses an aggregated version of all posts with pre-selected features. Note that a pre-selected feature ordering based on their importance is difficult because the number of features differs in every sample. E.g., in the social network example, it is difficult to statically determine the importance of a post's title, because each sample has a different number of posts.

In this article, we extend the original CwCF framework to naturally work with the structured data, which presents two main challenges. First, we need a way to process the data at the input. Deep Sets \citet{zaheer2017deep} is a technique to process variable-sized input and Hierarchical Multiple-Instance Learning (HMIL) is its extension for hierarchically nested data \citep{pevny2016discriminative}. It defines a special neural network architecture that accommodates to the specified data and creates its embedding. The second challenge lies in the fact that the original CwCF framework assumes a fixed number of features to select from and that the action space is \emph{static}. However, this assumption does not hold in our case -- the data contains lists of (possibly nested) objects, and only a part of the complete sample is visible at any moment. Since we map visible features with unknown values to actions, there is a different number of actions available to the algorithm at any moment. Moreover, there is no a priori known upper bound for the number of actions. Inspired by a technique from natural language processing \citep{morin2005hierarchical}, we take advantage of the hierarchical composition of the features and propose to decompose the policy analogically to their structure.

Finally, we demonstrate the extended CwCF framework with a set of experiments. First, we design a synthetic dataset which we use to analyze the algorithm's behavior. Second, we demonstrate the detection of malicious web domains with a real-world service. For this purpose, we created an offline dataset by collecting information about around 1~200 domains using the service's API. This dataset enables us to perform the experiments efficiently and credibly imitates real communication with the service. Third, we quantitatively test the methods in five more datasets adapted from public sources. 

Let us summarize the contributions of this manuscript:
\begin{enumerate}
  \item We formalize and bring the community's attention to a novel variant of an important problem (Section~\ref{sec:problem}).

  \item We extend the existing CwCF framework to work with structured data containing lists and nesting, which was not possible before. This includes processing the data on input and factorizing the dynamic action space to select individual features (Section~\ref{sec:method}). Other minor contributions include:
  \begin{itemize}
    \item We provide a formula to estimate the gradient of the policy entropy (required for the A2C algorithm) when only the probability of single action is known.
    \item We split the classifier and feature selection policy, leading to better sample complexity.
    \item We provide an unbiased loss for the classifier, weighted by the terminal action probability.
  \end{itemize}
   
  \item We evaluate our algorithm empirically and compare it to several alternatives, showing its superior performance. We execute the model with data from a real online service, proving its usefulness in a real-life scenario (Section~\ref{sec:experiments}).

  \item We release seven datasets in a unified format to benchmark algorithms for this problem (five datasets are adapted from existing public sources, and two are completely new). We also release the complete code with scripts to reproduce the experiments.
\end{enumerate}

This article is organized as follows. A detailed overview of the related work is presented in Section~\ref{sec:relatedwork}. Next, we describe the basic blocks we build upon in Section~\ref{sec:background}. Then we formalize the problem and formal changes to CwCF in Section~\ref{sec:problem}. Section~\ref{sec:method} focus on the algorithm and which practical changes are required. Experiments are presented in Section~\ref{sec:experiments}. Finally, Section~\ref{sec:discussion} provides answers to a few common questions and \ref{sec:conclusion} concludes the manuscript. Supplementary Material provides auxiliary information that did not fit the main text, such as dataset details, hyperparameters, visualizations, and training graphs.

\section{Related Work} \label{sec:relatedwork}
This work is a direct extension of the Classification with Costly Features (CwCF) framework, originally defined by \citet{dulac2012sequential} and lately advanced by \citet{janisch2019sequential,janisch2019classification}. All these algorithms are based on reinforcement learning (RL) but work only with fixed-length vectors. \citet{shim2018joint} proposes a method for sets of features, but cannot cope with nesting. We have covered some of the existing approaches \citep{xu2012greedy,kusner2014feature,xu2013cost,xu2014classifier,nan2015feature,nan2016pruning,nan2017adaptive,contardo2016recurrent,wang2014lp,wang2014model,ji2007cost,dulac2012sequential,janisch2019classification,janisch2019sequential,shim2018joint} and applications \citep{peng2018refuel,lee2020interactive,song2018deep,vivar2020peri,lee2020co,shpakova2021probabilistic,zhu2020learning,goldstein2020target,erion2022cost,banerjee2020deep,ali2020reinforcement,xu2021crowd,liu2018dependency,badr2022enabling} for the CwCF problem in Introduction.

Aside from the references mentioned above, multiple papers focus on a similar class of problems or improve the algorithms somehow. \citet{wang2015efficient} creates macro-features from different disjoint subsets of features. \citet{trapeznikov2013supervised} and \citet{liyanage2021dynamic} use a fixed order of features, while the latter provides an analytical solution to select them optimally. \citet{tan1993cost} analyzes a similar problem but requires memorization of all training examples. \citet{li2021active} uses RL with a generative surrogate model that provides intermediary rewards by assessing the information gain of newly acquired features and other side information.
\citet{bayer2005integrating} presents multiple approaches based on the AO* algorithm that searches the policy space, applicable in domains with discrete feature values. A case with a hard budget was explored in \citep{kapoor2005learning}. \citet{deng2007bandit} approached the problem with multi-armed bandit techniques. \citep{cesa2011efficient,zolghadr2013online} analyze the problem theoretically. \citet{kachuee2019opportunistic} uses heuristic reward to guide an RL-based algorithm.

A related problem is feature selection \citep{guyon2003introduction} which pre-selects a fixed set of features for all samples. However, in CwCF and similar approaches, the features are selected dynamically and sequentially. That is, for any particular sample, features are acquired one by one, and each decision is guided by the information gathered so far. This way, a different set of features is acquired for any particular sample. This approach requires more resources to train and execute but can provide higher performance (i.e., higher accuracy with the same average cost). Several approaches extend the feature selection to include costs of the features \citep{maldonado2017cost,bolon2014framework}. Still, they are designed to find a set of features common for the whole dataset and cannot work with structured data.


In this work, we use Hierarchical Multiple-Instance Learning (HMIL) to process the structured data \citep{pevny2017using,pevny2016discriminative,pevny2019approximation,mandlik2022jsongrinder}, which is an extension of Deep Sets \citet{zaheer2017deep}. In some deep RL problems, the action space is composed of orthogonal dimensions and existing techniques can be used to factorize it \citep{tang2020discretizing,chen2019effective,metz2017discrete}. In our case, the features are arranged in a tree-like structure and we factorize the corresponding action space with hierarchical softmax, a technique similar to the one used in natural language processing \citep{morin2005hierarchical,goodman2001classes}.

We optimize our model with the A2C algorithm derived from \citep{mnih2016asynchronous}, which belongs to a class of policy gradient RL algorithms \citep{sutton2018reinforcement}. It can be replaced with another algorithm from its class that works with discrete actions (e.g., TRPO \citep{schulman2015trust} or PPO \citep{schulman2017proximal}). While the use of the A2C algorithm is enough for the purposes of this paper, we note that any recent or future algorithm from the RL community may result in improved performance and better sample complexity.

The problem is distantly related to graph classification algorithms (e.g., \citet{zhou2018graph,hamilton2017inductive,perozzi2014deepwalk,kipf2016semi}). These algorithms either aim to classify graph nodes or the graph itself as a whole. In our case, we assume that the data is structured in a \emph{tree}, constructed around a point of interest (e.g., a particular web domain). For this kind of data, the HMIL algorithm is better suited and less expensive than the general message-passing. Moreover, the graph classification algorithms do not involve sequential feature acquisition, nor account for the costs of features.

\section{Preliminaries} \label{sec:background}
This section describes the methods we build upon in this work. Our method is based on the Classification with Costly Features (CwCF) \citep{janisch2019sequential,janisch2019classification} framework to set the objective and reformulate the problem as an MDP. However, structured data pose non-trivial challenges due to their variable input size and the variable number of actions. To create an embedding of the hierarchical input, we use an extension of Deep Sets \citep{zaheer2017deep} called Hierarchical Multiple-Instance Learning (HMIL) \citep{pevny2016discriminative,mandlik2022jsongrinder}. To select the performed actions, we use hierarchical softmax \citep{morin2005hierarchical,goodman2001classes}. To train our agent, we use Advantage Actor Critic (A2C) \citep{mnih2016asynchronous}, a reinforcement learning algorithm from the policy gradient family.

\subsection{Classification with Costly Features} \label{sec:cwcf}
Let us start by explaining the core concept of the Classification with Costly Features (CwCF) \citep{janisch2019classification,janisch2019sequential}. In CwCF, a data sample consists of \emph{features} (e.g., a user's name, reputation, etc.), each of which has a defined cost. Initially, the sample's feature values are unknown. The algorithm proceeds sequentially, and at each step, it decides whether to acquire another feature and which, or classify the sample. Note that the order of features is not fixed, but chosen dynamically. The objective is to optimally balance the total cost of features and classification accuracy, averaged over the dataset. Compared to feature selection \citep{guyon2003introduction}, this approach can achieve higher accuracy with the same cost because it can select a different set of features for each sample. The limitation of the framework is that it assumes that every sample contains exactly the same features and that they can be converted to a fixed-length vector. However, if the sample contains \textquote{a list of user's posts}, the original CwCF does not provide a way to process it. 

The following paragraph defines the problem formally. Let $\mathcal D$ be a dataset containing data points $(x, y)$, where $x$ is the sample and $y$ is its label. Let $\mathcal X$ be the input space and $\mathcal Y$ the set of all labels. We willingly do not define the $\mathcal X$ more precisely to allow a wider interpretation of what a feature value is (the CwCF framework defined it as $\mathcal X \subseteq \mathbf R^n$). Let $\mathcal F$ be the set of all possible features. Each feature has a predefined real-valued cost and the cost function $c: \mathcal \oldwp(\mathcal F) \rightarrow \mathbf R$ returns their sum, where the $\oldwp$ symbol denotes a power set. Let the tuple $(y_\theta, k_\theta)$ denote a model parametrized with $\theta$, where $y_\theta: \mathcal X \rightarrow \mathcal Y$ returns the label and $k_\theta: \mathcal X \rightarrow \oldwp(\mathcal F)$ returns the features used. The objective is:
\begin{equation}\label{eq:cwcf_problem}  
  \min_\theta \expect_{(x,y) \in \mathcal D} \big[ \ell_{rl}(y_\theta(x), y) + \lambda c(k_\theta(x)) \big]
\end{equation}
Here, $\ell_{rl}$ denotes a classification loss, commonly defined as binary (0 in case of mismatch, -1 otherwise). $\lambda \in \mathbf R$ is a trade-off factor between the accuracy and the cost. Minimizing this objective means minimizing the expected classification loss together with the $\lambda$-scaled per-sample cost.

Alternatively, CwCF provides \citep{janisch2019sequential} two other possible objectives. First, the algorithm can be modified to allow the user to specify directly a per-sample average budget $b \in \mathbf R$ and avoid $\lambda$. The objective then becomes:
\begin{equation}
  \min_\theta \expect_{(x,y) \in \mathcal D} \big[\ell_{rl}(y_\theta(x), y)\big], \; \text{s.t.} \; \expect_{(x,y) \in \mathcal D} \big[c(k_\theta(x))\big] \leq b
\end{equation}
Finally, it is possible to set a hard per-sample budget that cannot be exceeded for any sample. The objective is then:
\begin{equation}
  \min_\theta \expect_{(x,y) \in \mathcal D} \big[\ell_{rl}(y_\theta(x), y)\big], \; \text{s.t.} \; \forall x: c(k_\theta(x)) \leq b
\end{equation}

We chose to build our extensions with the objective in eq.~\eqref{eq:cwcf_problem}, as it corresponds to the vanilla algorithm, and the rest of the paper will mention only this one. If the application demands it, the other two objectives are also possible. We included them for completeness and reference. The interested reader can find more details about their implementation in \citep{janisch2019sequential}.

The way to solve eq.~\eqref{eq:cwcf_problem} is to construct a special Markov decision process (MDP), in which a single sample $(x, y)$ is analyzed per episode and the total episode reward $R$ is:
\[ R = -\big[ \ell_{rl}(y_\theta(x), y) + \lambda c(k_\theta(x)) \big] \]
Finding an optimal policy parametrized with $\theta$ equals to maximizing the expected reward, thus solving eq.~\eqref{eq:cwcf_problem}. The MDP is constructed as follows. In a particular episode with a sample $(x, y)$, the state space $\mathcal S$ consists of states $s = (x, y, \bar{\mathcal F})$, where $\bar{\mathcal F} \subseteq \mathcal F$ is the set of currently observed features. The agent only sees an observation $o(x, \mathcal{\bar F})$, which denotes only the parts of $x$ corresponding to features $\mathcal{\bar F}$. It also does not know the label $y$. Each episode starts with an initial state $s_0 = (x, y, \emptyset)$. The action space $\mathcal A$ corresponds to features and class labels, $\mathcal A = \mathcal A_f \cup \mathcal A_t$, where $\mathcal A_f = \mathcal F, \mathcal A_t = \mathcal Y$ ($t$ in $\mathcal A_t$ as \emph{terminal}). Typically, the already acquired features are removed from the selection, hence $\mathcal A_f(s) = \mathcal F \setminus \mathcal{\bar F}$. After performing an action selecting a feature, the reward is proportional to its negative cost, and the feature value is disclosed. After a classifying action, the episode terminates, and the reward is the negative loss of classification. Formally, the reward function $r: \mathcal S \times \mathcal A \rightarrow \mathbf R$ and transition function $t: \mathcal S \times \mathcal A \rightarrow \mathcal S$ are defined as follows:
\[ r(s, a) = 
          \begin{dcases*}
            -\lambda c(a)  & if $a \in \mathcal A_f$ \\
            -\ell_{rl}(a, y)    & if $a \in \mathcal A_t$ 
          \end{dcases*} \]
\[ t(s, a) = 
          \begin{dcases*}
            (x, y, \mathcal{\bar F} \cup a)   & if $a \in \mathcal A_f$ \\
            \mathcal{T}                       & if $a \in \mathcal A_t$
          \end{dcases*} \]
Here, $\mathcal T$ denotes the terminal state. When the episode terminates, the final action is a class prediction, and it is used as the model output $y_\theta$. Finally, the set of all acquired features is used as $k_\theta = \mathcal{\bar F}$.

The MDP defined above is solved with a deep reinforcement learning algorithm. The result is a policy $\pi_\theta$ that prescribes which actions to take in which states. In the original CwCF implementation, the RL algorithm was DQN \citep{mnih2015human} with several improvements (\citep{van2016deep,wang2016dueling,munos2016safe}). However, the method does not hinge on a particular algorithm, and another one can be easily used.

The eq.~\eqref{eq:cwcf_problem} poses a multi-criterial optimization problem that balances the classification accuracy in $\ell_{rl}$ and the cost of used features in $\lambda c$, for a fixed $\lambda$. The optimal behavior for $\lambda \to \infty$ is to refrain from acquiring any features and immediately classify with the most populous class, given the statistics of the training dataset. With the other extreme, $\lambda = 0$, a classifier that uses all features can be used to estimate a lower bound of the accuracy. Still, it is only a lower bound, since a different model may provide a better accuracy. For the points between, i.e., $\lambda \in (0, \infty)$, the issue is the same -- we can only find a lower bound (e.g., with baseline methods). Finally, note that eq.~\eqref{eq:cwcf_problem} focuses on the training set performance, but the ultimate goal is to find a model that generalizes to unseen data points.

\subsection{A2C Algorithm} \label{sec:a2c}
The method presented in this paper depends on hierarchical policy decomposition (explained in Section~\ref{sec:actionselection}), which is possible if the policy is probabilistic. However, the original CwCF uses the DQN algorithm that outputs a deterministic policy that cannot be easily factored. Therefore, we propose to use the Advantage Actor-Critic algorithm (A2C) \citep{mnih2016asynchronous}, a basic policy gradient algorithm to find the policy $\pi_\theta$. However, we note that any other algorithm from the policy gradient family with discrete actions (e.g., \citep{schulman2015trust,schulman2017proximal}) could be used in its place. This is an advantage of RL-based methods -- any recent or future improvement in deep RL algorithms can be immediately used with this method to improve its performance or sample complexity.

A detailed description of the A2C algorithm follows. An MDP is a tuple $(\mathcal{S}, \mathcal{A}, t, r, \gamma)$, where $\mathcal{S}$ represents the state space, $\mathcal{A}$ is a set of actions, $t(s, a)$ is a transition function returning a distribution of states after taking an action $a$ in a state $s$, $r(s, a, s') \in \mathbf R$ is a reward function that returns a reward for a transition from a state $s$ to $s'$ through an action $a$, and $\gamma \in (0, 1]$ is a discount factor. The A2C algorithm iteratively optimizes a policy $\pi_\theta: \mathcal S \rightarrow P(\mathcal A)$, where $P(\mathcal A)$ denotes a probability distribution over actions, and a value estimate $V_\theta: \mathcal S \rightarrow \mathbf R$ with model parameters $\theta$ to achieve the best cumulative reward in a given MDP. Let us define a state-action value function $Q(s, a) = \expect_{s' \sim t(s, a)} [r(s, a, s') + \gamma V_\theta(s')]$ and an advantage function $A(s, a) = Q(s, a) - V_\theta(s)$. Then, the policy gradient $\nabla_\theta J$ and the value function loss $L_V$ are:
\begin{equation}\label{eq:j}
\nabla_\theta J = \expect_{s, a \sim \pi_\theta, t} \Big[ A(s, a) \cdot \nabla_\theta \log \pi_\theta(a \mid  s) \Big]
\end{equation}
\begin{equation}\label{eq:lv}
L_V = \expect_{s, a, s' \sim \pi_\theta, t} \Big[ q(s, a, s') - V_\theta(s) \Big]^2
\end{equation}
\begin{equation}\label{eq:q}
 q(s, a, s') = r(s, a, s') + \gamma V_{\theta'}(s') 
\end{equation}
where $\theta'$ is a fixed copy of parameters $\theta$ and $\pi_\theta(a \mid s)$ denotes the probability of action $a$ under policy $\pi_\theta$ in state $s$.

To prevent premature convergence, a regularization term $L_H$ in the form of the average policy entropy is used:
\begin{equation}\label{eq:h}
L_H = \expect_{s \sim \pi_\theta, t} \Big[ H_{\pi_\theta}(s) \Big] \; ; \; H_{\pi}(s) = -\expect_{a \sim \pi(s)} \Big[ \log \pi(a\mid s) \Big] 
\end{equation}

The total loss is computed as $L_{pg} = -J + \alpha_v L_V - \alpha_h L_H$, with $\alpha_v, \alpha_h$ learning coefficients. The algorithm iteratively gathers sample runs according to a current policy $\pi_\theta$, and the traces are used as samples for the above expectations. Then, an arbitrary gradient descent method is used with the gradient $\nabla_\theta L_{pg}$. Often, multiple environments are run in parallel to get a better gradient estimate. Note that while \citet{mnih2016asynchronous} used asynchronous gradient updates, A2C performs the updates synchronously.

\subsection{Hierarchical Multiple-Instance Learning} \label{sec:hmil}
\begin{figure}[t]
  \centering
  \includegraphics[width=0.7\linewidth]{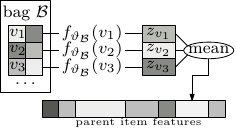}
  \caption{Illustration of the bag embedding in HMIL. Objects in the bag $\mathcal B$ are processed with $f_{\vartheta_\mathcal{B}}$ and aggregated. The result is used as the feature value for the parent object. The process recursively embeds the whole sample.}
  \label{fig:hmil}
\end{figure}

In our method, we need a way to process structured data. Our data samples are trees of features and they can contain nested lists of objects, similar to XML and JSON formats. To process this data on input, we use an extension of Deep Sets \citet{zaheer2017deep} for hierarchical data, called Hierarchical Multiple-Instance Learning (HMIL) \citep{pevny2016discriminative,mandlik2022jsongrinder}. For an illustration of how HMIL works, see Figure~\ref{fig:hmil}.

Let us start with MIL \citet{pevny2017using}, which presents a neural network architecture to learn an embedding of an unordered set (called a \emph{bag}) $\mathcal B$, composed of $m$ items $v_{\{1..m\}} \in \mathbf R^n$. The items are simultaneously processed into their embeddings $z_{v_i} = f_{\vartheta_\mathcal{B}}(v_i)$, where $f_{\vartheta_\mathcal{B}}$ is a non-linear function with parameters $\vartheta_\mathcal{B}$, shared for the bag $\mathcal B$. All embeddings are processed by an aggregation function $g$, commonly defined as an element-wise mean or max operator. The whole process creates a bag's embedding $z_\mathcal B = g_{i=1..m}( z_{v_i} )$, and is differentiable.

HMIL extends the framework so that it works with nested bags. In MIL, features are real scalars or vectors. In HMIL, a feature can also be a bag of items with the restriction that all the items share the same feature types. Different bags $\mathcal B$ have different parameters $\vartheta_{\mathcal{B}}$ and are recursively processed as in MIL, starting from the hierarchy's leaves and proceeding to the root. The resulting intermediary embeddings $z_{\mathcal B}$ are used as feature values (see Figure~\ref{fig:hmil}). The soundness of the hierarchical approach is theoretically studied by \citet{pevny2019approximation}.


\section{Problem} \label{sec:problem}
In this paper, we extend the CwCF framework (see Section~\ref{sec:cwcf}) to work with the structured data. This kind of data can be naturally processed with the HMIL architecture (Section~\ref{sec:hmil}). In this section, we describe what structured data means and how the problem formulation changes.

\subsection{Structured Data}
Compared to the data usually processed in machine learning, structured data, as we define it, cannot be described by fixed-length vectors. The main difference is that the samples can contain nested sets with a priori unknown cardinality. However, the structure of the samples is strictly defined. Below, we define the structured data with terms \emph{schema} and \emph{sample}.

\textbf{Dataset schema} recursively describes the structure, features, their types, and costs. Formally, let an \emph{object schema} be a collection of tuples \emph{(name, type, cost, children\_schema)}, where each tuple describes a single feature with its name, data-type, and non-negative real-valued cost. For features with \emph{type=set}, the \emph{children\_schema} is an object schema describing the objects in this set. For other features, \emph{children\_schema=$\emptyset$}. A dataset schema $\Sigma_{\mathcal D}$ is an object schema describing the whole sample.

\textbf{Data sample} is a collection of feature values, composed in a tree, and its structure strictly follows the schema $\Sigma_{\mathcal D}$. Formally, let an \emph{object} be a collection of feature values with types described by the corresponding object schema. We call each feature with \emph{type=set} a \emph{set feature}, and it is a collection of objects whose features are typed by the corresponding \emph{children\_schema}. Other features are called \emph{value features}.

\begin{figure}[t]
  \centering
  
  \begin{minipage}{0.5\linewidth}
    \fontsize{8}{8}
    \input{web_100k.txt}
  \end{minipage}
  \vspace{0.1cm}

  (a) schema of the \emph{threatcrowd} dataset

  \vspace{0.4cm}
  \includegraphics[width=1.0\linewidth]{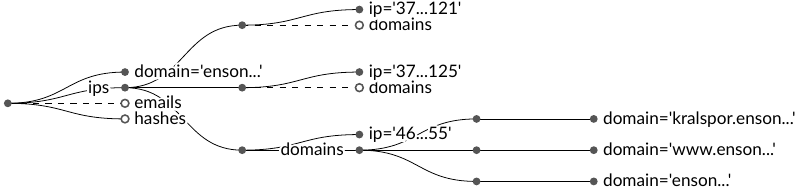}
  (b) a partial sample

  \caption{The schema and a partial sample for the \emph{threatcrowd} dataset. \textbf{(a)} The schema shows the feature names, their types, and their cost in parentheses. A \emph{set} type denotes that this feature contains a set of objects, whose features are described in the level below. \textbf{(b)} A partial sample. The full circles and lines denote features with known feature values. Among other information, the example shows that a list of domains was acquired for one of the IP addresses (\texttt{46..55}) with a reverse lookup.}
  \label{fig:schema_web}
\end{figure}

Both the schema and sample can be visualized as a tree. Figure~\ref{fig:schema_web}a shows an example of a schema \emph{threatcrowd} dataset. The schema specifies that each sample contains a free feature \emph{domain} with type \texttt{string} and sets of \emph{ips}, \emph{emails}, and \emph{hashes}. Objects in these sets have their own features (e.g., each IP address has a set of reversely translated \emph{domains}). Figure~\ref{fig:schema_web}b shows an incomplete sample as it would be seen by the augmented CwCF algorithm (only some of the features were acquired). Objects and their features are composed into a tree, according to the schema.

Note that our definition assumes that the cost of a particular feature across all samples is constant. While this assumption decreases the framework's flexibility, we argue that it is reasonable for real-world data where the cost of features can be usually precisely quantified upfront (e.g., the cost of an API request).

Last, it is useful to define a \emph{path} and \emph{prefix} of a feature in a particular sample. Let a path of a feature denote feature names and object positions in sets as a sequence from the root of the sample to the corresponding feature. We use the common programming syntax to denote the path. For example, we can write the path of features from the example in Figure~\ref{fig:schema_web}b as \textit{ips[0].ip} (the value of the first IP address), or \textit{ips[1].domains[0].domain} (the first domain of the second IP address). Let a prefix $pre(\kappa)$ of a feature $\kappa$ be its path without the last item. For example, $pre(ips[0].ip) = ips[0]$).

Note that while we address individual objects in a set by their index, we do this solely for the purposes of definitions and implementation. We assume that the order of objects does not have any predictive value.

\subsection{CwCF with Structured Data}
The original CwCF method (see Section~\ref{sec:cwcf}) worked with samples $x \in \mathbf R^n$. However, the data discussed in this paper cannot be easily converted to this Euclidean space. To accommodate for the issue, we present the following changes.

First, in CwCF, $\mathcal F$ denotes a set of all features. However, with structured data, the number of features is no longer constant across samples, as each sample can contain multiple objects in its sets. Therefore, let $\mathcal F(x)$ be a sample-dependent set of all features for a particular sample.

Second, a feature can be acquired only if its prefix has been obtained. For example, \emph{ips[0].ip} cannot be acquired before the set \emph{ips} or the object \emph{ips[0]} is obtained. Formally, we modify the available feature-selecting actions to $\mathcal A_f(s) = \{\kappa \in (\mathcal F(x) \setminus \mathcal{\bar F}) \mid pre(\kappa) \in \mathcal{\bar F} \}$. These actions correspond to features whose values are unknown, hence we call these features \emph{unobserved}.
As a minor optimization that facilitates training, we propose recursively processing the corresponding subtree and acquiring all features with zero cost, whenever a set feature is acquired.

Third, we decouple the classifier $y_\theta$ from the policy $\pi_\theta$. This change is not related to the structured data but results in improved performance and sample complexity. This is because the classifier can now be trained independently in every state and the policy is not burdened by the classification. Formally, we modify the set of terminal actions to include only a single terminal action $a_t$, $\mathcal A_t = \{ a_t \}$. The classifier $y_\theta$ is now separately trained on observations $o(x, \mathcal{\bar F})$ (remember that the observation discloses the parts of $x$ corresponding to features $\mathcal{\bar F}$). To simplify notation, let $\bar x = o(x, \mathcal{\bar F})$. The final prediction $y_\theta(\bar x)$ is used when the episode terminates. The reward function needs to reflect this change:
\[ r(s, a) = 
          \begin{dcases*}
            -\lambda c(a)  & if $a \in \mathcal A_f$ \\
            -\ell_{rl}(y_\theta(\bar x), y)    & if $a \in \mathcal A_t$ 
          \end{dcases*} \]
Note that we use parameters $\theta$ for both $\pi_\theta$ and $y_\theta$. Commonly, both of these functions are implemented as a neural network with shared layers and as such, their parameters overlap.

The original CwCF method solved a finite horizon MDP, since, for any dataset, there was a fixed number of features to acquire. To preserve this property in the modified framework, we need to add two assumptions. First, we assume that the dataset schema is finite, i.e., the feature hierarchy is limited in depth. The second assumption is that the number of objects in any set of any data sample is finite. These two assumptions together limit the number of features of any sample, therefore the modified method still operates within a finite horizon MDP.

Given these simple changes, the CwCF framework is \emph{formally} ready to work with structured data. However, the situation is more difficult implementation-wise, which is discussed in the following section.

\section{Method} \label{sec:method}
\begin{algorithm}[t]\footnotesize
\caption{HMIL-CwCF training}
\label{alg:training}
\begin{algorithmic}[1]
\State $\mathcal E$ - list of parallel environments, initialized as $(x,y,\mathcal{\bar F} = \emptyset), (x,y) \in \mathcal D$
\State $\Theta$ - model and its parameters (neural network)
\item[]
\Function{Train}{}
\While{not converged}
  \State batch $\mathfrak B = [\,]$
  \ForAll {$env \in \mathcal{E}$} \Comment{separate trace per environment}
    \State $s = (x, y, \mathcal{\bar F}), \bar x = o(x, \mathcal{\bar F})$; from $env.s$ \Comment{state and observation}
    \State $a, \pi(a \mid \bar x), \pi(a_t \mid \bar x), V, \varrho = \Theta(\bar x)$ \Comment{process with the model}

    \Comment{$a, a_t$ denote the selected and terminal actions}

    \State $r, s' = \Call{Step}{env, a, \argmax \varrho}$ \Comment{$\argmax \varrho$ needed only when $a = a_t$}
    \State append $s, a, r, s', \pi(a \mid \bar x), \pi(a_t \mid \bar x), V, \varrho$ to batch $\mathfrak B$
  \EndFor
  \State $L_{pg} = \Call{A2C}{\mathfrak B}$ \Comment{policy gradient loss}
  \State $L_{cls} = \expect_{\mathfrak B} \left [ \pi(a_t \mid \bar x) \cdot \ell_{cls}(\varrho(\bar x), y) \right ]$ \Comment{classifier loss (cross-entropy), eq.~\eqref{eq:classifier}}
  \State update $\Theta$ with $\nabla(L_{pg} + L_{cls})$ 
\EndWhile
\EndFunction
\item[]

\Function{Step}{environment $env$, action $a$, prediction $\hat y$}
    \State $(x, y, \mathcal{\bar F})$ = $env.s$ 
    \If{$a = a_t$}
      \State $r = -\ell_{rl}(\hat y, y)$ \Comment{RL loss (binary)} 
      \State sample new $(x', y')$ from $\mathcal D$, $\mathcal{\bar F} = \emptyset; env.s = (x', y', \mathcal{\bar F})$ \Comment{reset $env$}
      \State $s' = \mathcal T$
    \Else \Comment{$a \in \mathcal A_f \subseteq \mathcal F$}
      \State $r = -\lambda c(a)$ \Comment{cost of the feature}
      \State $\mathcal{\bar F} = \mathcal{\bar F} \cup a \cup \Call{GetFreeFeatures}{a}$
      \State $env.s = (x, y, \mathcal{\bar F}), s' = env.s$
    \EndIf
    \State \Return $r, s'$
\EndFunction

\item[]
\Function{A2C}{batch $\mathfrak B$} \Comment{with target clipping and sampled entropy}

  \State $\nabla J = \expect_{\mathfrak B} \Big[ A(\bar x, a) \cdot \nabla \log \pi(a \mid \bar x) \Big]$ \Comment{eq.~\eqref{eq:j}}

  \State $L_V = \expect_{\mathfrak B} \Big[ \text{clip}(r + \gamma V'(\bar x'), -\infty, 1.0) - V(\bar x) \Big]^2 $ \Comment{eqs.~\eqref{eq:lv},\eqref{eq:q2}} 
  
  \Comment{$V'$ is not updated; $V'(\bar x') = 0$ if $s' = \mathcal T$}

  \State $\nabla L_H = \expect_{\mathfrak B} \Big[ \log \pi(a \mid \bar x) \cdot \nabla \log \pi(a \mid \bar x) \Big] $ \Comment{eqs.~\eqref{eq:h},\eqref{eq:h2}}

  \State \Return $L_{pg} = -J + \alpha_v L_V - \alpha_h L_H$ \Comment{using auto-differentiation}
\EndFunction

\item[]
\Function{GetFreeFeatures}{$\kappa_0$} \Comment{recursively find free features}
  \State $\mathcal{\bar F} = \{ \}$
  \ForAll {$\kappa \mid pre(\kappa) = \kappa_0 \land c(\kappa) = 0 $} 
    \State $\mathcal{\bar F} = \mathcal{\bar F} \cup \kappa \cup \Call{GetFreeFeatures}{\kappa}$
  \EndFor
  \State \Return $\mathcal{\bar F}$
\EndFunction
\end{algorithmic}
\end{algorithm}

\begin{algorithm}[t]\footnotesize
\caption{HMIL-CwCF model}
\label{alg:model}
\begin{algorithmic}[1]
\Function{$\Theta$}{observation $\bar x$}
  \State $z_{\bar x} = \text{HMIL}(\bar x)$ \Comment{embed the observation}
  \State compute $V(z_{\bar x}); \nu_{a_t}(z_{\bar x}); \varrho(z_{\bar x})$ \Comment{separate heads in neural network}
  \State $a, \pi(a \mid \bar x), \pi(a_t \mid \bar x) = \Call{SelectAction}{\bar x, z_{\bar x}, \nu_{a_t}}$ \Comment{differentiable hierarchical softmax}
  \State \Return $a, \pi(a \mid \bar x), \pi(a_t \mid \bar x), V, \varrho$
\EndFunction

\item[]
\Function{HMIL}{bag $\mathcal B$}
  \ForAll {objects $v \in \mathcal B$}
    \ForAll {features $\kappa \in v \mid \text{type}(\kappa) = \text{set}$}
      \State $v_\kappa.\text{value} = \Call{HMIL}{v_\kappa.\text{items}}$ \Comment{recursively process set features}
      \State $v_\kappa.\text{mask} = \text{mean}_{v_i \in v_\kappa.\text{items}}(v_i.\text{mask})$ \Comment{\% of acquired features in sub-tree}
    \EndFor
    \State $v.\text{value} = \left[\forall \kappa \in v: v_\kappa.\text{value} \text{ if } \kappa \in \mathcal{\bar F} \text{ else } 0 \right] $ \Comment{concat the features' values}
    \State $v.\text{mask} = \left[\forall \kappa \in v: (1 \text{ or } v_\kappa.\text{mask}) \text{ if } \kappa \in \mathcal{\bar F} \text{ else } 0 \right] $ \Comment{use $v_\kappa.\text{mask}$ if $\text{type}(\kappa) = \text{set}$}
    \State $z_v = f_{\vartheta_{\mathcal B}}(v.\text{value}, v.\text{mask})$ \Comment{embed the object $v$}
  \EndFor
  \State \Return $\text{mean}_{v \in \mathcal B}(z_v)$ \Comment{average the vectors}
\EndFunction

\item[]
\Function{SelectAction}{bag $\mathcal B$, $z_{\bar x}$, $\nu_{a_t}$}
  \For{$i = 1..n$}
    \If{i = 1} \Comment{at first level, append $\nu_{a_t}$ to softmax}
      \State $\mathbb{P}(v, \kappa \text{ or } a_t \mid \bar x) = \softmax_{a_t,v,\kappa}\big(\nu_{a_t}, f_{\varphi_\mathcal{B}}(z_{\bar x}, z_v) : v \in \mathcal B \big)^\dagger$ \Comment{$z_v$ from \Call{HMIL}{}}
      \State sample $a_1 = (v, \kappa) \text{ or } a_t$ from $\mathbb P; \varpi_1 = \mathbb{P}(v, \kappa \text{ or } a_t \mid \bar x)$
      \State store $\pi(a_t \mid \bar x) = \mathbb{P}(a_t \mid \bar x)$
      \IfThen{$a_1 = a_t$}{break}
    \Else 
      \State $\mathbb{P}(v, \kappa \mid \bar x) = \softmax_{v,\kappa}\big(f_{\varphi_\mathcal{B}}(z_{\bar x}, z_v) : v \in \mathcal B \big)^\dagger$ 
      \State sample $a_i = (v, \kappa)$ from $\mathbb P; \varpi_i = \mathbb{P}(v, \kappa \mid \bar x) $
    \EndIf

    \If{$\text{type}(\kappa) = \text{set}$}
    \State $\mathcal B = v_\kappa.\text{items}$ \Comment{continue down the tree}
    \Else
    \State break \Comment{$v_\kappa$ is a leaf unobserved feature} 
    \EndIf
  \EndFor
  \State $a = [a_1, ..., a_n]; \pi(a \mid \bar x) = \prod_{i=1}^{n} \varpi_i$ \Comment{final action and its probability}
  \State \Return $a, \pi(a \mid \bar x), \pi(a_t \mid \bar x)$

  \item[]
  \State {$^\dagger$ to avoid choosing observed features, $f_{\varphi_\mathcal{B}}^{(\kappa)}(z_{\bar x}, z_v) = -\infty \text{ if } (v, \kappa) \in \mathcal{\bar F}$}
\EndFunction
\end{algorithmic}
\end{algorithm}

\begin{figure}[t]
  \centering
  \includegraphics[width=1.0\linewidth]{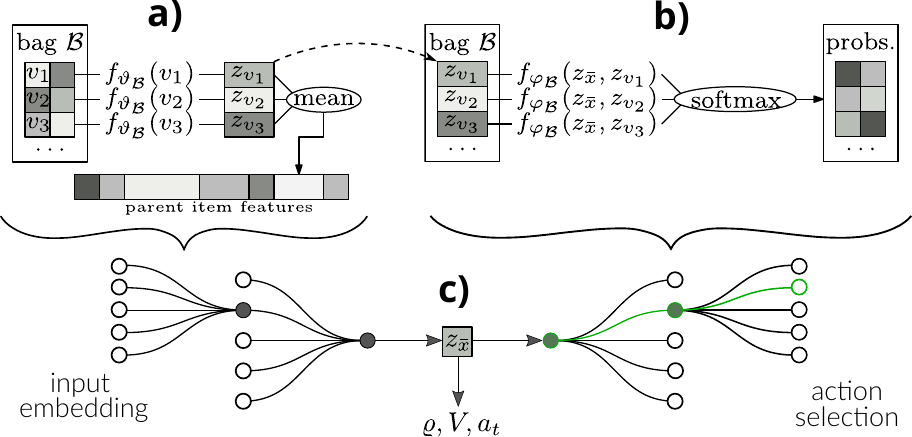}

  \caption{\textbf{(a)} The input $\bar x$ is recursively processed to create embeddings $z_v$ for each object $v$ in the tree and the sample-level embedding $z_{\bar x}$. \textbf{(c)} The embedding $z_{\bar x}$ is used to compute class probabilities $\varrho$, value estimate $V$, and the terminal action potential $a_t$. \textbf{(b)} An unobserved leaf feature is chosen with a sequence of stochastic decisions. Probabilities are determined by $f_{\varphi_\mathcal{B}}(z_{\bar x}, z_v)$. The whole architecture is end-to-end differentiable.}
  \label{fig:architecture}
\end{figure}

This section systematically introduces key details of our method to solve CwCF with structured data. The result is a model that is trained to solve eq.~\eqref{eq:cwcf_problem}. It is composed of several parts, as displayed in Figure~\ref{fig:architecture} and described by Algorithms~\ref{alg:training} and \ref{alg:model}. First, the input is processed with HMIL to create item-level and sample-level embeddings. Second, the sample-level embedding is used to create a class prediction, a value function estimate, and a terminal action value. Third, the action space is semantically factored with hierarchical softmax that creates a complete probability distribution over all actions.
Our model is a specialized end-to-end differentiable neural network, and we denote it with $\Theta$ and its parameters with $\theta$ (this includes parameters $\vartheta$ in HMIL, $\varphi$ in action selection and $\varrho, V, \pi$ output heads). To keep down the overall complexity of the final model, we minimize the number of layers used in each component. For example, we define the classifier $\varrho$ as a single neural network layer. However, this is not a limitation, since it uses the embeddings computed previously in the HMIL phase, and the whole network is updated end-to-end. When using our method, one may try to experiment with the number of layers to tune its performance for a concrete application. To declutter notation in the following text, we avoid using $\theta$ when describing gradients in $\nabla_\theta, \varrho_\theta, V_\theta$, and $\pi_\theta$. For reference, important symbols are summarized in Table~\ref{tab:symbols}.

\begin{table}[t]
    \centering
    \caption{Selected important symbols}

    \begin{tabular}{lp{0.75\linewidth}}
      \toprule
      symbol                        & description  \\
      \midrule                                
      $\lambda$                     & accuracy vs. cost trade-off factor \\
      $y_\theta, k_\theta$          & outputs of the model -- class and all acquired features \\
      $c$                           & cost function \\
      $x, y$                        & sample and class \\
      $\mathcal F(x)$               & all features of the sample $x$ \\ 
      $\mathcal{\bar F}$            & set of acquired features \\ 
      $\bar x = o(x, \mathcal{\bar F})$ & observation and observation function \\
      $\Sigma_{\mathcal D}$         & schema of a dataset $\mathcal D$ \\
      $a_t$                         & terminal action \\
      $\ell_{rl}$                   & classification loss for RL (binary) \\
      $\ell_{cls}$                  & classifier loss (cross-entropy) \\
      $\kappa$                      & a feature \\
      $pre(\kappa)$                   & prefix of the feature $\kappa$ \\
      \midrule                                
      $f_{\vartheta_{\mathcal B}}$     & HMIL embedding function for the bag $\mathcal B$ \\
      $f_{\varphi_{\mathcal B}}$       & pre-softmax embedding function for action selection for the bag $\mathcal B$ \\
      $z_v, z_{\bar x}$             & embeddings of an object $v$ and observation $\bar x$ \\
      $\pi$                         & action selection policy \\
      $\varrho$                     & classification probabilities \\
      $V$                           & value function \\
      $\nu_{a_t}$                   & pre-softmax value of terminal action $a_t$ \\
    \end{tabular}

    \label{tab:symbols}
\end{table}

\subsection{Input pre-processing}
The features in an observation $\bar x$ can be of different data types. Before processing with a neural network, they have to be converted into real vectors (only the features holding a value, not \emph{set} features). For strings, we observed good performance with character tri-gram histograms \citep{damashek1995gauging}. This hashing mechanism is simple, fast, and conserves similarities between strings. We used it for its simplicity and acknowledge that any other string processing mechanism is possible. One-hot encoding is used with categorical features.

For effectivity, the pre-processing step can take place before the training for the whole dataset. When the complete dataset is unavailable and the features are directly streamed upon request (e.g., during real-world inference), the values are converted on the fly.

During inference, the feature values can be unknown. In this case, a zero vector of the appropriate size is used. To help the model differentiate between observed and unobserved features, each feature in $x$ is augmented with a \emph{mask}. It is a single real value, either $1$ if the feature is observed or $0$ if not. In sets, the mask is the fraction of the corresponding branch that is observed, computed recursively.

\subsection{Input embedding} (Figure~\ref{fig:architecture}a, Algorithm~\ref{alg:model} \textsc{HMIL}) To process and embed the input, the first part of our fully differentiable model is HMIL (see Section~\ref{sec:hmil}). Its structure is determined by the dataset schema $\Sigma_{\mathcal D}$. Each set feature corresponds to a bag and the set of all such bags is $\{ \mathcal B_\kappa : \forall \kappa \in \Sigma_\mathcal{D} \; \mid \; \text{type}(\kappa) = \text{set}\}$. Before training, parameters $\vartheta_{\mathcal B_\kappa}$ are initialized for each bag $\mathcal B_\kappa$, which are later used for embedding items with the function $f_{\vartheta_{\mathcal B_\kappa}}$. We implement this function as one fully connected layer with LeakyReLU activation.

Let us clarify how HMIL is applied in our particular case to process an observation $\bar x$. The process starts with the leaves of the feature hierarchy and recursively proceeds toward the root. Each feature $\kappa$ with \emph{type=set} consists of a set of unordered objects $v$, collected in the bag $\mathcal B_\kappa$. All of these objects share the same type (enforced by the schema), i.e., they have the same features (however, not their values). The feature values of each object can be concatenated to $\mathbf R^n$, where $n$ is the size of the vector for the particular set $\kappa$. This is possible because the feature values are pre-processed, unknown features are replaced with zero vectors of the appropriate size, and the value of the set features is taken from the HMIL embedding of their contents. Each object $v \in \mathcal B_\kappa$ is processed by the embedding function $f_{\vartheta_{\mathcal B_\kappa}}(v) = z_v$, and the embeddings are saved to be used later. All items in the bag are mean-aggregated, and this value is used as the feature value of the parent object. Finally, when the whole tree is processed, the result is the root-level embedding $z_{\bar x}$.

\subsection{Classifier} (Figure~\ref{fig:architecture}c, Algorithm~\ref{alg:training} lines~8, 9, 13, 20)
The sample-level embedding $z_{\bar x}$ encodes the necessary information about the whole observation $\bar x$, and it is enough to compute the class probability distribution $\varrho(z_{\bar x})$ and the final decision $y_\theta(\bar x) = \argmax \varrho(z_{\bar x})$. We implement $\varrho$ as a single linear layer followed by softmax that converts the output to probabilities, and the classifier is trained parallelly to the policy $\pi$.

However, if we simply used every encountered state during training with the same weight, it would result in a biased classifier. 
This is because the classification is required only in terminal states and their reach probabilities need to be respected. Let $P_\pi(\bar x)$ denote a probability that the agent reaches $\bar x$ and terminates under policy $\pi$. The unbiased classification loss is then:
\begin{equation}\label{eq:classifier}  
  L_{cls} = \expect_{\bar x \sim P_\pi} \left[ \ell_{cls}(\varrho(\bar x), y) \right]
\end{equation}
To estimate the expectation in eq.~\eqref{eq:classifier}, we can either train the classifier only when the agent terminates, or we can use every encountered state weighted by the terminal action probability $\pi(a_t \mid \bar x)$. We use the latter because it provides an estimate with a lower variance. For $\ell_{cls}$, we use cross-entropy loss.

\subsection{Value function and terminal action} (Figure~\ref{fig:architecture}c, Algorithm~\ref{alg:model} line~3)
The embedding $z_{\bar x}$ is also used to compute the value function estimate $V(z_{\bar x})$ (required by the A2C algorithm) and pre-softmax value of the terminal action $\nu_{a_t}(z_{\bar x})$. Both functions are implemented as a single linear layer without any activation. The activation is not used in the value function, because its output should be unbounded, and it is commonly implemented in deep RL algorithms this way \citep{mnih2015human}. The output of $\nu_{a_t}(z_{\bar x})$ is converted to probability during the action selection.

\subsection{Action selection} \label{sec:actionselection}
\begin{figure}[t]
  \centering
  \includegraphics[width=1.0\linewidth]{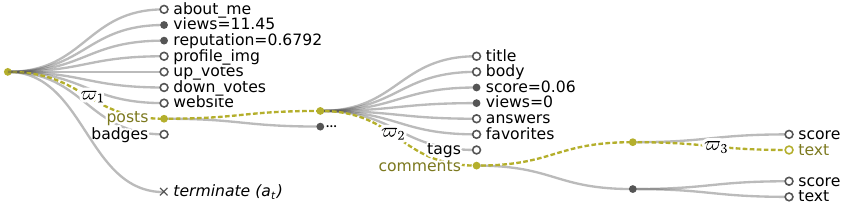}

  \caption{Visualization of how an action is selected. Sequentially, a path is created from the root to a leaf unobserved feature (or the terminal action) by a series of stochastic decisions. In set features, all items and their features are resolved at once. The probability of the performed action is a product of the partial probabilities on the path. In this example, the chosen action $a$ selects the \emph{posts[0].comments[0].text} feature with probability $\pi(a \mid \bar x) = \prod_{i=1}^3\varpi_i$.}
  \label{fig:hsoftmax_example}
\end{figure}

(Figure~\ref{fig:architecture}b and \ref{fig:hsoftmax_example}, Algorithm~\ref{alg:model} \textsc{SelectAction}) 
Let us describe the process of selecting an action. Remember that the observation $\bar x$ can be viewed as a tree, where value features are leaves and set features branch further. Note that this hierarchy is semantical, i.e., each set feature groups similar objects related to their parent. Therefore, it makes sense to use this semantical hierarchy for feature selection. We call the method below \emph{hierarchical softmax} and note that a similar technique was used in natural language processing \citep{morin2005hierarchical,goodman2001classes}.

For visualization, see Figures~\ref{fig:architecture}b and~\ref{fig:hsoftmax_example}. Oppositely to the input embedding procedure, the action selection starts at the root of $\bar x$ and a series of stochastic decisions are made at each node, continuing down the tree. The root node is regarded as a set with a single object. For each bag $\mathcal B \in \mathfrak B$, let the probability of selecting a feature $\kappa$ of an object $v$ be:
\begin{equation}\label{eq:hsoftmax}
  \mathbb{P}(v, \kappa \mid \bar x) = \softmax_{v,\kappa}\big(f_{\varphi_\mathcal{B}}(z_{\bar x}, z_v) : v \in \mathcal B\big)
\end{equation}
Here, $f_{\varphi_\mathcal{B}} : \mathbf R^n \rightarrow \mathbf R^m$ is a function that transforms the embeddings $z_{\bar x}$ and $z_v$ into a vector $\mathbf R^m$, where $n = \vert z_{\bar x} \vert + \vert z_{v} \vert$ and $m$ is the number of features for the object $v$. The bag-specific parameters $\varphi_\mathcal{B}$ are initialized prior training with the knowledge of the dataset schema for every possible bag $\mathcal B \in \mathfrak B$. In plain words, eq.~\eqref{eq:hsoftmax} means that all items in the bag $\mathcal B$ are processed with $f_{\varphi_\mathcal{B}}$, the outputs are concatenated are passed through the softmax function. This results in a single probability value for each feature in every object of $\mathcal B$, which are resolved at once.

Note that the function $f_{\varphi_\mathcal{B}}$ is a different function from $f_{\vartheta_{\mathcal B}}$. Its parameters are bag-specific, and it is implemented as a single fully connected layer with no activation function, since the output is later passed through softmax. Observed features and parts of the tree that are fully expanded (the mask of the corresponding features is $1$) are excluded from the softmax. We enforce this by setting the corresponding outputs of $f_{\varphi_\mathcal{B}}$ to $-\infty$, so the softmax returns $0$. At the root level, the terminal action potential $\nu_{a_t}(\bar x)$ is added to the softmax.

Now, remember that the action selection starts at the root of $\bar x$, iteratively samples from $\mathbb{P}(v, \kappa \mid \bar x)$ and proceeds down the tree, until it reaches a leaf feature (also, see Algorithm~\ref{alg:model} \textsc{SelectAction}). Let us define an action $a = [a_1, ..., a_n]$ as a list of the specific choices, $a_1 = (v_1, \kappa_1) \text{ or } a_t, a_2 = (v_2, \kappa_2), ..., a_n = (v_n, \kappa_n)$, where $n$ is the length of the path. 
We can write the probability of selecting the action $a$, given the observation $\bar x$, as a product of choice probabilities made on its path:
\begin{equation}\label{eq:pi}
  \pi(a \mid \bar x) = \prod_{i=1}^{n} \mathbb P(a_i \mid \bar x)
\end{equation}
Hence, any action $a \in \mathcal A_f \cup a_t$ (i.e., any currently unobserved leaf feature, or the terminal action) can be sequentially sampled from eq.~\eqref{eq:pi}.

The $\pi$ is a probability distribution of actions, hence it is a \emph{policy}. The decomposition according to eq.~\eqref{eq:pi} has several benefits. First, it was shown that a sensible policy decomposition introduces inductive biases to the model and speeds up the learning \citep{tang2020discretizing}. Our decomposition is logical because the decision on each level is made for objects that are semantically related. Second, it is interpretable, because it reveals which objects and features contributed to the decision. Third, it saves computational resources as only the probabilities on the selected path need to be computed. A drawback of the hierarchical softmax is that the decisions are made sequentially for each sample, which limits the parallel computation capabilities of modern GPUs. In our implementation, most of the time is spent on simulating the environment, and hence this drawback is negligible.

\subsection{Training} (Algorithm~\ref{alg:training} \textsc{Train} and \textsc{A2C})
We use the A2C algorithm (see Section~\ref{sec:a2c}) to optimize the policy $\pi$ with its parameters $\theta$, with the following changes. Note that we cannot train the model with value-based methods that were used with the original CwCF (e.g., DQN \citep{mnih2015human}), because they cannot optimize the policy itself.

First, we use the fact that the maximal $Q$ value is $1.0$ (the reward for correct prediction is $1.0$ and every other step has a negative reward) and clip the target $q$ in eq.~\eqref{eq:q} into $(-\infty, 1.0)$:
\begin{equation}\label{eq:q2}
  q(s, a, s') = \text{clip}(r(s, a, s') + \gamma V_{\theta'}(s'), -\infty, 1.0) 
\end{equation}
This reduces a maximization bias that occurs when learning a value function with neural networks \citep{van2016deep}.

Second, the computation of the policy entropy $L_H$ in eq.~\eqref{eq:h} requires knowledge of all action probabilities. However, the sequential nature of the hierarchical softmax means that only the $\pi(a \mid \bar x)$ for the actually performed action $a$ is computed. As the computation and gathering of probabilities for all actions are troublesome and unnecessary, we propose to estimate the entropy as follows. In the A2C algorithm, only the gradient $\nabla L_H$ is needed, and basic algebra shows that the correct way to estimate it is \citep{zhang2018efficient}:
\begin{equation}\label{eq:h2}
\nabla_\theta H_{\pi_\theta}(s) = -\expect_{a \sim \pi_\theta(s)} \Big [ \log \pi_\theta(a\mid s) \cdot \nabla_\theta \log \pi_\theta(a\mid s) \Big]
\end{equation}
Here, we use only the performed action to sample the expectation with zero bias, and the variance is decreased through large batches. For completeness, the derivation of eq.~\eqref{eq:h2} is in the Supplementary Material~\ref{apx:a2c}.

The A2C algorithm returns the loss $L_{pg}$ at each step. Simultaneously, the classification loss $L_{cls}$ is computed. Multiple parallel samples are processed at once to create a larger batch (see Supplementary Material~\ref{apx:implementation}  for further details). After each step, the model's parameters are updated in the direction of $-\nabla(L_{pg} + L_{cls})$. We believe that the A2C algorithm sufficiently demonstrates the method but note that any recent or future RL enhancement is likely to improve its performance. 

\subsection{Pretraining classifier}
The RL part of the algorithm optimizes eq. \eqref{eq:cwcf_problem}, which assumes a trained classifier. However, the classifier is trained simultaneously by minimizing eq. \eqref{eq:classifier}. As the classifier output appears in \eqref{eq:cwcf_problem} and eq. \eqref{eq:classifier} is based on the probability $P_\pi$, this introduces nonstationarity in both problems. To mitigate the issue and speed up convergence, we pretrain the classifier $\varrho$ with random observations (pruned samples). We cannot target a specific budget, since it is unknown before the training (only a tradeoff parameter $\lambda$ is specified). Hence, we cover the whole state space by generating observations $\bar x$ ranging from almost empty to complete. The exact details are in Supplementary Material~\ref{apx:implementation}.

\section{Experiments}\label{sec:experiments}
In this section, we describe several experiments that show the behavior of our algorithm and other tested methods. First, we describe the tested algorithms and the experiment setup. Then, we continue with a synthetic dataset designed to demonstrate the differences in algorithms' behaviors. Next, we apply the algorithm to a real-world problem of identifying malicious web domains. Finally, we gathered five more datasets for a quantitative evaluation. The complete code for all described algorithms and all datasets is shared publicly at \url{https://github.com/jaromiru/rcwcf}. For the reproducibility of our results, we also include the scripts to run the experiments and produce the plots.

\subsection{Tested Algorithms}
To our knowledge, there is no other method dealing specifically with costly hierarchical data. We constructed the following algorithms for comparison. Each of them represents certain class of algorithms and they can also be perceived as ablations of the main algorithm presented in this manuscript.

\textbf{HMIL} represents algorithms that disregard the costs and always use all available features. Alternatively, it can be seen as an ablation of the main algorithm, where we leave only the input embedding and classification parts. This method uses the complete information available, processes it directly with the HMIL algorithm and is trained in a supervised manner. This approach provides an estimate of achieveable accuracy, but also with the highest cost. In practice, using all features at once makes the algorithm prone to overfitting, which we mitigated by using aggressive weight decay regularization \citep{loshchilov2018decoupled}.

\textbf{RandFeats} represents a naive approach to the hierarchical composition of features, which are now selected randomly. With this, we can estimate the influence of the informed feature selection. It is an ablation of the full algorithm, implemented by replacing the policy with a random sampling. The algorithm acquires features randomly until a specified budget is exceeded. All other parts of the algorithm are kept the same. Since this algorithm is uninformed, we expect it to underperform the complete algorithm and give a lower bound estimate for accuracy.

\textbf{Flat-CwCF}: In this case, we demonstrate the original CwCF algorithm, which requires a fixed number of features. We achieve this by flattening the data -- only the root-level features are selectable, and the algorithm observes the complete sub-tree (embedded with HMIL) whenever such a feature is selected.  This algorithm behaves the same as the full algorithm on the root level but lacks fine control over which features it requests deeper in the structure. Because of that, we expect the method to underperform the full algorithm with lower budgets, but to reach the performance of \emph{HMIL} gradually. 

One could argue that we could also engineer a fixed set of features for each dataset and apply the original CwCF or a similar algorithm. For example, the engineered features for the \emph{threatcrowd} dataset (see Figure~\ref{fig:intro_example}-right for its schema) could include its domain and aggregated hashes of five random IP addresses, emails, and malware hashes. However, there can be more or fewer of these objects in the actual data sample. Given the variability of individual samples, the automatic selection of a static set of features is difficult, and the standard approaches to feature selection do not work with structured data.

In the original CwCF paper \citep{janisch2019sequential}, the authors proposed a heuristic baseline method that acquired features in a precomputed order sorted by their importance. For each subset, a specific classifier was trained to estimate the accuracy at this point, resulting in a point in the accuracy vs cost plane. The original CwCF method was shown to outperform this baseline, due to its ability to select per-sample specific features in a unique order. In our case, it is unclear how to apply this baseline to the hierarchical data where each sample has a different number of objects in its sets and a different number of features overall.

Finally, we refer to the full method described in this paper as \textbf{HMIL-CwCF}.
We searched for the optimal set of hyperparameters for each algorithm and dataset using validation data, and the complete table with all settings is in Supplementary Material~\ref{apx:implementation}.

\subsection{Experiment Setup}\label{sec:setup}
For each dataset, we ran \emph{HMIL} with ten different seeds, \emph{RandFeats} with 30 different budgets linearly covering either $[0, 10]$, $[0, 20]$ or $[0, 40]$ range (depending on the dataset) and \emph{Flat-CwCF} and \emph{HMIL-CwCF} with 30 different values of $\lambda$, logarithmically spaced in $[10^{-4}, 1.0]$ range. For each run, we selected the best epoch based on the validation data (for more details, see convergence graphs in Supplementary Material~\ref{apx:convgraphs}).

To visualize the results, we select the best runs that are on the Pareto front of the validation dataset, using the cost and accuracy criteria. We plot the best runs as a scatter plot with the average cost on the \emph{x}-axis and accuracy on the \emph{y}-axis and also visualize \emph{their} Pareto front with the testing set. To estimate variance, all other runs are visualized with faint color. For better comparison, we show the mean performance (± one standard deviation) of \emph{HMIL} across the whole \emph{x}-axis.

Apart from the graph form, the results are also reported as normalized Area Under the Trade-off Curve (AUTC). The AUTC metric describes the overall performance across the whole range of budgets. It is computed as the area under the visualized Pareto front, normalized by the total area of the graph, and the area below the prior of the most populous class is subtracted. The AUTC would return 0 for an algorithm that always predicts the most populous class and 1 for an algorithm with perfect classification. See Supplementary Material~\ref{apx:auc} for more details.

\subsection{Experiment A: Synthetic Dataset}
This experiment is aimed to demonstrate the behavior of our and other tested algorithms on purposefully crafted data. Note that this synthetic dataset is \emph{designed} to demonstrate the differences between the algorithms and therefore our method (\emph{HMIL-CwCF}) performs the best.  

Let us first explain the dataset's structure (follow its schema in Figure~\ref{fig:toy_schema}). A sample contains two sets (\emph{set\_a} and \emph{set\_b}), each with ten items. Each item has two features -- free feature \emph{item\_key} with a value \emph{0} and \emph{item\_value} containing a random label. Randomly, a single item in one of the sets is chosen, and its \emph{item\_key} is changed to \emph{1} and its \emph{item\_value} to the correct sample label. Further, the feature \emph{which\_set} contains the information about which set contains the indicative item. The idea is that the algorithm can learn a correct label by retrieving the \emph{which\_set} feature, opening the correct set, and retrieving the value for the item with \emph{item\_key=1}. Uniquely for this dataset, we test the algorithms directly on the training data.

\begin{figure}[t]
  \centering
  \begin{minipage}{0.5\linewidth}
    \fontsize{8}{8}
    \input{toy_b.txt}
  \end{minipage}
  \caption{The schema of the \emph{synthetic} dataset. The numbers in parentheses denote the costs of the corresponding features.}
  \label{fig:toy_schema}
\end{figure}

\begin{figure}[t]
  \begin{minipage}{0.5\linewidth}
  1: \includegraphics[align=c,scale=0.55]{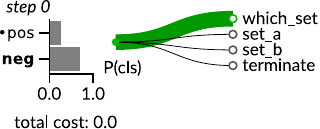}\\
  2: \includegraphics[align=c,scale=0.55]{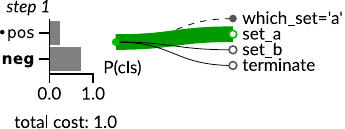}\\
  3: \includegraphics[align=c,scale=0.55]{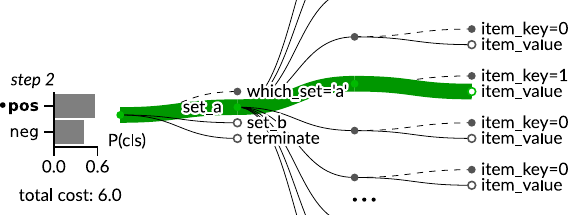}\\
  4: \includegraphics[align=c,scale=0.55]{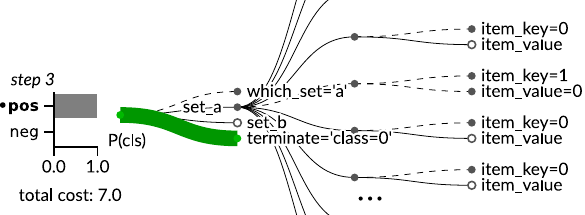}
  \end{minipage}%
  \begin{minipage}{0.5\linewidth}
    \centering
    \includegraphics[width=1.0\linewidth]{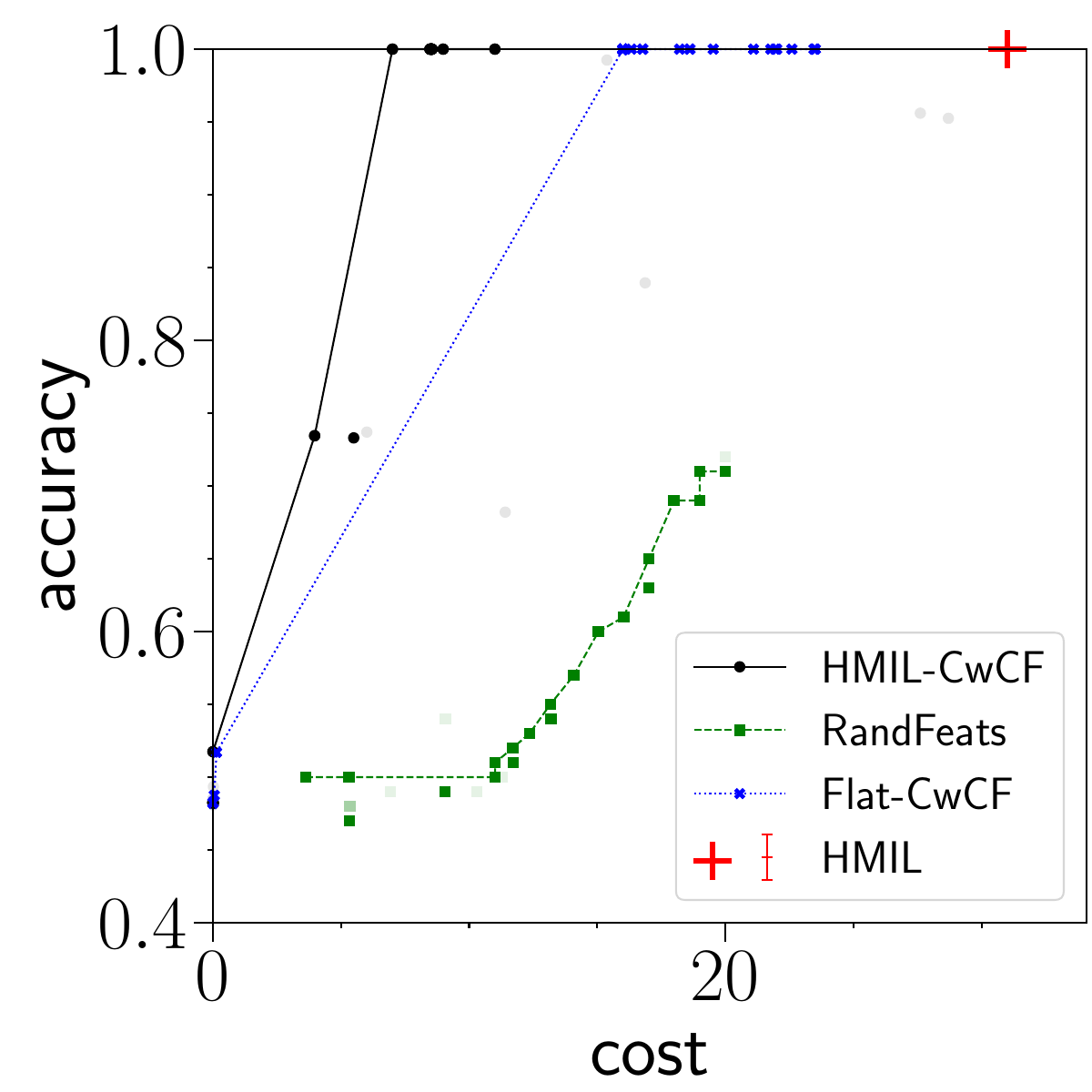}
  \end{minipage}

  \centering
  \caption{Results in the \emph{synthetic} dataset. \textbf{(left)} The process of feature selection. In this example, the algorithm optimally requests the \emph{which\_set} feature, opens \emph{set\_a}, and learns the label in the indicative item. \textbf{(right)} Performance of all algorithms across different budget settings (\emph{x}-axis). We show our method (\emph{HMIL-CwCF}), its ablation with a random policy (\emph{RandFeats}), ablation with flattened data (\emph{Flat-CwCF}), and the \emph{HMIL} algorithm trained with complete information. We train 30 instances per each algorithm (\emph{HMIL-CwCF}, \emph{RandFeats}, and \emph{Flat-CwCF}), each targeting a different budget. We plot the best runs and their Pareto front. We also show the results of all runs as faint points for information about variance. Uniquely for this dataset, the train, validation and test sets are the same.}
  \label{fig:toy_results}
\end{figure}

Figure~\ref{fig:toy_results}-right shows the performance of the tested algorithms in this dataset and Table~\ref{tab:resultsauc} shows the AUTC metric. \emph{HMIL} (the ablation with complete data) reaches 100\% accuracy with a total cost of 31 (cost of all features). The \emph{Flat-HMIL} is able to reduce the cost by acquiring only the correct set, but it has to retrieve all of its objects. Hence, it also reaches 100\% accuracy, but with a cost of 16 (1 for \emph{which\_set} feature, 5 for one of the sets, and 10 for all values inside). Contrarily, the complete \emph{HMIL-CwCF} method reaches 100\% accuracy with only the cost of 7, since it can retrieve only the single indicative value from the correct set. Moreover, it is able to reduce the cost even further by sacrificing accuracy, as seen in the clustering around the cost of 6 and 0.75 accuracy, something that \emph{Flat-HMIL} cannot do. This is one of the strengths of the proposed method -- because it has greater control over which features it acquires, the user can \emph{choose} to sacrifice the accuracy for a lower cost. Lastly, the \emph{RandFeats} method selects the features randomly, and hence, its accuracy is well below \emph{HMIL-CwCF} for corresponding budgets. The accuracy is influenced by the probability of getting the indicative item, which raises with the allocated budget and would reach 100\% with the cost of 31 (we run the method with budgets from $[0, 20]$).

We selected one of the \emph{HMIL-CwCF} models that was trained to reach 100\% accuracy and examined how it behaves (see Figure~\ref{fig:toy_results}-left). We see that it indeed learned to acquire \emph{which\_set} feature, open the corresponding \emph{set\_a} or \emph{set\_b} and select the \emph{item\_value} of the item with \emph{item\_key=1} to learn the right label.

This experiment validates the correct behavior of our method and demonstrates the need for all its parts. Compared to \emph{HMIL} and \emph{Flat-CwCF}, the complete method reaches comparable accuracy with lower cost. Moreover, compared to \emph{Flat-CwCF}, it has better control over which features it requests, achieving better accuracy even in the low-cost region. Finally, the order in which the features are acquired matters, as shown in comparison with \emph{RandFeats}.

\subsection{Experiment B: Threatcrowd}
\begin{figure}[t]
  \includegraphics[width=1.0\linewidth]{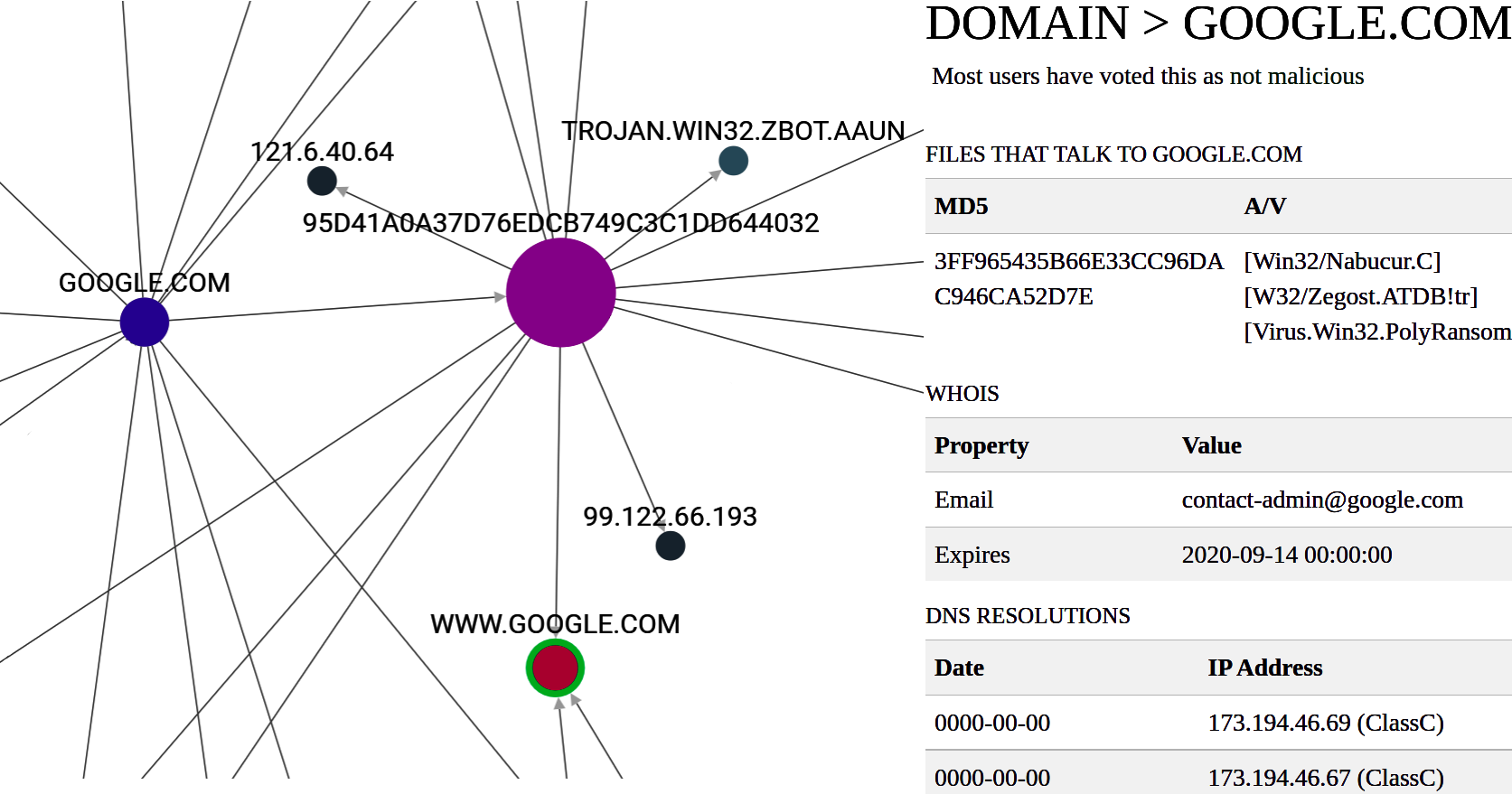}
  \caption{Threatcrowd interface. The left side shows a part of the information graph, unfolded to a limited depth. Various information is available for each node, and the right side displays the information about the currently focused node.}
  \label{fig:threatcrowd_interface}
\end{figure}

Let us focus on one of the real-world cases that motivated this work. Threatcrowd is a service providing rich security-oriented information about domains, such as known malware binaries communicating with the domain (identified by their hashes), WHOIS information, DNS resolutions, subdomains, associated email addresses, and, in some cases, a flag that the domain is known to be malicious (see an example of its interface in Figure~\ref{fig:threatcrowd_interface}). This information is stored in a graph structure, but only a part around the current query is visible to the user. However, the user can easily request more information about the connected objects. For example, after probing the main domain \texttt{google.com}, the user can focus on one of its multiple IP addresses to analyze its reverse DNS lookups, or which other domains are involved with a particular malware.
To make the queries, Threatcrowd provides an API with a limited number of requests per unit of time, which makes it a scarce resource. We are interested in the following task:
\textit{Classify a specified domain using the information provided through the API, minimizing the number of requests.}

\begin{figure}[t]
  \begin{minipage}{0.5\linewidth}
  \end{minipage}
  \begin{minipage}{0.65\linewidth}
  \includegraphics[align=c,scale=0.55]{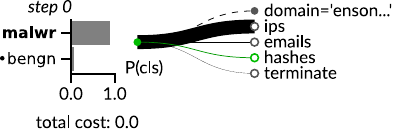}\\
  \includegraphics[align=c,scale=0.55]{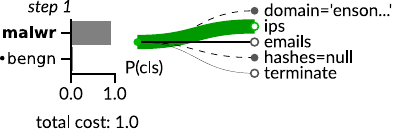}
  \includegraphics[align=c,scale=0.55]{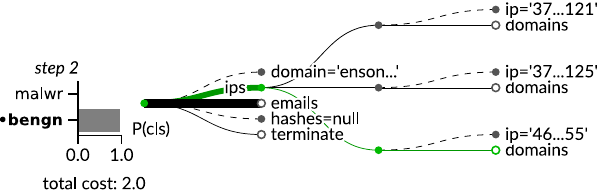}
  \end{minipage}%
  \begin{minipage}{0.35\linewidth}
    \fontsize{8}{8}
    \input{web_100k.txt}
  \end{minipage}
  \includegraphics[align=c,scale=0.55]{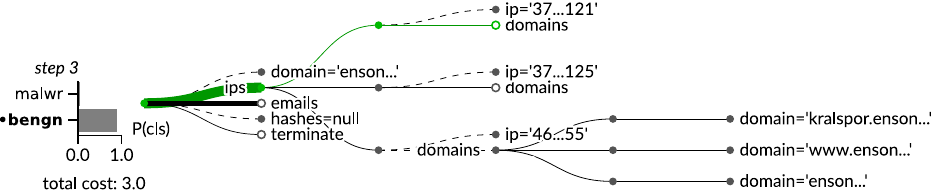}
  \includegraphics[align=c,scale=0.55]{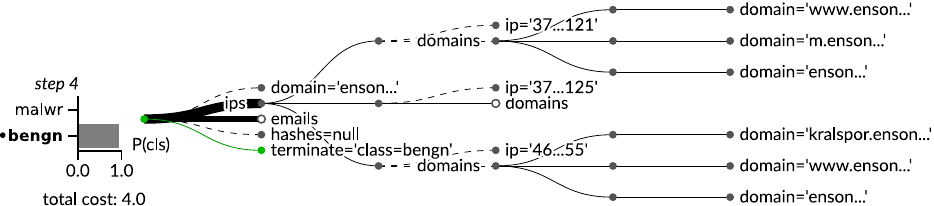}

  \caption{\textbf{(left)} Classification of a potentially malicious domain (\emph{threatcrowd} dataset). At each step, acquired features (full circles) and possible actions (empty circles; unobserved features and terminal action) are shown. The policy is visualized as line thickness and the selection with a green line. The method sequentially requests features: First, it retrieves (step 0) a list of known malware hashes communicating with the domain, then (step 1) a list of associated IP addresses, and finally (steps 2 and 3) performs reverse IP lookups. The correct class is highlighted with a dot. Note that the number of actions differs at each step and the size of sets (IPs, hashes, and emails) differs between samples. \textbf{(right)} The dataset's schema in \textit{feature:type(cost)} format. In this dataset, the costs represent API requests.}
  \label{fig:intro_example}
\end{figure}

To make the experimentation easier and reproducible, we sourced an offline dataset directly from the Threatcrowd service through their API, with their permission. Programmatically, we gathered information about 1~171 domains within a depth of three API requests (including one request for the domain itself) around the original domain and split them into training, validation, and test sets. We chose three API requests because we assume that most of the indicative information is located in the close neighborhood of the root object. Each domain contains its URL as a free feature and a list of associated IP addresses, emails, and malware hashes. These objects can be further reverse-looked up for other domains. This offline dataset perfectly simulates real-life communication with the original service but in a swift and error-free manner. The dataset's schema can be viewed in Figure~\ref{fig:intro_example}-right.

\begin{figure}[t]
  \centering
  {
  \setlength{\tabcolsep}{0pt}\small
  \begin{tabular}{cc}
    \includegraphics[width=0.49\linewidth]{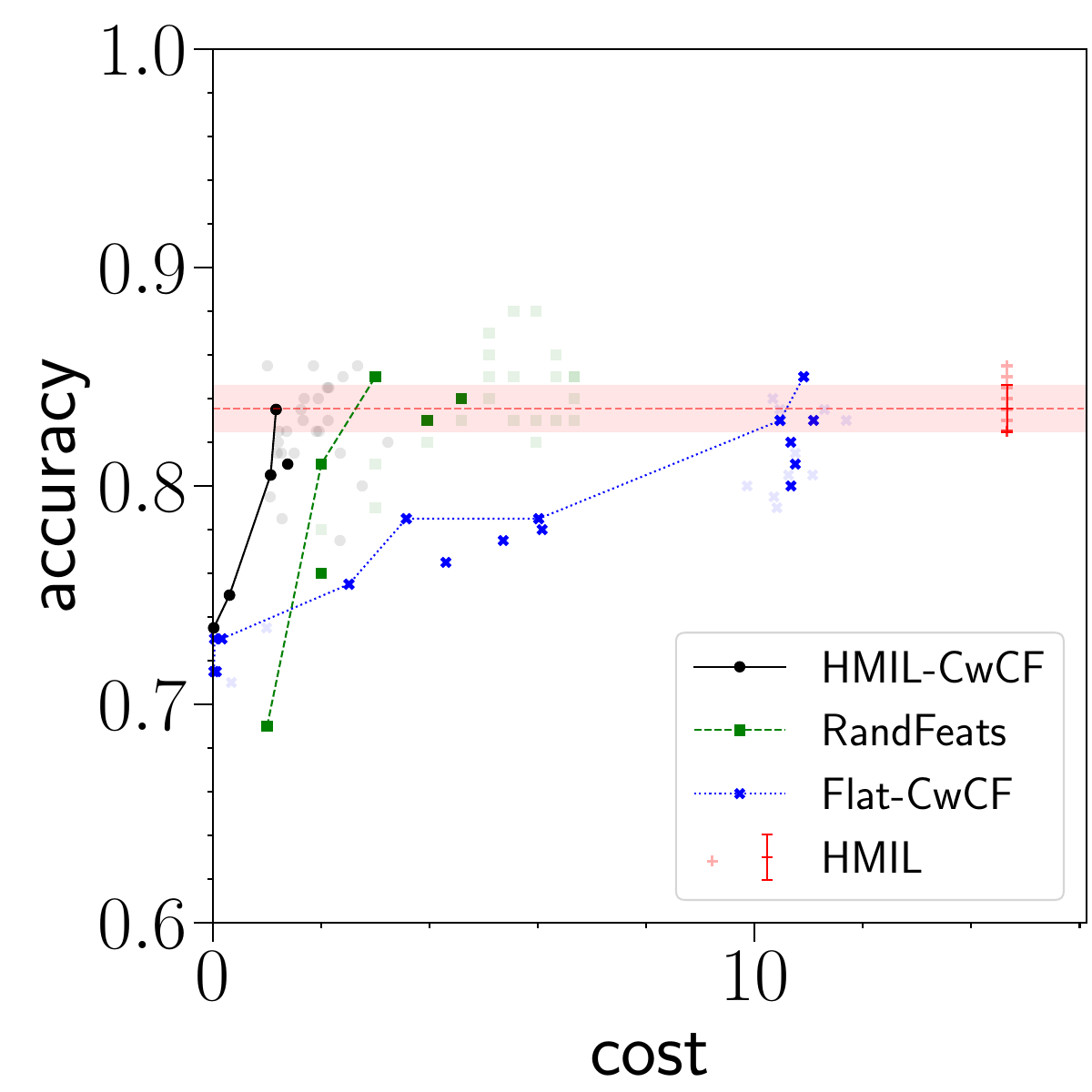} &
    \includegraphics[width=0.49\linewidth]{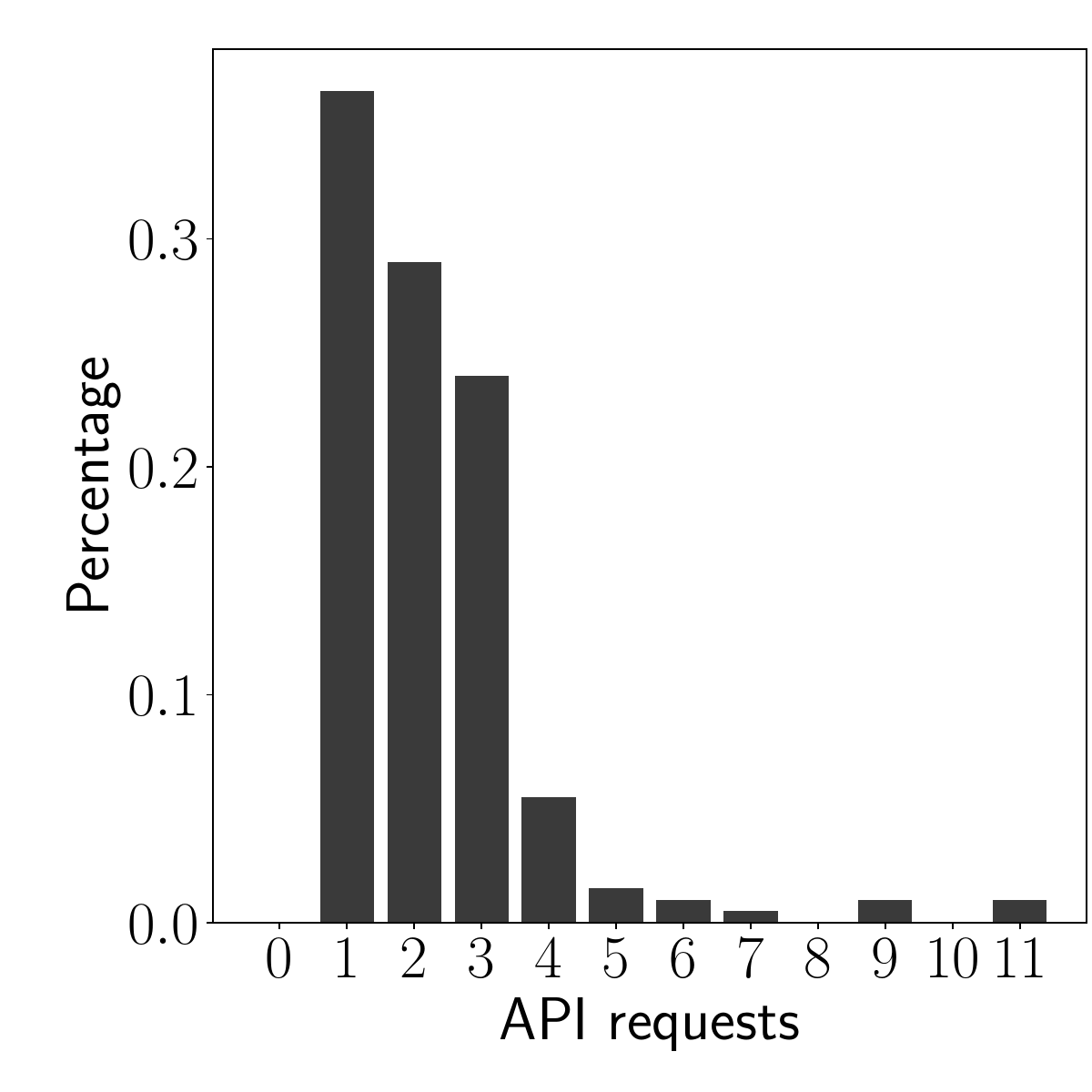}\\
    (a) results & (b) histogram of used API requests
  \end{tabular}
  }

  \caption{\textbf{(a)} Results in \emph{threatcrowd} dataset. The shaded area shows ± one standard deviation around the mean performance of \emph{HMIL} (10 runs), across the whole \emph{x}-axis for comparison. \textbf{(b)} Histogram of used API requests for a trained model that uses two requests on average.}
  \label{fig:results_threatcrowd}
\end{figure}

We ran all of the algorithms with the sourced data, and the results of the experiment are shown in Figure~\ref{fig:results_threatcrowd}a and Table~\ref{tab:resultsauc}. The \emph{HMIL} reaches the mean accuracy of $0.83$ with a cost of $15$ (on average, one needs to make 15 requests to gather all information within the depth of three). Other algorithms reach the same accuracy with a lower cost -- \emph{Flat-CwCF} with $11$, \emph{RandFeats} with $5$, and \emph{HMIL-CwCF} with only $2$ (results are rounded). That means that \textbf{our method needs only two API requests on average} to reach the same accuracy as \emph{HMIL} (which uses complete information), resulting in $7.5\times$ savings. To better understand what these two requests on average mean, we analyzed a single trained model and plotted a histogram of API requests across the whole test set in Figure~\ref{fig:results_threatcrowd}b. For example, with a single request, the algorithm can learn a list of all IP addresses (without further details) or a list of associated malware hashes. The histogram shows that in about 36\% of samples, a single request is enough for classification, 29\% requires two, 23\% three, and 12\% four requests or more. 

Surprisingly, \emph{RandFeats} performs better than \emph{Flat-CwCF}, indicating that only a fraction of information is required, even if randomly sampled. The \emph{Flat-CwCF} algorithm always acquires a complete sub-tree for a specific feature (e.g., a complete list of IP addresses with their reverse lookups, up to the defined depth), resulting in unnecessarily high cost. 

To get better insight into our algorithm's behavior and to showcase its explainability, we visualize how a trained model works with a single sample in Figure~\ref{fig:intro_example}-left. Initially, only the domain name itself is known, without any additional details and the classification would be \emph{malware} if the model decided to terminate at this point. However, the terminal action probability is low, and the model requests a list of malware hashes (there are not any) and a list of IP addresses instead (steps 0 and 1). The prediction changes to \emph{benign}, likely because no malware communicates with the domain nor any malicious IP address is in the list. Still, the model performs a reverse DNS lookup for two IP addresses, which does not change the prediction (steps 2 and 3). Finally, the algorithm finishes with a correct classification \emph{benign}. With four requests, the method was able to probe and classify an unknown domain.

To conclude, this experiment shows that the complete method leads to substantial savings while achieving the same accuracy. When deployed to production, this could mean that the method can classify much more samples with the same budget, or that the budget can be lowered, leading to monetary savings. To apply the model in a real-life scenario, the only thing required is an interface connecting the model's input and decisions with the Threatcrowd API. After that, the model would be able to perform the classification online.
The experiment also verifies that all parts of the algorithm are required. Specifically, the comparison with the \emph{Flat-CwCF} and \emph{RandFeats} baselines showed that flattening the features results in degraded performance and that selecting features based on the knowledge gathered so far is crucial.

\subsection{Experiment C: Other Datasets}
To further evaluate our method, how it scales with small and large datasets and how it performs in binary and multi-class settings, we sourced five more datasets from various domains. Because our method targets a novel problem, we did not find datasets in appropriate format -- i.e., datasets with hierarchical structure and cost information. Therefore, we transformed existing public relational datasets into hierarchical forms by fixing the root object (different for each sample) and expanding its neighborhood into a defined depth. We also manually added costs to the features in a non-uniform way, respecting that in reality, some features are more costly than others (e.g., getting a patient's age is easier than doing a blood test). In practice, the costs would be assigned to the real value of the required resources. The depth of the datasets was chosen so that they completely fit into the memory. 

\subsubsection{Dataset descriptions}
We provide brief descriptions of the used datasets below. The statistics are summarized in Table~\ref{tab:datasets}. For reproducibility, we published the processed versions, along with a library to load them. More details on how we obtained and processed the datasets, their splits, structure, and feature costs are in Supplementary Material~\ref{apx:schemas}.

\begin{table*}[t]
    \centering
    \caption{Statistics of the used datasets. The \emph{features} column shows the number of features (tree leaves) across all completely observed samples in the corresponding dataset.}
    \small
    \begin{tabular}{lrrrrrr}
      \toprule
      dataset         & samples      & class distribution       & features       & depth  \\
                      & (all splits)         &                          & (min/mean/max) &  \\
      \midrule                                
      synthetic       & 12           & 0.5 / 0.5                & 43 / 43.0 / 43 & 2      \\
      threatcrowd     & 1 171        & 0.27 / 0.73              & 4 / 701.7 / 3706 & 3      \\
      hepatitis       & 500          & 0.41 / 0.59              & 7 / 121.7 / 1065 & 2      \\
      mutagenesis     & 188          & 0.34 / 0.66              & 173 / 332.2 / 517 & 3      \\
      ingredients     & 39 774       & 0.01$\sim$0.20           & 2 / 11.8 / 66 & 2      \\
      sap             & 35 602       & 0.5 / 0.5                & 16 / 31.8 / 52 & 2      \\
      stats           & 8 318        &  0.49 / 0.38 / 0.13      & 9 / 52.5 / 21979 & 3      \\
    \end{tabular}
    \label{tab:datasets}
\end{table*}

\textbf{Hepatitis}: A relatively small medical dataset containing patients infected with hepatitis, types B or C. Each patient has various features (e.g., sex, age, etc.) and three sets of indications. The task is to determine the type of disease.

\textbf{Mutagenesis}: Extremely small dataset (188 samples) consisting of molecules that were tested on a particular bacteria for mutagenicity. The molecules themselves have several features and consist of atoms with features and bonds.

\textbf{Ingredients}: Large dataset containing recipes with a single list of ingredients. The task is to determine the type of cuisine of the recipe. The main challenge is to decide when to stop analyzing the ingredients optimally.

\textbf{SAP}:
In this large artificial dataset, the task is to determine whether a particular customer will buy a new product based on a list of past sales. A customer is defined by various features and a list of sales.

\textbf{Stats}: An anonymized content dump from a real website Stats StackExchange. We extracted a list of users to become samples and set an artificial goal of predicting their age category. Each user has several features, a list of posts, and a list of achievements. The posts also contain their own features and a list of tags and comments.

\subsubsection{Results}
\begin{figure}[t]
  \centering 
  {
  \setlength{\tabcolsep}{0pt}\small
  \begin{tabular}{ccc}
    \includegraphics[width=0.33\linewidth]{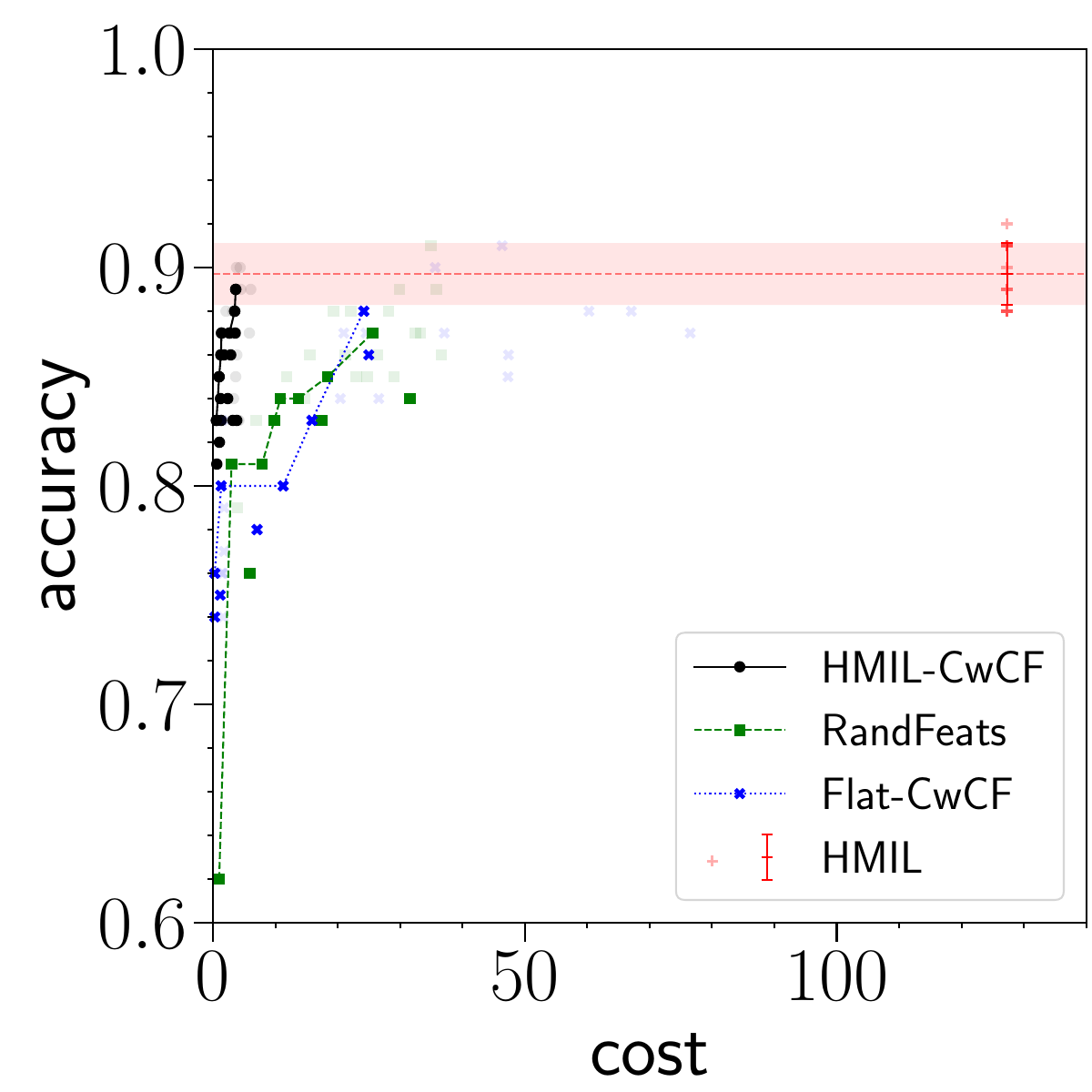} &
    \includegraphics[width=0.33\linewidth]{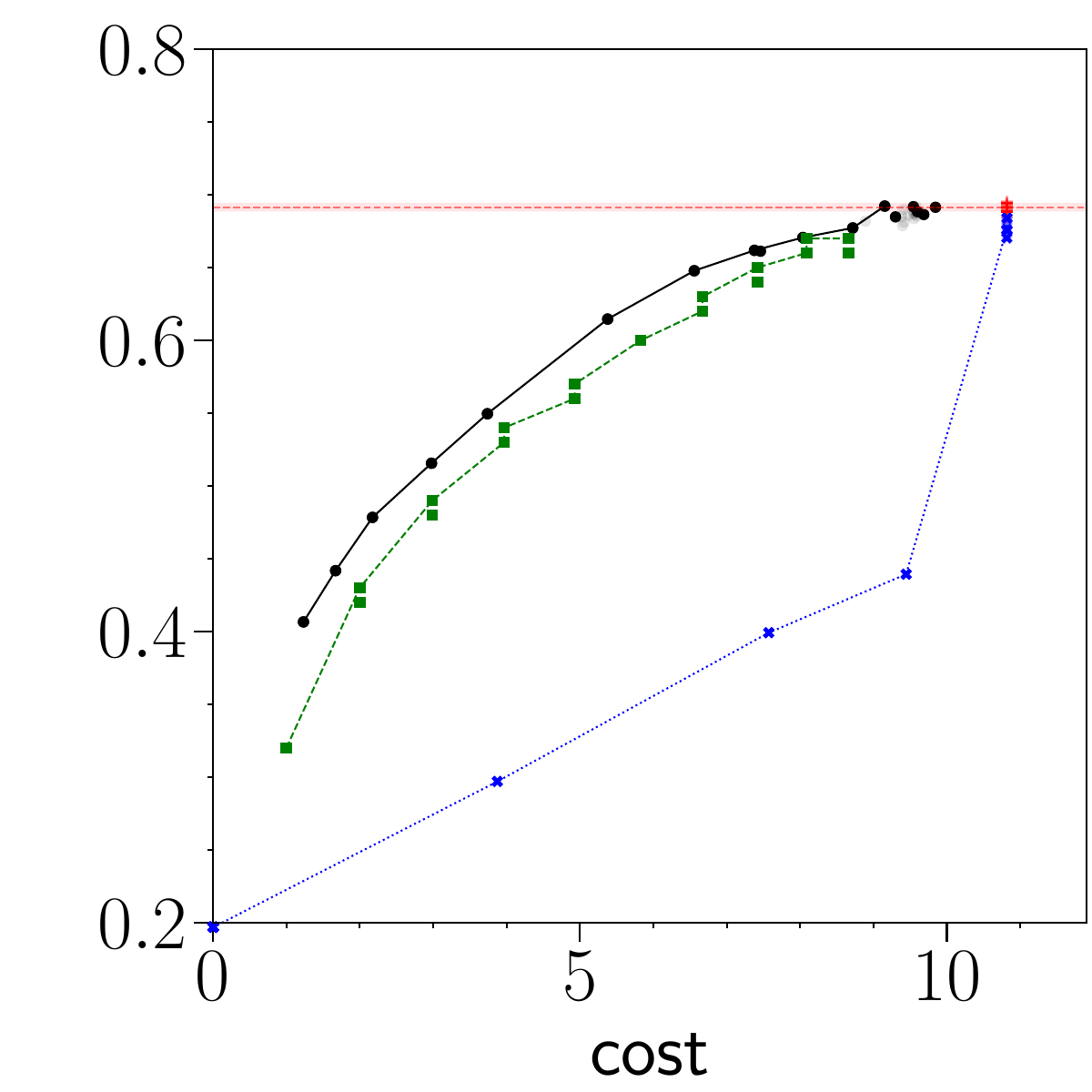} &
    \includegraphics[width=0.33\linewidth]{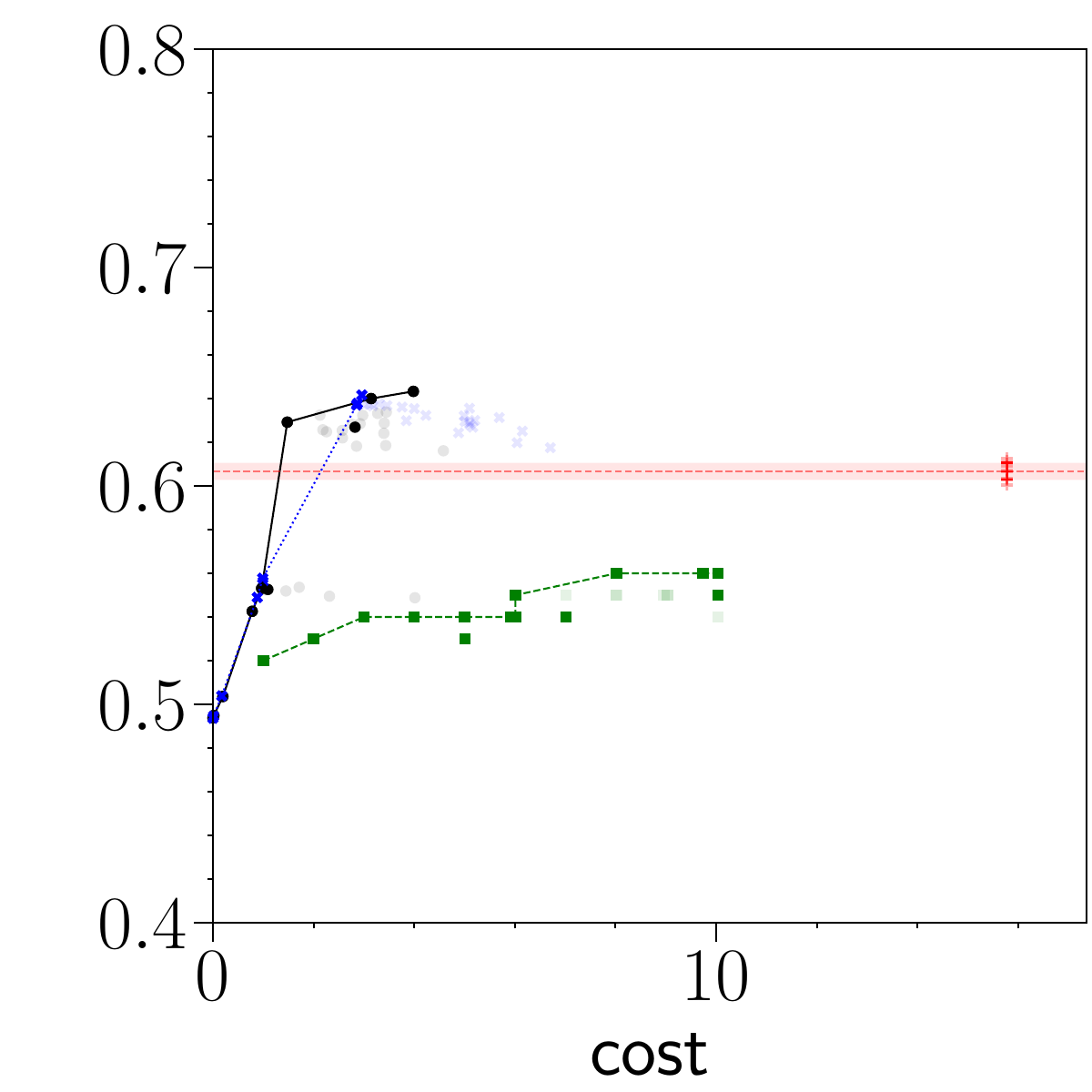} \\
    (a) hepatitis & (b) ingredients & (c) sap \\
    \includegraphics[width=0.33\linewidth]{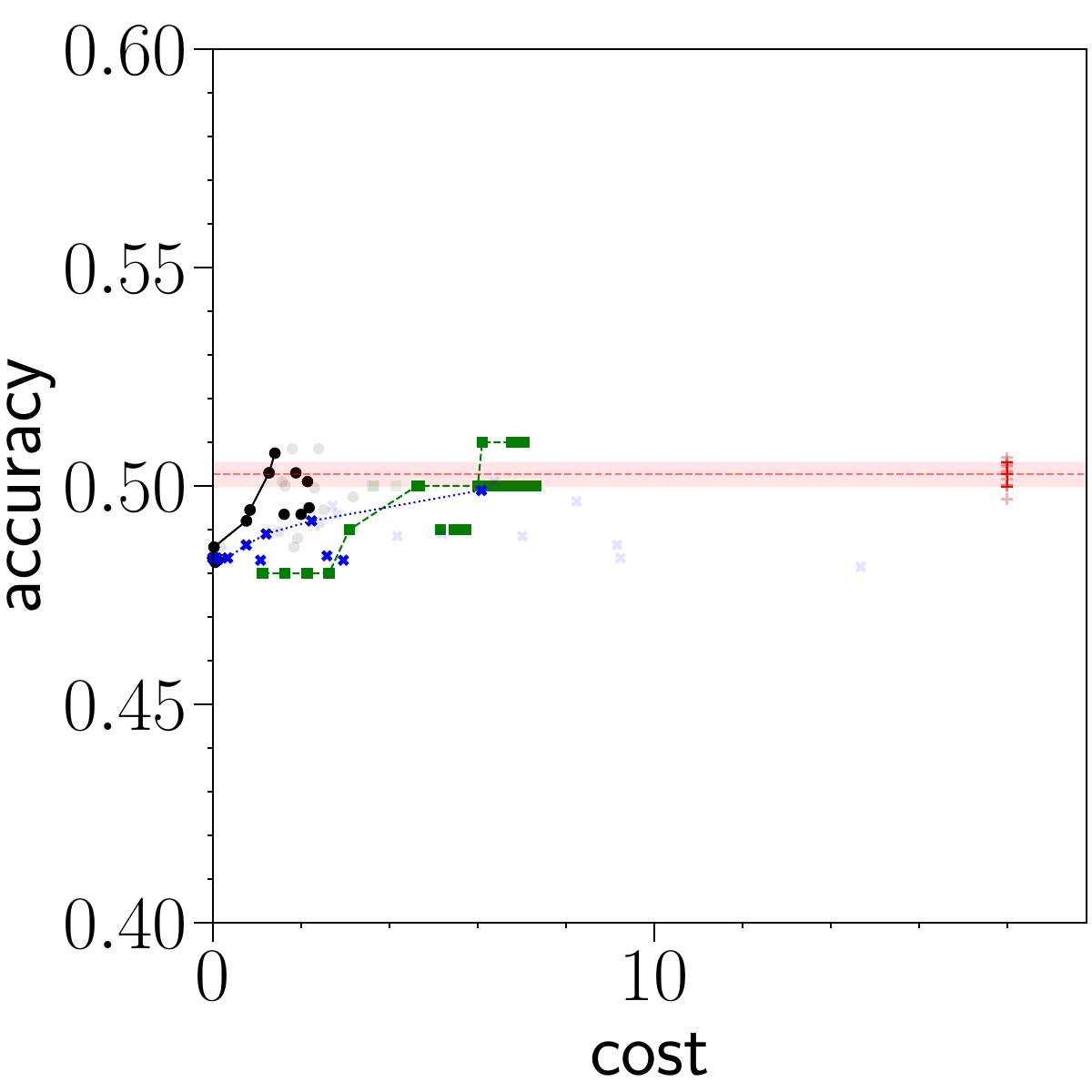} &
    \includegraphics[width=0.33\linewidth]{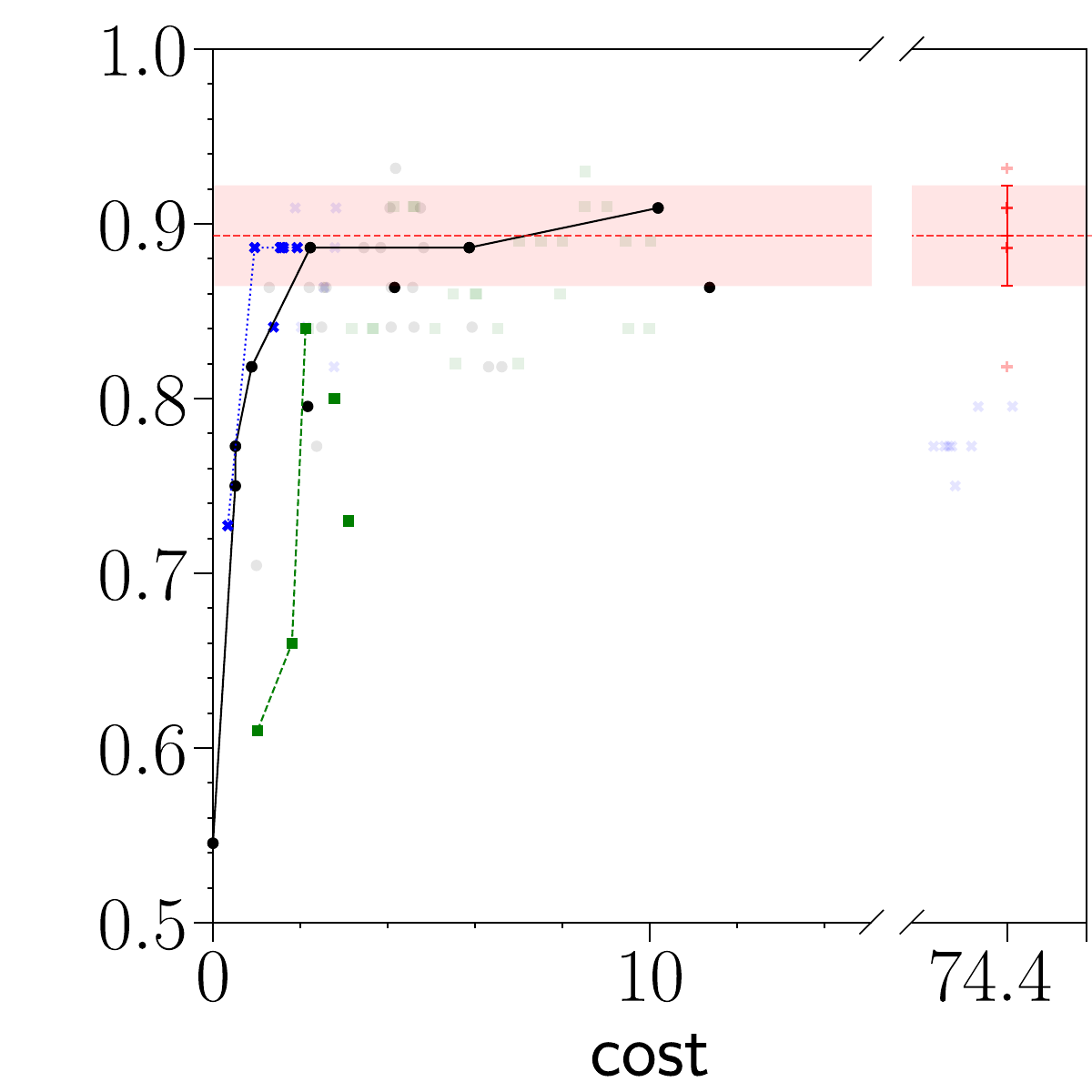} & \\
    (d) stats & (e) mutagenesis \\
  \end{tabular}
  }

  \caption{The performance of the algorithms in five datasets, shown in the cost vs. accuracy plane. We show our method (\emph{HMIL-CwCF}), its ablation with a random policy (\emph{RandFeats}), ablation with flattened data (\emph{Flat-CwCF}) and the \emph{HMIL} algorithm trained with complete information. We train 30 instances per each algorithm (\emph{HMIL-CwCF}, \emph{RandFeats} and \emph{Flat-CwCF}), each targeting a different budget. We plot the best runs, selected using validation sets and their Pareto front. For information about variance, we also show the results of all runs as faint points. The \emph{HMIL} is run 10 times, and we plot the mean ± one standard deviation (the bar visualizes the metrics across the whole range of budgets for comparison). }
  \label{fig:results}
\end{figure}

\begin{table}[t]
    \centering
    \caption{Normalized area under the trade-off curve (AUTC; see Section~\ref{sec:setup} for description).}

    \begin{tabular}{lrrrr}
      \toprule
      dataset & HMIL-CwCF & Flat-CwCF & RandFeats &   HMIL \\
      \midrule
      synthetic   &      \textbf{0.88} &      0.75 &      0.32 &  0.50 \\
      hepatitis   &      \textbf{0.74} &      0.70 &      0.69 &  0.38 \\
      mutagenesis &      \textbf{0.71} &      0.68 &      0.60 &  0.36 \\
      ingredients &      \textbf{0.47} &      0.19 &      0.44 &  0.31 \\
      sap         &      \textbf{0.24} &      0.23 &      0.11 &  0.11 \\
      stats       &      \textbf{0.03} &      0.02 &      \textbf{0.03} &  0.02 \\
      threatcrowd &      \textbf{0.36} &      0.25 &      \textbf{0.36} &  0.18 \\
    \end{tabular}

    \label{tab:resultsauc}
\end{table}

The results are shown in Figure~\ref{fig:results} and in Table~\ref{tab:resultsauc}. Let us select interesting facts and describe them below. The \emph{HMIL} algorithm shows what accuracy is possible to achieve when using all features at once. The variance of its results indicates what should be considered normal in the corresponding dataset. Especially in \emph{hepatitis} and \emph{mutagenesis} (Fig.~\ref{fig:results}ae), the variance of the results is high, which is given by the datasets' small sizes.

The results in \emph{sap} (Fig.~\ref{fig:results}c) are noteworthy. Here, the top accuracy of \emph{HMIL} is exceeded by \emph{HMIL-CwCF} and \emph{Flat-CwCF}. We investigated what is happening and concluded that \emph{HMIL} overfits the training data, despite aggressive regularization -- we tuned the weight decay to maximize the validation accuracy. Surprisingly, \emph{HMIL-CwCF} and \emph{Flat-CwCF} do not suffer from this issue, with fewer features. We hypothesize that the \emph{sap} dataset contains some features deep in the hierarchy that are very informative on the training set, but do not translate well to the test set. The well-performing methods are able to circumvent the issue by selecting fewer features, which results in less overfitting.

Generally, the \emph{HMIL-CwCF} is among the best-performing algorithms in all datasets, i.e., it reaches the same accuracy with lower cost (in \emph{sap} and \emph{mutagenesis}, it performs comparatively to \emph{Flat-CwCF}).  Compared to \emph{HMIL}, the cost is reduced about $26\times$ in \emph{hepatitis}, $1.2\times$ in \emph{ingredients}, $8\times$ in \emph{sap}, $6\times$ in \emph{stats} and $15\times$ in \emph{mutagenesis}, which are significant savings.
\emph{Flat-CwCF} generally exhibits low performance in the low-cost region, due to its limited control over which features it gathers. 

Lastly, let us point out the result of \emph{HMIL-CwCF} compared to \emph{RandFeats} in \emph{ingredients} (Fig.~\ref{fig:results}b). This dataset contains a single set of ingredients, which are objects with a single feature. The best any algorithm can do is to randomly sample the ingredients and stop optimally. While \emph{RandFeats} always uses the given budget, \emph{HMIL-CwCF} can acquire more features in some cases and compensate for that with other samples. Hence, it can reach higher accuracy with the same \emph{average} cost as \emph{RandFeats}. 

The \emph{Flat-CwCF} algorithm can either acquire the whole set of ingredients, or nothing. It achieves different points in Fig.~\ref{fig:results}b by randomization, i.e., it discloses the list of ingredients for some samples, or not for others. Note that the number of ingredients in each recipe varies and ranges from 1 to 65. One could argue that we could use a different encoding of the ingredients -- e.g., one-hot encoding of the ingredients that are in a recipe. However, there are 6707 unique ingredients, while the mean number of ingredients in a recipe is around 11. Flattening the data this way would result in a very sparse and long binary feature vector. Applying the original CwCF method with such data would not work very well, since most of the features would encode a \emph{missing ingredient}.
This was already exemplified in \citep{janisch2019sequential}, where training in a dataset with categorical values encoded to multiple one-hot encoded features (with a length of 40, compared to the required 6707 in case of \emph{ingredient}) took an order of magnitude longer time to train, compared to similarly-sized dataset without such features.

To conclude, the results in Figure~\ref{fig:results} show that our method consistently performs better or comparatively to other methods -- i.e., achieves a similar accuracy with much fewer features. The AUTC metric in Table~\ref{tab:resultsauc} aggregates the performance for the whole range of costs and confirms the conclusion.

\subsection{Remarks}
\subsubsection{Explainability}
Unlike the standard classification algorithms (e.g., \emph{HMIL}), the sequential nature of \emph{HMIL-CwCF} enables easier analysis of its behavior. Figures~\ref{fig:intro_example} and~\ref{fig:toy_results} present two examples of the feature acquisition process and give insight into the agent's decisions. The weights the model assigns to different features in different samples and steps can be used to assess the agent's rationality or learn more about the dataset. We present more visualizations in Supplementary Material~\ref{apx:sampleruns}.

\subsubsection{Classifier pretraining}
\begin{figure}[t]
  \centering
  \includegraphics[width=1.0\linewidth]{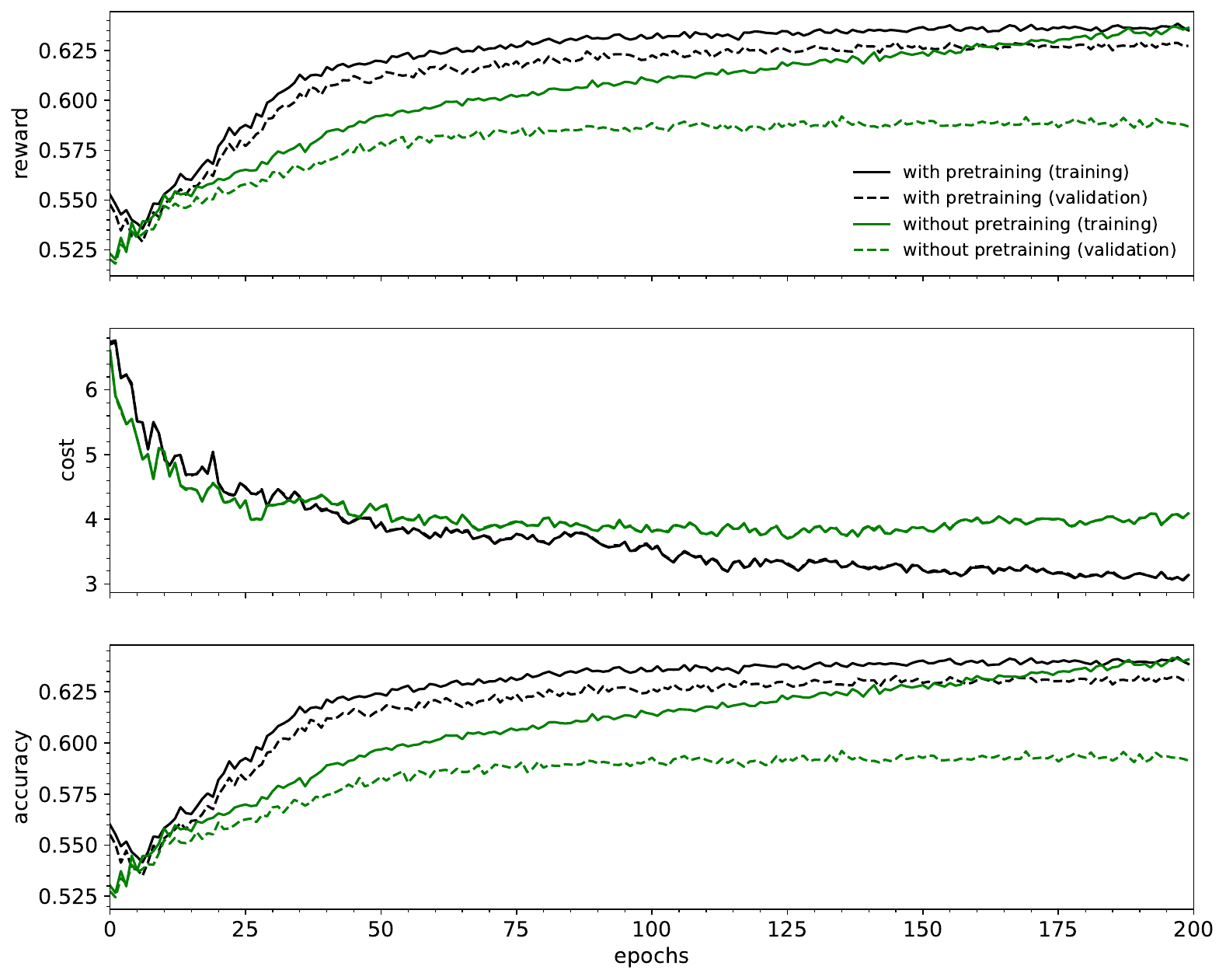}
  \caption{Training of a model, with and without the classifier pretraining. Performed on the \emph{sap} dataset with $\lambda=0.00108264$; an average of 10 runs.}
  \label{fig:pretraining}
\end{figure}

The positive role of pretraining was already established in the original CwCF paper \citep{janisch2019classification}. However, as we separate the classifier from the RL algorithm, it is worth to assess how the situation changes. We performed an ablation experiment with the \emph{sap} dataset and a fixed $\lambda$, where we ran the experiment 10 times with and without pretraining. The results in Figure~\ref{fig:pretraining} show that the pretraining improves the speed of convergence and the performance on validation data.

\subsubsection{Computational requirements}
We measured the training times using a single core of Intel Xeon Gold 6146 3.2GHz and 4GB of memory. We used only CPU because the most time-consuming part of the training was the environment's simulation and it cannot benefit from the use of GPU. The measured times are displayed in Table~\ref{tab:trainingtimes}. We show the \emph{synthetic} dataset separately because it was much faster to learn.
Note that the training times are for a single run (i.e., a single point in Figure~\ref{fig:results}), but the runs are independent and are easily parallelized. After training, the inference time is negligible for all methods.

Note that while the training time of \emph{HMIL-CwCF} is much longer than in the case of \emph{HMIL}, it is easily compensated by the fact that our method can save a large amount of resources if correctly deployed. Moreover, computational power rises exponentially every year (resulting in faster training), while resources like CO\textsubscript{2} production, patients' discomfort, or response time of an antivirus software only gain importance.

\begin{table}[t]
    \centering
    \caption{Training times for a single instance (i.e., single setting of $\lambda$ in \emph{HMIL-CwCF}). Note that most of the time is spent on simulating the environment.}

    \begin{tabular}{lrrrr}
      \toprule
      dataset & HMIL-CwCF & RandFeats & Flat-CwCF & HMIL \\
      \midrule                                
      synthetic   & 1 hour   & 30 minutes & 1 hour & 1 minute \\
      other (average) & 19 hours & 14 hours & 9 hours & 1 hour \\
    \end{tabular}

    \label{tab:trainingtimes}
\end{table}

\section{Discussion}\label{sec:discussion}
\paragraph{Comparison with Graph Neural Networks (GNNs).}
Instead of HMIL, we could use a GNN to perform the input embedding. However, note that the data we work with are hierarchical and constructed around a central root. Hence it makes sense to model the data as \emph{trees}, not as general graphs, and use a method tailored to work with trees. In our case, generic message passing is unnecessary, and a single pass from leaves to the tree's root is sufficient to embed all information correctly. \citet{mandlik2020mapping} provides a deeper discussion about using HMIL and GNNs in sample-centric applications.

In some special cases, the same object could be located in multiple places (e.g., the same IP address accessible by multiple paths). In our method, we still handle the sample as a tree. If such a situation occurs, the data have to be \emph{unrolled}, i.e., different places of the same object are considered to be different objects.

\paragraph{Is the depth of the tested datasets sufficient?}
We argue that most of the relevant information is within the near neighborhood of the central object of interest. Increasing the depth exponentially increases the available feature space and space requirements and slows down training. As the experiments showed that there are substantial differences between the methods, we conclude that the used depth is sufficient.

\paragraph{How to obtain credible cost assignment?}
In a real-life application, it should be possible to measure the costs of features up front. For example, the time required to perform an experiment, electricity consumed to retrieve a piece of data, or, as in the Threatcrowd experiment, every feature can represent a single API request.

\paragraph{Advantages and disadvantages of the proposed method}
Our solution provides the following advantages, some of which are inherited from the original CwCF framework:
\begin{itemize}\small
    \item It directly optimizes the objective in eq.~(1) and although the deep RL has not the same theoretical guarantees as tabular RL, it searches for the optimal solution. In contrast, some related work used heuristics (e.g., proxy rewards \citep{kachuee2019opportunistic} in the flat CwCF case) -- such algorithms are not guaranteed to aim for the optimal solution.
    \item The used HMIL algorithm used to process the hierarchical input is theoretically sound -- \citet{pevny2019approximation} generalizes the universal approximation theorem \citep{hornik1991approximation,leshno1993multilayer} to HMIL networks.
    \item As our method is based on a standard deep RL technique, its performance is likely to be improved with advancements in the RL field itself, since it is an actively developed area.
    \item The novel method can directly utilize many of the extensions developed for CwCF. This includes (1) problems with hard budget, (2) specifying the budget directly and automatic search for an optimal $\lambda$, (3) missing features (e.g., features of some objects may be inaccessible, possibly because the training data is incomplete), and (4) using an external high-performance classifier as one of the features. Points (1-3) are discussed in \citep{janisch2019sequential}, (4) is explored in \citep{janisch2019classification}.
    \item The original CwCF paper \citep{janisch2019sequential} has already established the competitive performance of the method in the flat data case. Therefore, we believe that the novel algorithm serves as a highly competitive baseline as well.
\end{itemize} 
Below, we state the drawbacks of our algorithm we are aware of:
\begin{itemize}\small
    \item Being RL-based, the algorithm is sample inefficient, i.e., it requires a long training. As mentioned, training in the more complicated datasets took about 19 hours on average.
    \item Data must be hierarchical, e.g., it must not contain references to the same object in different places in the hierarchy, nor cycles. As mentioned in the discussion about GNNs, if such structures appear in the data, it must be \emph{unrolled} (e.g., the same object would have to be copied to different places) so that the result is hierarchical.
    \item With some datasets, there could be non-negligible variance in the performance of trained models. The user is advised to repeat training several times and select the best-performing model, based on validation data.
\end{itemize} 

\paragraph{Alternative approaches}
Generally, there are two ways to make the existing algorithms work with the hierarchical data: a) modifying the data, b) modifying the algorithm. Below, we suggest several different approaches to these options. Keep in mind that each of these suggestions would require substantial research to implement, and might not be possible at all.

a) Modifying the data can be done in the way we did in the case of \emph{Flat-CwCF}, but there could be other ways, for example:
\begin{itemize}
    \item It may be possible to decrease the granularity of choice to the set level by considering each path in the \emph{schema} as a separate feature. While this approach would result in a fixed number of features for all samples, it brings several issues. For example, since sets can contain multiple objects, it is unclear how to choose one of them. An algorithm selecting the objects randomly would have inherently lesser control over which objects to select, and would not be able to utilize possible conditional dependencies between objects' features. In the \emph{RandFeats} baseline, we have already shown that such loss of control results in degraded performance. Second, if it is allowed to get the same feature multiple times (to cover different objects in a set), it is unclear how to aggregate and process these multiple values. 
    \item Another way could be to treat all features in the tree as a set of tuples \emph{(path, type, value)}, each encoded into a $\mathbf R^n$ space, and use algorithms designed to process sets \citep{shim2018joint}. While this approach would preserve all information, it is unclear how to efficiently encode paths of various lengths that can branch in sets, or values of different types.
    \item Also, one could manually engineer features based on the known data structure. However, this step is laborious, suboptimal, and may be difficult to apply, because the individual samples vary in size of their sets. Note that the standard approaches to feature selection do not work with hierarchical data.
\end{itemize}

b) Let us also discuss the possible modification of the existing algorithms, where the problem is twofold. First, the algorithm needs to be modified to accept hierarchical data with varying size. In some cases, it could be solved by embedding the data sample into a smaller, fixed space, e.g., with the HMIL algorithm, as we did in our case. However, many algorithms for the CwCF problem depend on access to the actual feature values, such as decision trees \citep{maliah2018mdp}, random forests \citep{nan2015feature,nan2016pruning,nan2017adaptive} or cascade classifiers \citep{xu2014classifier} and may not work with such transformations.
Second, the modified algorithm needs to be able to select features within the hierarchy. This could be done through direct selection of the corresponding output (as we do in our method, or as the \citep{shim2018joint} would do with the formerly proposed modification), or through some other way of identifying the specific feature (possibly by returning its encoded path). 

Again, while believe that many of these problems are solvable, they would require non-trivial further research.
\section{Conclusion}\label{sec:conclusion}
We presented an augmented Classification with Costly Features framework that can process hierarchically structured data. Contrarily to existing algorithms, our method can process this kind of data in its natural form and select features directly in the hierarchy. In several experiments, we demonstrated that our method substantially outperforms an algorithm that uses complete information, in terms of the cost of used features. We also showed how the original CwCF would work if the data was flattened so the method could process it. As our augmented HMIL-CwCF model has the ability to choose features with greater precision, it leads to superior performance. In a separate experiment, we applied our method to a real-life problem of classification of malicious web domains, where it also outperformed the other algorithms. The sequential nature of our algorithm and its hierarchical action selection contribute to its explainability, as the features are semantically grouped, and the user can view which of them are considered important at different time steps.

\backmatter

\bmhead{Supplementary information}
The complete code for our algorithm and baselines, along with all prepared datasets and scripts to run the experiments, is publicly available at \url{https://github.com/jaromiru/rcwcf}.

\bmhead{Acknowledgments}
The GPU used for this research was donated by the NVIDIA Corporation. Some computational resources were supplied by the project \textquote{e-Infrastruktura CZ} (e-INFRA LM2018140) provided within the program Projects of Large Research, Development and Innovations Infrastructures. The authors acknowledge the support of the OP VVV funded project CZ.02.1.01/0.0/0.0/16\_019/0000765 ``Research Center for Informatics''. This research was supported by The Czech Science Foundation (grants no. 22-32620S and 22-26655S).

\bmhead{Authors’ contributions}
Jaromír Janisch designed and implemented the method, performed the experiments and wrote the manuscript. Tomáš Pevný and Viliam Lisý supervised and consulted the work.

\pagebreak

\bibliography{sn-bibliography}


\begin{thebibliography}{74}
\ifx \bisbn   \undefined \def \bisbn  #1{ISBN #1}\fi
\ifx \binits  \undefined \def \binits#1{#1}\fi
\ifx \bauthor  \undefined \def \bauthor#1{#1}\fi
\ifx \batitle  \undefined \def \batitle#1{#1}\fi
\ifx \bjtitle  \undefined \def \bjtitle#1{#1}\fi
\ifx \bvolume  \undefined \def \bvolume#1{\textbf{#1}}\fi
\ifx \byear  \undefined \def \byear#1{#1}\fi
\ifx \bissue  \undefined \def \bissue#1{#1}\fi
\ifx \bfpage  \undefined \def \bfpage#1{#1}\fi
\ifx \blpage  \undefined \def \blpage #1{#1}\fi
\ifx \burl  \undefined \def \burl#1{\textsf{#1}}\fi
\ifx \doiurl  \undefined \def \doiurl#1{\url{https://doi.org/#1}}\fi
\ifx \betal  \undefined \def \betal{\textit{et al.}}\fi
\ifx \binstitute  \undefined \def \binstitute#1{#1}\fi
\ifx \binstitutionaled  \undefined \def \binstitutionaled#1{#1}\fi
\ifx \bctitle  \undefined \def \bctitle#1{#1}\fi
\ifx \beditor  \undefined \def \beditor#1{#1}\fi
\ifx \bpublisher  \undefined \def \bpublisher#1{#1}\fi
\ifx \bbtitle  \undefined \def \bbtitle#1{#1}\fi
\ifx \bedition  \undefined \def \bedition#1{#1}\fi
\ifx \bseriesno  \undefined \def \bseriesno#1{#1}\fi
\ifx \blocation  \undefined \def \blocation#1{#1}\fi
\ifx \bsertitle  \undefined \def \bsertitle#1{#1}\fi
\ifx \bsnm \undefined \def \bsnm#1{#1}\fi
\ifx \bsuffix \undefined \def \bsuffix#1{#1}\fi
\ifx \bparticle \undefined \def \bparticle#1{#1}\fi
\ifx \barticle \undefined \def \barticle#1{#1}\fi
\bibcommenthead
\ifx \bconfdate \undefined \def \bconfdate #1{#1}\fi
\ifx \botherref \undefined \def \botherref #1{#1}\fi
\ifx \url \undefined \def \url#1{\textsf{#1}}\fi
\ifx \bchapter \undefined \def \bchapter#1{#1}\fi
\ifx \bbook \undefined \def \bbook#1{#1}\fi
\ifx \bcomment \undefined \def \bcomment#1{#1}\fi
\ifx \oauthor \undefined \def \oauthor#1{#1}\fi
\ifx \citeauthoryear \undefined \def \citeauthoryear#1{#1}\fi
\ifx \endbibitem  \undefined \def \endbibitem {}\fi
\ifx \bconflocation  \undefined \def \bconflocation#1{#1}\fi
\ifx \arxivurl  \undefined \def \arxivurl#1{\textsf{#1}}\fi
\csname PreBibitemsHook\endcsname

\bibitem{janisch2019sequential}
\begin{barticle}
\bauthor{\bsnm{Janisch}, \binits{J.}},
\bauthor{\bsnm{Pevný}, \binits{T.}},
\bauthor{\bsnm{Lisý}, \binits{V.}}:
\batitle{Classification with costly features as a sequential decision-making
  problem}.
\bjtitle{Machine Learning}
\bvolume{109}(\bissue{8}),
\bfpage{1587}--\blpage{1615}
(\byear{2020})
\end{barticle}
\endbibitem

\bibitem{shim2018joint}
\begin{bchapter}
\bauthor{\bsnm{Shim}, \binits{H.}},
\bauthor{\bsnm{Hwang}, \binits{S.J.}},
\bauthor{\bsnm{Yang}, \binits{E.}}:
\bctitle{Joint active feature acquisition and classification with variable-size
  set encoding}.
In: \bbtitle{Advances in Neural Information Processing Systems},
pp. \bfpage{1375}--\blpage{1385}
(\byear{2018})
\end{bchapter}
\endbibitem

\bibitem{dulac2012sequential}
\begin{barticle}
\bauthor{\bsnm{Dulac-Arnold}, \binits{G.}},
\bauthor{\bsnm{Denoyer}, \binits{L.}},
\bauthor{\bsnm{Preux}, \binits{P.}},
\bauthor{\bsnm{Gallinari}, \binits{P.}}:
\batitle{Sequential approaches for learning datum-wise sparse representations}.
\bjtitle{Machine learning}
\bvolume{89}(\bissue{1-2}),
\bfpage{87}--\blpage{122}
(\byear{2012})
\end{barticle}
\endbibitem

\bibitem{xu2012greedy}
\begin{bchapter}
\bauthor{\bsnm{Xu}, \binits{Z.}},
\bauthor{\bsnm{Weinberger}, \binits{K.}},
\bauthor{\bsnm{Chapelle}, \binits{O.}}:
\bctitle{The greedy miser: learning under test-time budgets}.
In: \bbtitle{Proceedings of the 29th International Coference on International
  Conference on Machine Learning},
pp. \bfpage{1299}--\blpage{1306}
(\byear{2012}).
\bcomment{Omnipress}
\end{bchapter}
\endbibitem

\bibitem{kusner2014feature}
\begin{bchapter}
\bauthor{\bsnm{Kusner}, \binits{M.}},
\bauthor{\bsnm{Chen}, \binits{W.}},
\bauthor{\bsnm{Zhou}, \binits{Q.}},
\bauthor{\bsnm{Xu}, \binits{Z.}},
\bauthor{\bsnm{Weinberger}, \binits{K.}},
\bauthor{\bsnm{Chen}, \binits{Y.}}:
\bctitle{Feature-cost sensitive learning with submodular trees of classifiers}.
In: \bbtitle{AAAI Conference on Artificial Intelligence},
pp. \bfpage{1939}--\blpage{1945}
(\byear{2014})
\end{bchapter}
\endbibitem

\bibitem{xu2013cost}
\begin{bchapter}
\bauthor{\bsnm{Xu}, \binits{Z.}},
\bauthor{\bsnm{Kusner}, \binits{M.}},
\bauthor{\bsnm{Weinberger}, \binits{K.}},
\bauthor{\bsnm{Chen}, \binits{M.}}:
\bctitle{Cost-sensitive tree of classifiers}.
In: \bbtitle{International Conference on Machine Learning},
pp. \bfpage{133}--\blpage{141}
(\byear{2013})
\end{bchapter}
\endbibitem

\bibitem{xu2014classifier}
\begin{barticle}
\bauthor{\bsnm{Xu}, \binits{Z.}},
\bauthor{\bsnm{Kusner}, \binits{M.}},
\bauthor{\bsnm{Weinberger}, \binits{K.}},
\bauthor{\bsnm{Chen}, \binits{M.}},
\bauthor{\bsnm{Chapelle}, \binits{O.}}:
\batitle{Classifier cascades and trees for minimizing feature evaluation cost}.
\bjtitle{Journal of Machine Learning Research}
\bvolume{15}(\bissue{1}),
\bfpage{2113}--\blpage{2144}
(\byear{2014})
\end{barticle}
\endbibitem

\bibitem{nan2015feature}
\begin{bchapter}
\bauthor{\bsnm{Nan}, \binits{F.}},
\bauthor{\bsnm{Wang}, \binits{J.}},
\bauthor{\bsnm{Saligrama}, \binits{V.}}:
\bctitle{Feature-budgeted random forest}.
In: \bbtitle{International Conference on Machine Learning},
pp. \bfpage{1983}--\blpage{1991}
(\byear{2015})
\end{bchapter}
\endbibitem

\bibitem{nan2016pruning}
\begin{bchapter}
\bauthor{\bsnm{Nan}, \binits{F.}},
\bauthor{\bsnm{Wang}, \binits{J.}},
\bauthor{\bsnm{Saligrama}, \binits{V.}}:
\bctitle{Pruning random forests for prediction on a budget}.
In: \bbtitle{Advances in Neural Information Processing Systems},
pp. \bfpage{2334}--\blpage{2342}
(\byear{2016})
\end{bchapter}
\endbibitem

\bibitem{nan2017adaptive}
\begin{bchapter}
\bauthor{\bsnm{Nan}, \binits{F.}},
\bauthor{\bsnm{Saligrama}, \binits{V.}}:
\bctitle{Adaptive classification for prediction under a budget}.
In: \bbtitle{Advances in Neural Information Processing Systems},
pp. \bfpage{4730}--\blpage{4740}
(\byear{2017})
\end{bchapter}
\endbibitem

\bibitem{contardo2016recurrent}
\begin{bchapter}
\bauthor{\bsnm{Contardo}, \binits{G.}},
\bauthor{\bsnm{Denoyer}, \binits{L.}},
\bauthor{\bsnm{Artieres}, \binits{T.}}:
\bctitle{Recurrent neural networks for adaptive feature acquisition}.
In: \bbtitle{International Conference on Neural Information Processing},
pp. \bfpage{591}--\blpage{599}
(\byear{2016}).
\bcomment{Springer}
\end{bchapter}
\endbibitem

\bibitem{wang2014lp}
\begin{bchapter}
\bauthor{\bsnm{Wang}, \binits{J.}},
\bauthor{\bsnm{Trapeznikov}, \binits{K.}},
\bauthor{\bsnm{Saligrama}, \binits{V.}}:
\bctitle{An lp for sequential learning under budgets}.
In: \bbtitle{Artificial Intelligence and Statistics},
pp. \bfpage{987}--\blpage{995}
(\byear{2014})
\end{bchapter}
\endbibitem

\bibitem{wang2014model}
\begin{bchapter}
\bauthor{\bsnm{Wang}, \binits{J.}},
\bauthor{\bsnm{Bolukbasi}, \binits{T.}},
\bauthor{\bsnm{Trapeznikov}, \binits{K.}},
\bauthor{\bsnm{Saligrama}, \binits{V.}}:
\bctitle{Model selection by linear programming}.
In: \bbtitle{European Conference on Computer Vision},
pp. \bfpage{647}--\blpage{662}
(\byear{2014}).
\bcomment{Springer}
\end{bchapter}
\endbibitem

\bibitem{ji2007cost}
\begin{barticle}
\bauthor{\bsnm{Ji}, \binits{S.}},
\bauthor{\bsnm{Carin}, \binits{L.}}:
\batitle{Cost-sensitive feature acquisition and classification}.
\bjtitle{Pattern Recognition}
\bvolume{40}(\bissue{5}),
\bfpage{1474}--\blpage{1485}
(\byear{2007})
\end{barticle}
\endbibitem

\bibitem{janisch2019classification}
\begin{bchapter}
\bauthor{\bsnm{Janisch}, \binits{J.}},
\bauthor{\bsnm{Pevný}, \binits{T.}},
\bauthor{\bsnm{Lisý}, \binits{V.}}:
\bctitle{Classification with costly features using deep reinforcement
  learning}.
In: \bbtitle{Proceedings of 33\textsuperscript{rd} AAAI Conference on
  Artificial Intelligence}
(\byear{2019})
\end{bchapter}
\endbibitem

\bibitem{peng2018refuel}
\begin{bchapter}
\bauthor{\bsnm{Peng}, \binits{Y.-S.}},
\bauthor{\bsnm{Tang}, \binits{K.-F.}},
\bauthor{\bsnm{Lin}, \binits{H.-T.}},
\bauthor{\bsnm{Chang}, \binits{E.}}:
\bctitle{Refuel: Exploring sparse features in deep reinforcement learning for
  fast disease diagnosis}.
In: \bbtitle{Advances in Neural Information Processing Systems},
pp. \bfpage{7322}--\blpage{7331}
(\byear{2018})
\end{bchapter}
\endbibitem

\bibitem{lee2020interactive}
\begin{bchapter}
\bauthor{\bsnm{Lee}, \binits{M.H.}},
\bauthor{\bsnm{Siewiorek}, \binits{D.P.}},
\bauthor{\bsnm{Smailagic}, \binits{A.}},
\bauthor{\bsnm{Bernardino}, \binits{A.}},
\bauthor{\bparticle{Berm{\'u}dez~i} \bsnm{Badia}, \binits{S.}}:
\bctitle{Interactive hybrid approach to combine machine and human intelligence
  for personalized rehabilitation assessment}.
In: \bbtitle{Proceedings of the ACM Conference on Health, Inference, and
  Learning},
pp. \bfpage{160}--\blpage{169}
(\byear{2020})
\end{bchapter}
\endbibitem

\bibitem{song2018deep}
\begin{bchapter}
\bauthor{\bsnm{Song}, \binits{C.}},
\bauthor{\bsnm{Chen}, \binits{C.}},
\bauthor{\bsnm{Li}, \binits{Y.}},
\bauthor{\bsnm{Wu}, \binits{X.}}:
\bctitle{Deep reinforcement learning apply in electromyography data
  classification}.
In: \bbtitle{2018 IEEE International Conference on Cyborg and Bionic Systems
  (CBS)},
pp. \bfpage{505}--\blpage{510}
(\byear{2018}).
\bcomment{IEEE}
\end{bchapter}
\endbibitem

\bibitem{vivar2020peri}
\begin{bchapter}
\bauthor{\bsnm{Vivar}, \binits{G.}},
\bauthor{\bsnm{Mullakaeva}, \binits{K.}},
\bauthor{\bsnm{Zwergal}, \binits{A.}},
\bauthor{\bsnm{Navab}, \binits{N.}},
\bauthor{\bsnm{Ahmadi}, \binits{S.-A.}}:
\bctitle{Peri-diagnostic decision support through cost-efficient feature
  acquisition at test-time}.
In: \bbtitle{International Conference on Medical Image Computing and
  Computer-Assisted Intervention},
pp. \bfpage{572}--\blpage{581}
(\byear{2020}).
\bcomment{Springer}
\end{bchapter}
\endbibitem

\bibitem{lee2020co}
\begin{barticle}
\bauthor{\bsnm{Lee}, \binits{M.H.}},
\bauthor{\bsnm{Siewiorek}, \binits{D.P.}},
\bauthor{\bsnm{Smailagic}, \binits{A.}},
\bauthor{\bsnm{Bernardino}, \binits{A.}},
\bauthor{\bparticle{Berm{\'u}dez~i} \bsnm{Badia}, \binits{S.}}:
\batitle{Co-design and evaluation of an intelligent decision support system for
  stroke rehabilitation assessment}.
\bjtitle{Proceedings of the ACM on Human-Computer Interaction}
\bvolume{4}(\bissue{CSCW2}),
\bfpage{1}--\blpage{27}
(\byear{2020})
\end{barticle}
\endbibitem

\bibitem{shpakova2021probabilistic}
\begin{barticle}
\bauthor{\bsnm{Shpakova}, \binits{T.}},
\bauthor{\bsnm{Sokolovska}, \binits{N.}}:
\batitle{Probabilistic personalised cascade with abstention}.
\bjtitle{Pattern Recognition Letters}
\bvolume{147},
\bfpage{8}--\blpage{15}
(\byear{2021})
\end{barticle}
\endbibitem

\bibitem{zhu2020learning}
\begin{bchapter}
\bauthor{\bsnm{Zhu}, \binits{M.}},
\bauthor{\bsnm{Zhu}, \binits{H.}}:
\bctitle{Learning a cost-effective strategy on incomplete medical data}.
In: \bbtitle{International Conference on Database Systems for Advanced
  Applications},
pp. \bfpage{175}--\blpage{191}
(\byear{2020}).
\bcomment{Springer}
\end{bchapter}
\endbibitem

\bibitem{goldstein2020target}
\begin{barticle}
\bauthor{\bsnm{Goldstein}, \binits{O.}},
\bauthor{\bsnm{Kachuee}, \binits{M.}},
\bauthor{\bsnm{Karkkainen}, \binits{K.}},
\bauthor{\bsnm{Sarrafzadeh}, \binits{M.}}:
\batitle{Target-focused feature selection using uncertainty measurements in
  healthcare data}.
\bjtitle{ACM Transactions on Computing for Healthcare}
\bvolume{1}(\bissue{3}),
\bfpage{1}--\blpage{17}
(\byear{2020})
\end{barticle}
\endbibitem

\bibitem{erion2022cost}
\begin{botherref}
\oauthor{\bsnm{Erion}, \binits{G.}},
\oauthor{\bsnm{Janizek}, \binits{J.D.}},
\oauthor{\bsnm{Hudelson}, \binits{C.}},
\oauthor{\bsnm{Utarnachitt}, \binits{R.B.}},
\oauthor{\bsnm{McCoy}, \binits{A.M.}},
\oauthor{\bsnm{Sayre}, \binits{M.R.}},
\oauthor{\bsnm{White}, \binits{N.J.}},
\oauthor{\bsnm{Lee}, \binits{S.-I.}}:
A cost-aware framework for the development of {AI} models for healthcare
  applications.
Nature Biomedical Engineering,
1--15
(2022)
\end{botherref}
\endbibitem

\bibitem{banerjee2020deep}
\begin{bchapter}
\bauthor{\bsnm{Banerjee}, \binits{S.}},
\bauthor{\bsnm{Pratiher}, \binits{S.}},
\bauthor{\bsnm{Chattoraj}, \binits{S.}},
\bauthor{\bsnm{Gupta}, \binits{R.}},
\bauthor{\bsnm{Patra}, \binits{P.}},
\bauthor{\bsnm{Saikia}, \binits{B.}},
\bauthor{\bsnm{Thakur}, \binits{S.}},
\bauthor{\bsnm{Mondal}, \binits{S.}},
\bauthor{\bsnm{Mukherjee}, \binits{A.}}:
\bctitle{Deep reinforcement learning for variability prediction in latent heat
  flux from low-cost meteorological parameters}.
In: \bbtitle{Optics and Photonics for Advanced Dimensional Metrology},
vol. \bseriesno{11352},
pp. \bfpage{305}--\blpage{311}
(\byear{2020}).
\bcomment{SPIE}
\end{bchapter}
\endbibitem

\bibitem{ali2020reinforcement}
\begin{barticle}
\bauthor{\bsnm{Ali}, \binits{B.}},
\bauthor{\bsnm{Moriyama}, \binits{K.}},
\bauthor{\bsnm{Kalintha}, \binits{W.}},
\bauthor{\bsnm{Numao}, \binits{M.}},
\bauthor{\bsnm{Fukui}, \binits{K.-I.}}:
\batitle{Reinforcement learning based metric filtering for evolutionary
  distance metric learning}.
\bjtitle{Intelligent Data Analysis}
\bvolume{24}(\bissue{6}),
\bfpage{1345}--\blpage{1364}
(\byear{2020})
\end{barticle}
\endbibitem

\bibitem{xu2021crowd}
\begin{barticle}
\bauthor{\bsnm{Xu}, \binits{J.}},
\bauthor{\bsnm{Sun}, \binits{Z.}},
\bauthor{\bsnm{Ma}, \binits{C.}}:
\batitle{Crowd aware summarization of surveillance videos by deep reinforcement
  learning}.
\bjtitle{Multimedia Tools and Applications}
\bvolume{80}(\bissue{4}),
\bfpage{6121}--\blpage{6141}
(\byear{2021})
\end{barticle}
\endbibitem

\bibitem{liu2018dependency}
\begin{bchapter}
\bauthor{\bsnm{Liu}, \binits{X.}},
\bauthor{\bsnm{Kumar}, \binits{B.}},
\bauthor{\bsnm{Yang}, \binits{C.}},
\bauthor{\bsnm{Tang}, \binits{Q.}},
\bauthor{\bsnm{You}, \binits{J.}}:
\bctitle{Dependency-aware attention control for unconstrained face recognition
  with image sets}.
In: \bbtitle{Proceedings of the European Conference on Computer Vision (ECCV)},
pp. \bfpage{548}--\blpage{565}
(\byear{2018})
\end{bchapter}
\endbibitem

\bibitem{badr2022enabling}
\begin{botherref}
\oauthor{\bsnm{Badr}, \binits{Y.}}:
Enabling intrusion detection systems with dueling double deep {Q}-learning.
Digital Transformation and Society
(ahead-of-print)
(2022)
\end{botherref}
\endbibitem

\bibitem{zaheer2017deep}
\begin{bchapter}
\bauthor{\bsnm{Zaheer}, \binits{M.}},
\bauthor{\bsnm{Kottur}, \binits{S.}},
\bauthor{\bsnm{Ravanbakhsh}, \binits{S.}},
\bauthor{\bsnm{Poczos}, \binits{B.}},
\bauthor{\bsnm{Salakhutdinov}, \binits{R.R.}},
\bauthor{\bsnm{Smola}, \binits{A.J.}}:
\bctitle{Deep sets}.
In: \bbtitle{Advances in Neural Information Processing Systems},
pp. \bfpage{3391}--\blpage{3401}
(\byear{2017})
\end{bchapter}
\endbibitem

\bibitem{pevny2016discriminative}
\begin{bchapter}
\bauthor{\bsnm{Pevný}, \binits{T.}},
\bauthor{\bsnm{Somol}, \binits{P.}}:
\bctitle{Discriminative models for multi-instance problems with tree
  structure}.
In: \bbtitle{Proceedings of the 2016 ACM Workshop on Artificial Intelligence
  and Security},
pp. \bfpage{83}--\blpage{91}
(\byear{2016}).
\bcomment{ACM}
\end{bchapter}
\endbibitem

\bibitem{morin2005hierarchical}
\begin{bchapter}
\bauthor{\bsnm{Morin}, \binits{F.}},
\bauthor{\bsnm{Bengio}, \binits{Y.}}:
\bctitle{Hierarchical probabilistic neural network language model.}
In: \bbtitle{Aistats},
vol. \bseriesno{5},
pp. \bfpage{246}--\blpage{252}
(\byear{2005}).
\bcomment{Citeseer}
\end{bchapter}
\endbibitem

\bibitem{wang2015efficient}
\begin{bchapter}
\bauthor{\bsnm{Wang}, \binits{J.}},
\bauthor{\bsnm{Trapeznikov}, \binits{K.}},
\bauthor{\bsnm{Saligrama}, \binits{V.}}:
\bctitle{Efficient learning by directed acyclic graph for resource constrained
  prediction}.
In: \bbtitle{Advances in Neural Information Processing Systems},
pp. \bfpage{2152}--\blpage{2160}
(\byear{2015})
\end{bchapter}
\endbibitem

\bibitem{trapeznikov2013supervised}
\begin{bchapter}
\bauthor{\bsnm{Trapeznikov}, \binits{K.}},
\bauthor{\bsnm{Saligrama}, \binits{V.}}:
\bctitle{Supervised sequential classification under budget constraints}.
In: \bbtitle{Artificial Intelligence and Statistics},
pp. \bfpage{581}--\blpage{589}
(\byear{2013})
\end{bchapter}
\endbibitem

\bibitem{liyanage2021dynamic}
\begin{botherref}
\oauthor{\bsnm{Liyanage}, \binits{Y.W.}},
\oauthor{\bsnm{Zois}, \binits{D.-S.}},
\oauthor{\bsnm{Chelmis}, \binits{C.}}:
Dynamic instance-wise joint feature selection and classification.
IEEE Transactions on Artificial Intelligence
(2021)
\end{botherref}
\endbibitem

\bibitem{tan1993cost}
\begin{barticle}
\bauthor{\bsnm{Tan}, \binits{M.}}:
\batitle{Cost-sensitive learning of classification knowledge and its
  applications in robotics}.
\bjtitle{Machine Learning}
\bvolume{13}(\bissue{1}),
\bfpage{7}--\blpage{33}
(\byear{1993})
\end{barticle}
\endbibitem

\bibitem{li2021active}
\begin{bchapter}
\bauthor{\bsnm{Li}, \binits{Y.}},
\bauthor{\bsnm{Oliva}, \binits{J.}}:
\bctitle{Active feature acquisition with generative surrogate models}.
In: \bbtitle{International Conference on Machine Learning},
pp. \bfpage{6450}--\blpage{6459}
(\byear{2021}).
\bcomment{PMLR}
\end{bchapter}
\endbibitem

\bibitem{bayer2005integrating}
\begin{barticle}
\bauthor{\bsnm{Bayer-Zubek}, \binits{V.}},
\bauthor{\bsnm{Dietterich}, \binits{T.G.}}:
\batitle{Integrating learning from examples into the search for diagnostic
  policies}.
\bjtitle{Journal of Artificial Intelligence Research}
\bvolume{24},
\bfpage{263}--\blpage{303}
(\byear{2005})
\end{barticle}
\endbibitem

\bibitem{kapoor2005learning}
\begin{bchapter}
\bauthor{\bsnm{Kapoor}, \binits{A.}},
\bauthor{\bsnm{Greiner}, \binits{R.}}:
\bctitle{Learning and classifying under hard budgets}.
In: \bbtitle{European Conference on Machine Learning},
pp. \bfpage{170}--\blpage{181}
(\byear{2005}).
\bcomment{Springer}
\end{bchapter}
\endbibitem

\bibitem{deng2007bandit}
\begin{bchapter}
\bauthor{\bsnm{Deng}, \binits{K.}},
\bauthor{\bsnm{Bourke}, \binits{C.}},
\bauthor{\bsnm{Scott}, \binits{S.}},
\bauthor{\bsnm{Sunderman}, \binits{J.}},
\bauthor{\bsnm{Zheng}, \binits{Y.}}:
\bctitle{Bandit-based algorithms for budgeted learning}.
In: \bbtitle{Seventh IEEE International Conference on Data Mining (ICDM 2007)},
pp. \bfpage{463}--\blpage{468}
(\byear{2007}).
\bcomment{IEEE}
\end{bchapter}
\endbibitem

\bibitem{cesa2011efficient}
\begin{barticle}
\bauthor{\bsnm{Cesa-Bianchi}, \binits{N.}},
\bauthor{\bsnm{Shalev-Shwartz}, \binits{S.}},
\bauthor{\bsnm{Shamir}, \binits{O.}}:
\batitle{Efficient learning with partially observed attributes}.
\bjtitle{Journal of Machine Learning Research}
\bvolume{12}(\bissue{Oct}),
\bfpage{2857}--\blpage{2878}
(\byear{2011})
\end{barticle}
\endbibitem

\bibitem{zolghadr2013online}
\begin{bchapter}
\bauthor{\bsnm{Zolghadr}, \binits{N.}},
\bauthor{\bsnm{Bart{\'o}k}, \binits{G.}},
\bauthor{\bsnm{Greiner}, \binits{R.}},
\bauthor{\bsnm{Gy{\"o}rgy}, \binits{A.}},
\bauthor{\bsnm{Szepesv{\'a}ri}, \binits{C.}}:
\bctitle{Online learning with costly features and labels.}
In: \bbtitle{Advances in Neural Information Processing Systems},
pp. \bfpage{1241}--\blpage{1249}
(\byear{2013})
\end{bchapter}
\endbibitem

\bibitem{kachuee2019opportunistic}
\begin{bchapter}
\bauthor{\bsnm{Kachuee}, \binits{M.}},
\bauthor{\bsnm{Goldstein}, \binits{O.}},
\bauthor{\bsnm{Karkkainen}, \binits{K.}},
\bauthor{\bsnm{Darabi}, \binits{S.}},
\bauthor{\bsnm{Sarrafzadeh}, \binits{M.}}:
\bctitle{Opportunistic learning: Budgeted cost-sensitive learning from data
  streams}.
In: \bbtitle{International Conference on Learning Representations}
(\byear{2019})
\end{bchapter}
\endbibitem

\bibitem{guyon2003introduction}
\begin{barticle}
\bauthor{\bsnm{Guyon}, \binits{I.}},
\bauthor{\bsnm{Elisseeff}, \binits{A.}}:
\batitle{An introduction to variable and feature selection}.
\bjtitle{Journal of machine learning research}
\bvolume{3}(\bissue{Mar}),
\bfpage{1157}--\blpage{1182}
(\byear{2003})
\end{barticle}
\endbibitem

\bibitem{maldonado2017cost}
\begin{barticle}
\bauthor{\bsnm{Maldonado}, \binits{S.}},
\bauthor{\bsnm{P{\'e}rez}, \binits{J.}},
\bauthor{\bsnm{Bravo}, \binits{C.}}:
\batitle{Cost-based feature selection for support vector machines: An
  application in credit scoring}.
\bjtitle{European Journal of Operational Research}
\bvolume{261}(\bissue{2}),
\bfpage{656}--\blpage{665}
(\byear{2017})
\end{barticle}
\endbibitem

\bibitem{bolon2014framework}
\begin{barticle}
\bauthor{\bsnm{Bol{\'o}n-Canedo}, \binits{V.}},
\bauthor{\bsnm{Porto-D{\'\i}az}, \binits{I.}},
\bauthor{\bsnm{S{\'a}nchez-Maro{\~n}o}, \binits{N.}},
\bauthor{\bsnm{Alonso-Betanzos}, \binits{A.}}:
\batitle{A framework for cost-based feature selection}.
\bjtitle{Pattern Recognition}
\bvolume{47}(\bissue{7}),
\bfpage{2481}--\blpage{2489}
(\byear{2014})
\end{barticle}
\endbibitem

\bibitem{pevny2017using}
\begin{bchapter}
\bauthor{\bsnm{Pevný}, \binits{T.}},
\bauthor{\bsnm{Somol}, \binits{P.}}:
\bctitle{Using neural network formalism to solve multiple-instance problems}.
In: \bbtitle{International Symposium on Neural Networks},
pp. \bfpage{135}--\blpage{142}
(\byear{2017}).
\bcomment{Springer}
\end{bchapter}
\endbibitem

\bibitem{pevny2019approximation}
\begin{botherref}
\oauthor{\bsnm{Pevný}, \binits{T.}},
\oauthor{\bsnm{Kovařík}, \binits{V.}}:
Approximation capability of neural networks on spaces of probability measures
  and tree-structured domains.
arXiv preprint arXiv:1906.00764
(2019)
\end{botherref}
\endbibitem

\bibitem{mandlik2022jsongrinder}
\begin{barticle}
\bauthor{\bsnm{Mandlík}, \binits{{\v{S}}.}},
\bauthor{\bsnm{Račinský}, \binits{M.}},
\bauthor{\bsnm{Lisý}, \binits{V.}},
\bauthor{\bsnm{Pevný}, \binits{T.}}:
\batitle{Jsongrinder. jl: automated differentiable neural architecture for
  embedding arbitrary json data}.
\bjtitle{Journal of Machine Learning Research}
\bvolume{23}(\bissue{298}),
\bfpage{1}--\blpage{5}
(\byear{2022})
\end{barticle}
\endbibitem

\bibitem{tang2020discretizing}
\begin{bchapter}
\bauthor{\bsnm{Tang}, \binits{Y.}},
\bauthor{\bsnm{Agrawal}, \binits{S.}}:
\bctitle{Discretizing continuous action space for on-policy optimization}.
In: \bbtitle{Proceedings of the AAAI Conference on Artificial Intelligence},
vol. \bseriesno{34},
pp. \bfpage{5981}--\blpage{5988}
(\byear{2020})
\end{bchapter}
\endbibitem

\bibitem{chen2019effective}
\begin{botherref}
\oauthor{\bsnm{Chen}, \binits{Y.-E.}},
\oauthor{\bsnm{Tang}, \binits{K.-F.}},
\oauthor{\bsnm{Peng}, \binits{Y.-S.}},
\oauthor{\bsnm{Chang}, \binits{E.Y.}}:
Effective medical test suggestions using deep reinforcement learning.
arXiv preprint arXiv:1905.12916
(2019)
\end{botherref}
\endbibitem

\bibitem{metz2017discrete}
\begin{botherref}
\oauthor{\bsnm{Metz}, \binits{L.}},
\oauthor{\bsnm{Ibarz}, \binits{J.}},
\oauthor{\bsnm{Jaitly}, \binits{N.}},
\oauthor{\bsnm{Davidson}, \binits{J.}}:
Discrete sequential prediction of continuous actions for deep rl.
arXiv preprint arXiv:1705.05035
(2017)
\end{botherref}
\endbibitem

\bibitem{goodman2001classes}
\begin{bchapter}
\bauthor{\bsnm{Goodman}, \binits{J.}}:
\bctitle{Classes for fast maximum entropy training}.
In: \bbtitle{2001 IEEE International Conference on Acoustics, Speech, and
  Signal Processing. Proceedings (Cat. No. 01CH37221)},
vol. \bseriesno{1},
pp. \bfpage{561}--\blpage{564}
(\byear{2001}).
\bcomment{IEEE}
\end{bchapter}
\endbibitem

\bibitem{mnih2016asynchronous}
\begin{bchapter}
\bauthor{\bsnm{Mnih}, \binits{V.}},
\bauthor{\bsnm{Badia}, \binits{A.P.}},
\bauthor{\bsnm{Mirza}, \binits{M.}},
\bauthor{\bsnm{Graves}, \binits{A.}},
\bauthor{\bsnm{Lillicrap}, \binits{T.}},
\bauthor{\bsnm{Harley}, \binits{T.}},
\bauthor{\bsnm{Silver}, \binits{D.}},
\bauthor{\bsnm{Kavukcuoglu}, \binits{K.}}:
\bctitle{Asynchronous methods for deep reinforcement learning}.
In: \bbtitle{International Conference on Machine Learning},
pp. \bfpage{1928}--\blpage{1937}
(\byear{2016})
\end{bchapter}
\endbibitem

\bibitem{sutton2018reinforcement}
\begin{bbook}
\bauthor{\bsnm{Sutton}, \binits{R.S.}},
\bauthor{\bsnm{Barto}, \binits{A.G.}}:
\bbtitle{Reinforcement Learning: An Introduction (2nd Ed.)}.
\bpublisher{MIT press},
\blocation{Cambridge, MA}
(\byear{2018})
\end{bbook}
\endbibitem

\bibitem{schulman2015trust}
\begin{bchapter}
\bauthor{\bsnm{Schulman}, \binits{J.}},
\bauthor{\bsnm{Levine}, \binits{S.}},
\bauthor{\bsnm{Abbeel}, \binits{P.}},
\bauthor{\bsnm{Jordan}, \binits{M.}},
\bauthor{\bsnm{Moritz}, \binits{P.}}:
\bctitle{Trust region policy optimization}.
In: \bbtitle{International Conference on Machine Learning},
pp. \bfpage{1889}--\blpage{1897}
(\byear{2015}).
\bcomment{PMLR}
\end{bchapter}
\endbibitem

\bibitem{schulman2017proximal}
\begin{botherref}
\oauthor{\bsnm{Schulman}, \binits{J.}},
\oauthor{\bsnm{Wolski}, \binits{F.}},
\oauthor{\bsnm{Dhariwal}, \binits{P.}},
\oauthor{\bsnm{Radford}, \binits{A.}},
\oauthor{\bsnm{Klimov}, \binits{O.}}:
Proximal policy optimization algorithms.
arXiv preprint arXiv:1707.06347
(2017)
\end{botherref}
\endbibitem

\bibitem{zhou2018graph}
\begin{botherref}
\oauthor{\bsnm{Zhou}, \binits{J.}},
\oauthor{\bsnm{Cui}, \binits{G.}},
\oauthor{\bsnm{Zhang}, \binits{Z.}},
\oauthor{\bsnm{Yang}, \binits{C.}},
\oauthor{\bsnm{Liu}, \binits{Z.}},
\oauthor{\bsnm{Sun}, \binits{M.}}:
Graph neural networks: {A} review of methods and applications.
arXiv preprint arXiv:1812.08434
(2018)
\end{botherref}
\endbibitem

\bibitem{hamilton2017inductive}
\begin{bchapter}
\bauthor{\bsnm{Hamilton}, \binits{W.}},
\bauthor{\bsnm{Ying}, \binits{Z.}},
\bauthor{\bsnm{Leskovec}, \binits{J.}}:
\bctitle{Inductive representation learning on large graphs}.
In: \bbtitle{Advances in Neural Information Processing Systems},
pp. \bfpage{1024}--\blpage{1034}
(\byear{2017})
\end{bchapter}
\endbibitem

\bibitem{perozzi2014deepwalk}
\begin{bchapter}
\bauthor{\bsnm{Perozzi}, \binits{B.}},
\bauthor{\bsnm{Al-Rfou}, \binits{R.}},
\bauthor{\bsnm{Skiena}, \binits{S.}}:
\bctitle{Deepwalk: Online learning of social representations}.
In: \bbtitle{Proceedings of the 20th ACM SIGKDD International Conference on
  Knowledge Discovery and Data Mining},
pp. \bfpage{701}--\blpage{710}
(\byear{2014}).
\bcomment{ACM}
\end{bchapter}
\endbibitem

\bibitem{kipf2016semi}
\begin{botherref}
\oauthor{\bsnm{Kipf}, \binits{T.N.}},
\oauthor{\bsnm{Welling}, \binits{M.}}:
Semi-supervised classification with graph convolutional networks.
arXiv preprint arXiv:1609.02907
(2016)
\end{botherref}
\endbibitem

\bibitem{mnih2015human}
\begin{barticle}
\bauthor{\bsnm{Mnih}, \binits{V.}},
\bauthor{\bsnm{Kavukcuoglu}, \binits{K.}},
\bauthor{\bsnm{Silver}, \binits{D.}},
\bauthor{\bsnm{Rusu}, \binits{A.A.}},
\bauthor{\bsnm{Veness}, \binits{J.}},
\bauthor{\bsnm{Bellemare}, \binits{M.G.}},
\bauthor{\bsnm{Graves}, \binits{A.}},
\bauthor{\bsnm{Riedmiller}, \binits{M.}},
\bauthor{\bsnm{Fidjeland}, \binits{A.K.}},
\bauthor{\bsnm{Ostrovski}, \binits{G.}}, \betal:
\batitle{Human-level control through deep reinforcement learning}.
\bjtitle{Nature}
\bvolume{518}(\bissue{7540}),
\bfpage{529}--\blpage{533}
(\byear{2015})
\end{barticle}
\endbibitem

\bibitem{van2016deep}
\begin{bchapter}
\bauthor{\bsnm{Van~Hasselt}, \binits{H.}},
\bauthor{\bsnm{Guez}, \binits{A.}},
\bauthor{\bsnm{Silver}, \binits{D.}}:
\bctitle{Deep reinforcement learning with double {Q}-learning}.
In: \bbtitle{AAAI Conference on Artificial Intelligence},
pp. \bfpage{2094}--\blpage{2100}
(\byear{2016})
\end{bchapter}
\endbibitem

\bibitem{wang2016dueling}
\begin{bchapter}
\bauthor{\bsnm{Wang}, \binits{Z.}},
\bauthor{\bsnm{Schaul}, \binits{T.}},
\bauthor{\bsnm{Hessel}, \binits{M.}},
\bauthor{\bsnm{Hasselt}, \binits{H.}},
\bauthor{\bsnm{Lanctot}, \binits{M.}},
\bauthor{\bsnm{Freitas}, \binits{N.}}:
\bctitle{Dueling network architectures for deep reinforcement learning}.
In: \bbtitle{International Conference on Machine Learning},
pp. \bfpage{1995}--\blpage{2003}
(\byear{2016})
\end{bchapter}
\endbibitem

\bibitem{munos2016safe}
\begin{bchapter}
\bauthor{\bsnm{Munos}, \binits{R.}},
\bauthor{\bsnm{Stepleton}, \binits{T.}},
\bauthor{\bsnm{Harutyunyan}, \binits{A.}},
\bauthor{\bsnm{Bellemare}, \binits{M.}}:
\bctitle{Safe and efficient off-policy reinforcement learning}.
In: \bbtitle{Advances in Neural Information Processing Systems},
pp. \bfpage{1054}--\blpage{1062}
(\byear{2016})
\end{bchapter}
\endbibitem

\bibitem{damashek1995gauging}
\begin{barticle}
\bauthor{\bsnm{Damashek}, \binits{M.}}:
\batitle{Gauging similarity with n-grams: Language-independent categorization
  of text}.
\bjtitle{Science}
\bvolume{267}(\bissue{5199}),
\bfpage{843}--\blpage{848}
(\byear{1995})
\end{barticle}
\endbibitem

\bibitem{zhang2018efficient}
\begin{botherref}
\oauthor{\bsnm{Zhang}, \binits{Y.}},
\oauthor{\bsnm{Vuong}, \binits{Q.H.}},
\oauthor{\bsnm{Song}, \binits{K.}},
\oauthor{\bsnm{Gong}, \binits{X.-Y.}},
\oauthor{\bsnm{Ross}, \binits{K.W.}}:
Efficient entropy for policy gradient with multidimensional action space.
arXiv preprint arXiv:1806.00589
(2018)
\end{botherref}
\endbibitem

\bibitem{loshchilov2018decoupled}
\begin{bchapter}
\bauthor{\bsnm{Loshchilov}, \binits{I.}},
\bauthor{\bsnm{Hutter}, \binits{F.}}:
\bctitle{Decoupled weight decay regularization}.
In: \bbtitle{International Conference on Learning Representations}
(\byear{2018})
\end{bchapter}
\endbibitem

\bibitem{mandlik2020mapping}
\begin{botherref}
\oauthor{\bsnm{Mandlík}, \binits{{\v{S}}.}}:
Mapping the internet — modelling entity interactions in complex heterogeneous
  networks.
Master's thesis,
Czech technical university in Prague
(2020)
\end{botherref}
\endbibitem

\bibitem{hornik1991approximation}
\begin{barticle}
\bauthor{\bsnm{Hornik}, \binits{K.}}:
\batitle{Approximation capabilities of multilayer feedforward networks}.
\bjtitle{Neural networks}
\bvolume{4}(\bissue{2}),
\bfpage{251}--\blpage{257}
(\byear{1991})
\end{barticle}
\endbibitem

\bibitem{leshno1993multilayer}
\begin{barticle}
\bauthor{\bsnm{Leshno}, \binits{M.}},
\bauthor{\bsnm{Lin}, \binits{V.Y.}},
\bauthor{\bsnm{Pinkus}, \binits{A.}},
\bauthor{\bsnm{Schocken}, \binits{S.}}:
\batitle{Multilayer feedforward networks with a nonpolynomial activation
  function can approximate any function}.
\bjtitle{Neural networks}
\bvolume{6}(\bissue{6}),
\bfpage{861}--\blpage{867}
(\byear{1993})
\end{barticle}
\endbibitem

\bibitem{maliah2018mdp}
\begin{bchapter}
\bauthor{\bsnm{Maliah}, \binits{S.}},
\bauthor{\bsnm{Shani}, \binits{G.}}:
\bctitle{Mdp-based cost sensitive classification using decision trees.}
In: \bbtitle{AAAI Conference on Artificial Intelligence},
pp. \bfpage{3746}--\blpage{3753}
(\byear{2018})
\end{bchapter}
\endbibitem

\bibitem{cturelational}
\begin{botherref}
\oauthor{\bsnm{Motl}, \binits{J.}},
\oauthor{\bsnm{Schulte}, \binits{O.}}:
The {CTU} {P}rague relational learning repository.
arXiv preprint arXiv:1511.03086
(2015)
\end{botherref}
\endbibitem

\bibitem{ba2016layer}
\begin{botherref}
\oauthor{\bsnm{Ba}, \binits{J.L.}},
\oauthor{\bsnm{Kiros}, \binits{J.R.}},
\oauthor{\bsnm{Hinton}, \binits{G.E.}}:
Layer normalization.
arXiv preprint arXiv:1607.06450
(2016)
\end{botherref}
\endbibitem

\end{thebibliography}

\begin{appendices}
\onecolumn
\setcounter{page}{1}
\def\appendixname{Part}

{
\centering {\Huge Supplementary Material}

\vspace{2mm}
{\Large Classification with Costly Features in Hierarchical Deep Sets}

}

\section{A2C Algorithm} \label{apx:a2c}

The A2C algorithm was explained in the main text, therefore we only provide the derivation of eq.~\eqref{eq:h2} below. For readability, we omit $\theta$ in $\nabla_\theta$ and $\pi_\theta$:
\begin{proof}[Derivation of eq.~\eqref{eq:h2}]
\begin{align*}
  \nabla H_\pi(s) &= -\nabla \sum_{a} \pi(a \mid  s) \cdot \log \pi(a \mid  s) \\
        &= -\sum_{a} \nabla \pi(a \mid  s) \cdot \log \pi(a \mid  s) - \sum_{a} \pi(a \mid  s) \cdot \nabla \log \pi(a \mid  s)
\end{align*}
Using the fact $x \cdot \nabla \log x = \nabla x$, we show that the second term is zero:
$$ \sum_{a} \pi(a \mid  s) \cdot \nabla \log \pi(a \mid  s) = \sum_{a} \nabla \pi(a \mid  s) = \nabla \sum_{a} \pi(a \mid  s) = \nabla 1 = 0$$
Let's continue with the remaining term and use the fact $\nabla x = x \cdot \nabla \log x$ again:
\begin{align*}
  \nabla H_\pi(s) &= -\sum_{a} \nabla \pi(a \mid  s) \cdot \log \pi(a \mid  s) \\
        &= -\sum_{a} \pi(a \mid  s) \cdot \nabla \log \pi(a \mid  s) \cdot \log \pi(a \mid  s) \\
        &= -\expect_{a \sim \pi(s)} \Big[ \nabla \log \pi(a \mid  s) \cdot \log \pi(a \mid  s) \Big] 
\end{align*}
\end{proof}
\clearpage
\section{Dataset Details} \label{apx:schemas}
The \emph{hepatitis}, \emph{mutagenesis}, \emph{sap} and \emph{stats} datasets were retrieved from \citet{cturelational} and processed into trees by fixing the root and unfolding the graph into a defined depth, if required. The \emph{threatcrowd} dataset was sourced from the Threatcrowd service, with permission to share. The \emph{ingredients} dataset was retrieved from Kaggle\footnote{\url{https://kaggle.com/alisapugacheva/recipes-data}}.

Float values in all datasets are normalized. Strings were processed with the tri-gram histogram method \citep{damashek1995gauging}, with modulo~13 index hashing. The datasets were split into training, validation, and testing sets as shown in Table~\ref{tab:dataset_params}. The dataset schemas, including the feature costs, are in Figure~\ref{fig:dataset_structure}.

\newcommand{\dtreefloat}[3]{
  \begin{minipage}{#1\linewidth}
    \input{#2}
    \centering
    #3
  \end{minipage}
}

\begin{table*}[ht]
    \centering
    \small
    \caption{Dataset statistics.}
    \begin{tabular}{l|rrrr|r}
      \toprule
      dataset         & samples & \#train & \#validation & \#test & class distribution \\
      \midrule          
      synthetic       & 12     & 4 & 4 & 4               & 0.5 / 0.5 \\                  
      threatcrowd     & 1 171  & 771 & 200 & 200         & 0.27 / 0.73                             \\
      hepatitis       & 500    & 300 & 100 & 100         & 0.41 / 0.59                             \\
      mutagenesis     & 188    & 100 &  44 &  44         & 0.34 / 0.66                             \\
      ingredients     & 39 774 & 29774 & 5000 & 5000     & 0.01$\sim$0.20                             \\
      sap             & 35 602 & 15602 & 10000 & 10000   & 0.5 / 0.5                             \\
      stats           & 8 318  & 4318 & 2000 & 2000      &  0.49 / 0.38 / 0.12                             \\
    \end{tabular}
    \label{tab:dataset_params}
\end{table*}

\begin{figure}[ht]
  \centering
  \fontsize{8}{8}
  
  \begin{minipage}{0.3\linewidth}
    \input{web_100k.txt}
    \centering
    (a) threatcrowd
  \end{minipage}

  \begin{minipage}{0.3\linewidth}
    \input{stats_full.txt}
    \centering
    (b) stats
  \end{minipage}

  \begin{minipage}{0.3\linewidth}
    \input{hepa.txt}
    \centering
    (c) hepatitis
  \end{minipage}

  \begin{minipage}{0.4\linewidth}
    \input{muta.txt}
    \centering
    (g) mutagenesis
  \end{minipage}

  \begin{minipage}{0.4\linewidth}
    \input{sap_balanced.txt}
    \centering
    (f) sap
  \end{minipage}

  \begin{minipage}{0.4\linewidth}
    \input{toy_b.txt}
    \centering
    (e) synthetic
  \end{minipage}

  \begin{minipage}{0.4\linewidth}
    \input{recipes.txt}
    \centering
    (d) ingredients
  \end{minipage}

  \caption{Datasets schemas used in this work. The trees show the feature names, their types, and their cost in parentheses. Features with a \emph{set} type contain an arbitrary number of same-typed items.}
  \label{fig:dataset_structure}
\end{figure}

\clearpage
\section{Implementation and Hyperparameters} \label{apx:implementation}
The hyperparameters for each method are summarized in Table~\ref{tab:hyperparams}. We searched for an optimal set of hyperparameters: batch size in $\{128, 256\}$, learning rate in $[3\times 10^{-4}, 1\times 10^{-3}, 3\times 10^{-3}, 1\times 10^{-2}]$, embedding size in $\{64, 128\}$, weight decay in range $[1\times 10^{-4}, 3.0]$ and $\alpha_h$ in $\{0.0025, 0.025, 0.5, 0.1\}$. We report that properly tuned weight decay and $\alpha_h$ is crucial for good performance.

The model's architecture and parameters are initialized according to the provided dataset schema, and parameters $\vartheta_{\mathcal B}, \varphi_{\mathcal B}$ are created for each bag $\mathcal B$. LeakyReLU is used as the activation function. The aggregation function is \emph{mean} with layer normalization \citep{ba2016layer}. The policy entropy controlling weight ($\alpha_h$) decays with $\frac{1}{T}$ schedule every 10 epochs. We use AdamW optimizer \citep{loshchilov2018decoupled} with weight decay regularization. The learning rate of the main training in annealed exponentially by a factor of $0.5$ every 10 epochs. The gradients are clipped to a norm of $1.0$. For each dataset, we select the best performing iteration based on its validation reward. 

\subsection{Pretraining}
Before main training, the classifier $\varrho$ and the value output $V$ are initialized by pretraining. The classifier is pretrained with randomly generated partial samples from the dataset, with cross-entropy loss and the initial learning rate. To cover the whole state space, the partial samples are constructed as follows: A probability $p \sim \mathcal U_{[0,1]}$ is chosen and, starting from the root of the sample, features are included with probability $p$. All objects in sets are recursively processed in the same manner and the same probability $p$. The pretraining proceeds for a whole epoch (with the same batch size as in the main training) and subsequently the validation loss is estimated using the complete samples every $\frac{1}{10}$ of an epoch. Early stopping is used to terminate the pretraining.

\begin{sidewaystable}[ht] \footnotesize
    \centering
    \caption{Hyper-parameters. If a hyper-parameter is not specified, it is the same as in \emph{HMIL-CWCF}.}
    \begin{tabular}{l|rrrrrrrr}
      \toprule
      parameter & \textit{shared} & synthetic & hepatitis & mutagenesis & ingredients & sap & stats & threatcrowd \\
      \midrule
      \textbf{HMIL-CwCF} \\
      steps in epoch & 1000 & 100 &  &  &  &  &  &  \\
      train time (epochs) & 200 &  &  &  & 300 &  &  &  \\
      batch size & 256 &  &  &  &  &  &  &  \\
      embedding size ($f_{\theta_{\mathcal B}}$) & 128 &  &  &  &  &  &  &  \\
      learning rate (initial) & 1.0e-3 &  &  &  &  &  &  &  \\
      learning rate (final) & 1.0e-3 / 30 &  &  &  &  &  &  &  \\
      learning rate decay factor & \multicolumn{2}{r}{0.5 (every 10 epochs)} &  &  &  &  &  &  \\
      gradient max norm & 1.0 &  &  &  &  &  &  &  \\
      $\gamma$ - discount factor & 0.99 &  &  &  &  &  &  & \\
      $\alpha_v$ & 0.5 &  &  &  &  &  &  &  \\
      $\alpha_h$ (initial) &  & 0.025 & 0.1 & 0.025 & 0.05 & 0.1 & 0.05 & 0.05 \\
      $\alpha_h$ (final) &  & 0.00025 & 0.005 & 0.00025 & 0.0025 & 0.005 & 0.0025 & 0.0025 \\
      weight decay &  & 1e-4 & 1e-4 & 1e-4 & 0.3 & 1e-4 & 1e-4 & 3.0 \\
      \midrule      
      \textbf{HMIL} \\
      steps in epoch & 100 &  &  &  &  &  &  &  \\
      train time (epochs) & 50 &  &  &  & 200 &  &  &  \\
      learning rate (initial) & 3.0e-3 &  &  &  &  &  &  &  \\
      learning rate (final) & 3.0e-3 / 30 &  &  &  &  &  &  &  \\
      weight decay &  & 1.0 & 1.0 & 1.0 & 1.0 & 0.1 & 3.0 & 3.0 \\
      \midrule      
      \textbf{Flat-CwCF} \\
      steps in epoch & 1000 & 200 &  &  &  &  &  &  \\
      train time (epochs) & 200 &  &  &  &  &  &  &  \\
      $\alpha_h$ (initial) & 0.05 &  & 0.1 &  &  &  &  &  \\
      $\alpha_h$ (final) & 0.0025 &  & 0.005 &  &  &  &  &  \\
      weight decay &  & 1e-4 & 1.0 & 1.0 & 1.0 & 0.1 & 1.0 & 3.0 \\
      \midrule      
      \textbf{Random} \\
      steps in epoch & 1000 & 100 &  &  &  &  &  &  \\
      train time (epochs) &  & 20 & 20 & 20 & 100 & 40 & 40 & 20 \\
      learning rate (initial) & 3.0e-3 &  &  &  &  &  &  &  \\
      learning rate (final) & 3.0e-3 / 30 &  &  &  &  &  &  &  \\
      weight decay &  & 1e-4 & 1e-4 & 1.0 & 1.0 & 0.1 & 3.0 & 3.0 \\
    \end{tabular}
    \centering
    \label{tab:hyperparams}
\end{sidewaystable}

\clearpage
\section{AUTC metric} \label{apx:auc}
In the Experiments, we use a custom Area Under Trade-off Curve (AUTC) metric to compare algorithms. Its purpose is to evaluate the performance of an algorithm across the whole range of budgets, but output only a single number. It is defined such that an algorithm that always predicts the most populous class with any budget would get~0 and an algorithm that always classifies perfectly would get~1. 

\begin{figure}[h]
  \centering
  \includegraphics[width=0.5\linewidth]{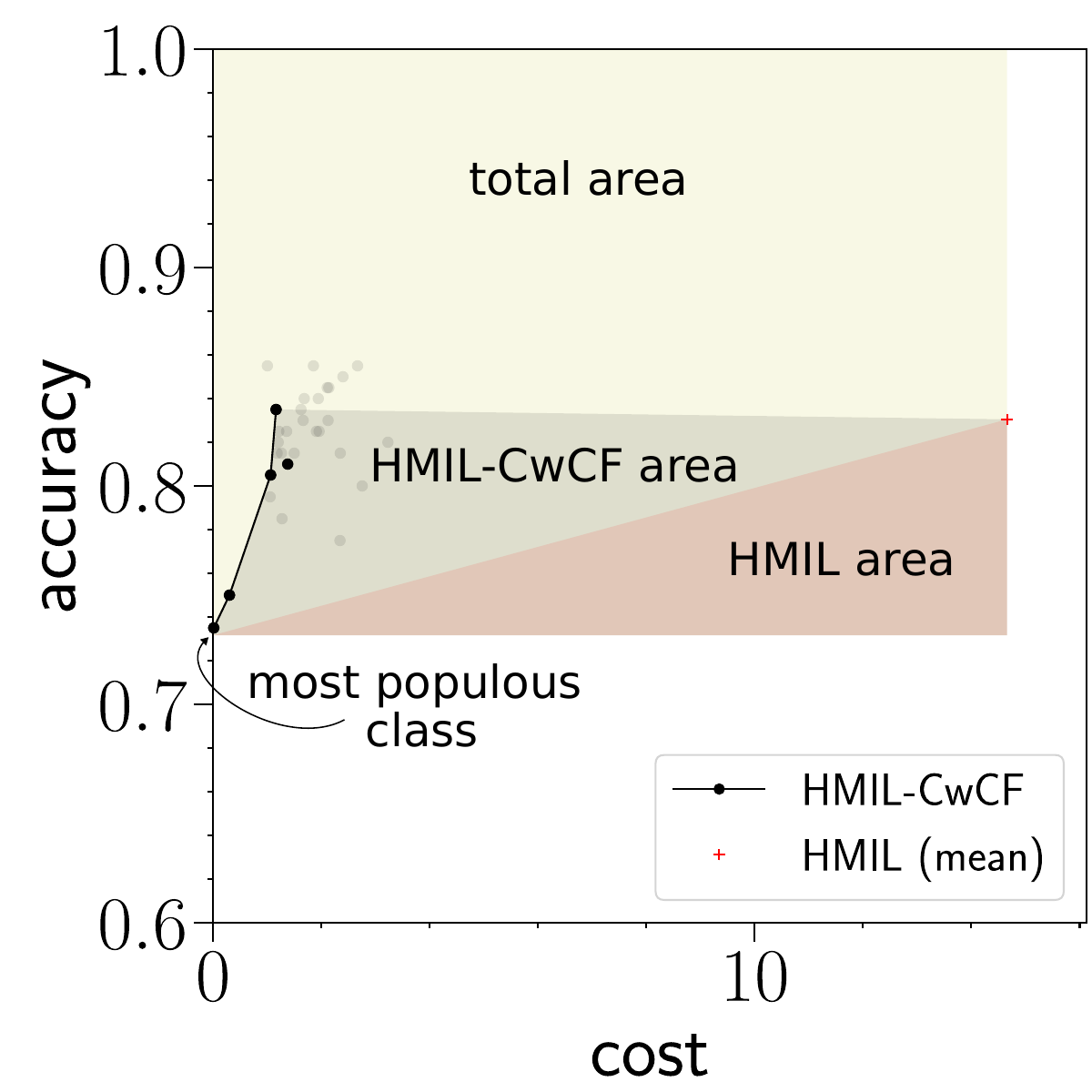}
  \caption{Visualization of the AUTC metric which is computed as the area under the curve normalized by the total area.}
  \label{fig:auc}
\end{figure}

The metric is defined as a normalized area under the visualized Pareto front, where we assume that the minimal accuracy is the prior of the most populous class and that the accuracy with all features is equal to \emph{HMIL} (see Figure~\ref{fig:auc}). The area below the prior accuracy is subtracted. The result is normalized by the area of a rectangle between $[0, prior\_accuracy]$ and $[max\_cost, 1.0]$ within the $[cost, accuracy]$ coordinates. Note that $max\_cost$ is the mean of the total cost of all features across all samples.
For \emph{HMIL}, we define the AUTC as the normalized area under the line connecting the $[0, prior\_accuracy]$ and $[max\_cost, hmil\_accuracy]$ points. The reasoning is that the user can always choose between the two.

\clearpage
\section{Sample Runs} \label{apx:sampleruns}
Below we show one selected sample from each dataset as processed by the proposed algorithm (we omit hepatitis and mutagenesis, as the samples are too large to visualize). Read left-to-right, top-to-down. At each step, the current observation is shown, the line thickness visualizes the action probabilities and the green line marks the selected action. On the left, there is a visualization of the current state value and class probabilities. The correct class is visualized with a dot, and the current prediction is in bold. All displayed models were trained with $\lambda=0.00108264$.
\begin{figure}[ht]
  0:~\includegraphics[align=c,scale=0.5]{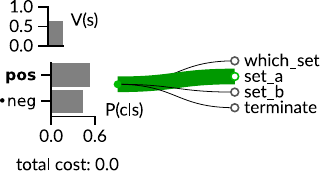}
  1:~\includegraphics[align=c,scale=0.5]{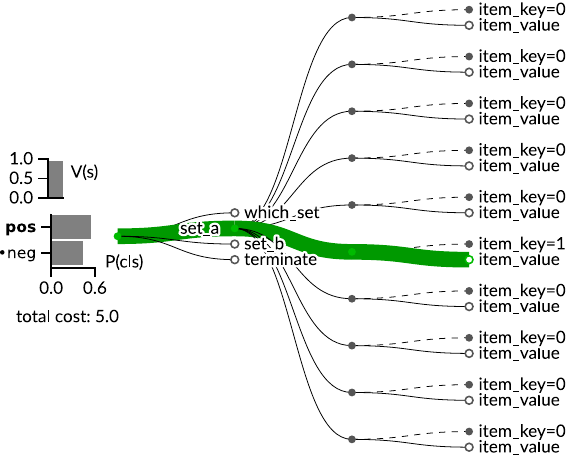}\\
  2:~\includegraphics[align=c,scale=0.5]{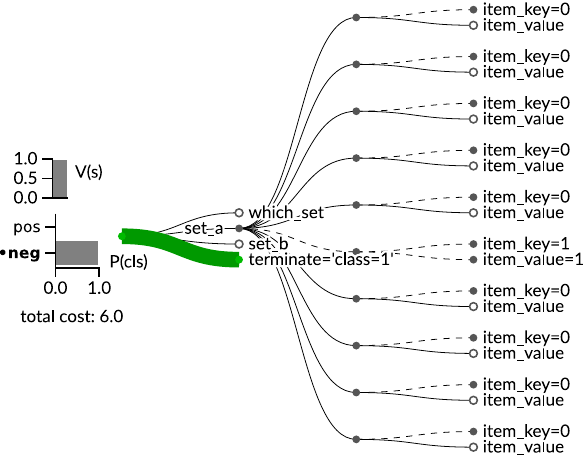}
  \caption{Sample from dataset \emph{synthetic}}
\end{figure}
\begin{figure}[ht]
  0:~\includegraphics[align=c,scale=0.5]{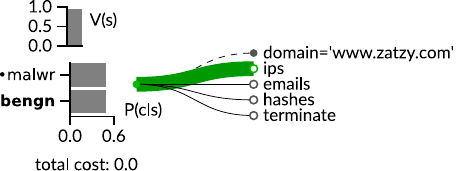}
  1:~\includegraphics[align=c,scale=0.5]{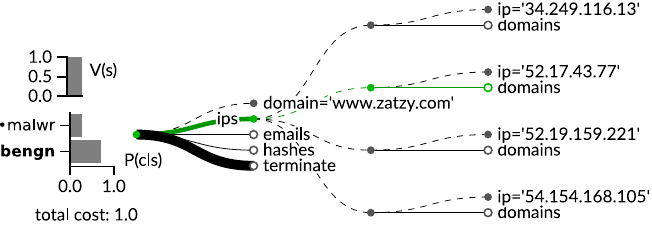}\\
  2:~\includegraphics[align=c,scale=0.5]{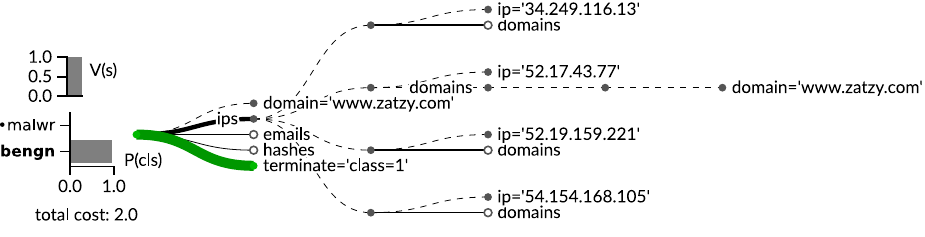}
  \caption{Sample from dataset \emph{threatcrowd}}
\end{figure}

\begin{figure}[ht]
  0:~\includegraphics[align=c,scale=0.5]{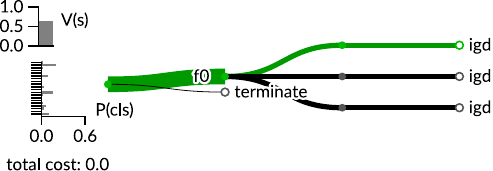}
  1:~\includegraphics[align=c,scale=0.5]{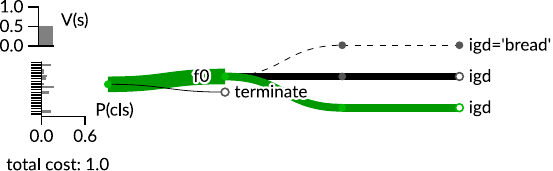}\\
  2:~\includegraphics[align=c,scale=0.5]{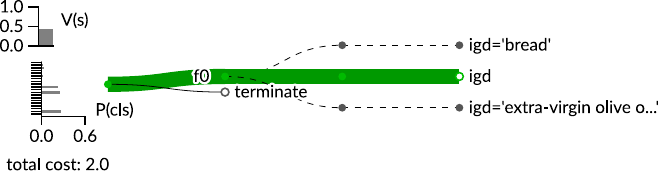}
  3:~\includegraphics[align=c,scale=0.5]{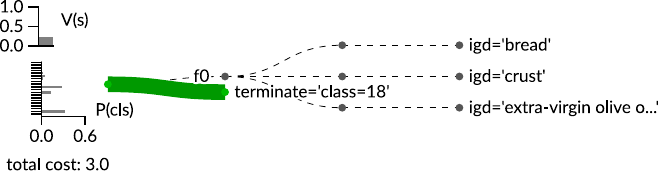}
  \caption{Sample from dataset \emph{ingredients}}
\end{figure}
\begin{figure}[ht]
  0:~\includegraphics[align=c,scale=0.5]{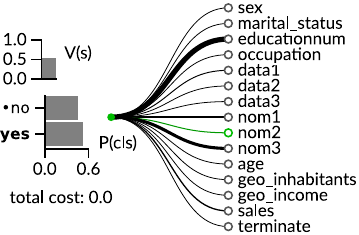}
  1:~\includegraphics[align=c,scale=0.5]{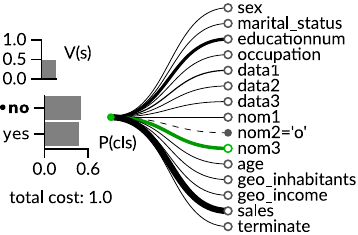}
  2:~\includegraphics[align=c,scale=0.5]{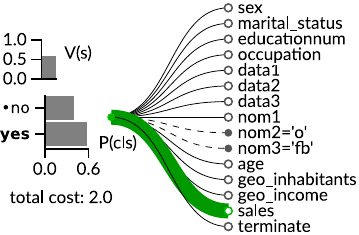}\\
  3:~\includegraphics[align=c,scale=0.5]{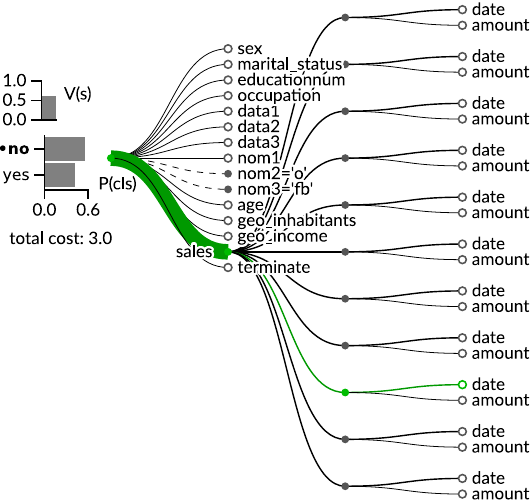}
  4:~\includegraphics[align=c,scale=0.5]{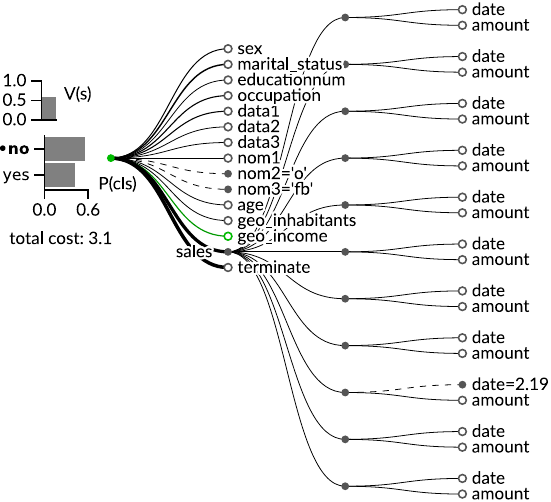}\\
  5:~\includegraphics[align=c,scale=0.5]{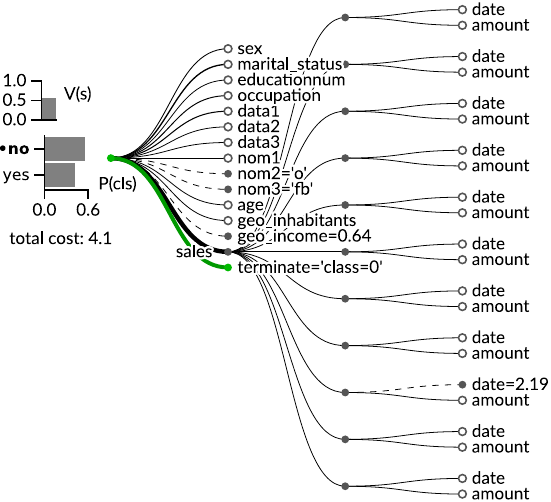}
  \caption{Sample from dataset \emph{sap}}
\end{figure}
\begin{figure}[ht]
  0:~\includegraphics[align=c,scale=0.5]{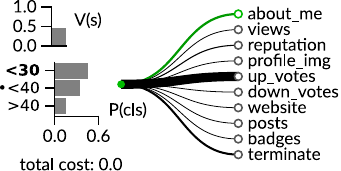}
  1:~\includegraphics[align=c,scale=0.5]{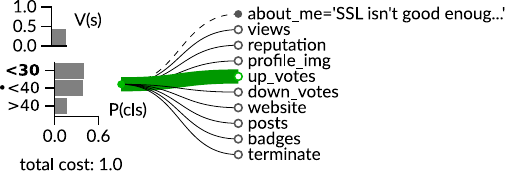}
  2:~\includegraphics[align=c,scale=0.5]{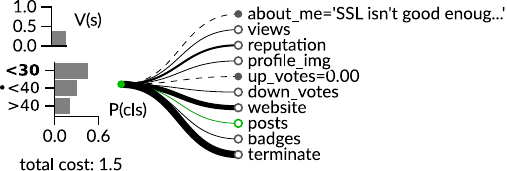}
  3:~\includegraphics[align=c,scale=0.5]{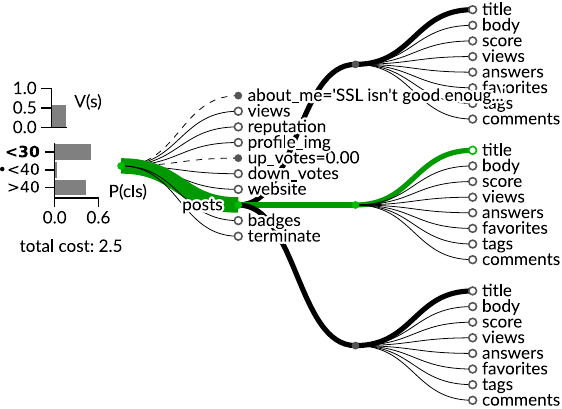}
  4:~\includegraphics[align=c,scale=0.5]{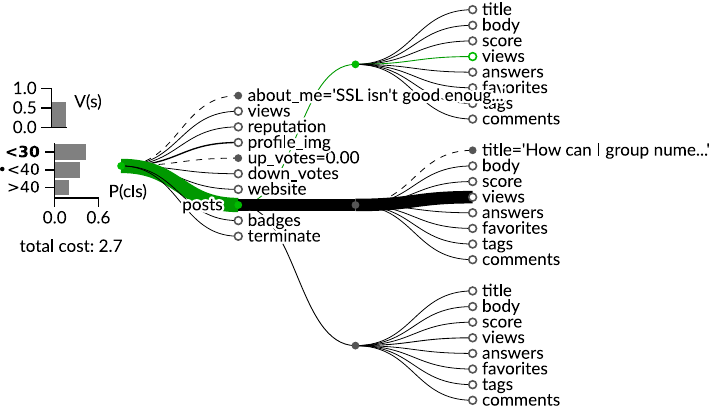}
  5:~\includegraphics[align=c,scale=0.5]{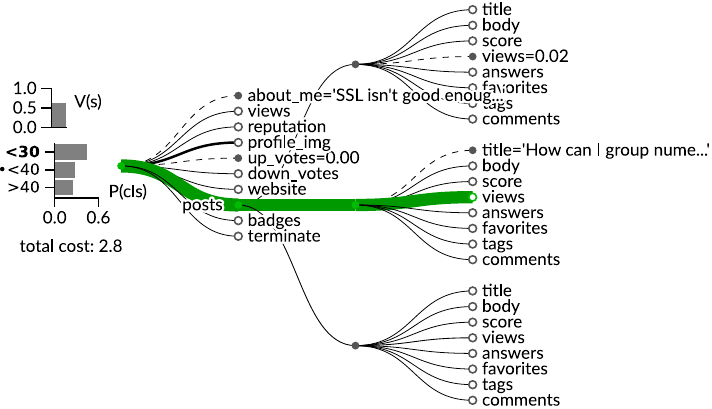}
  6:~\includegraphics[align=c,scale=0.5]{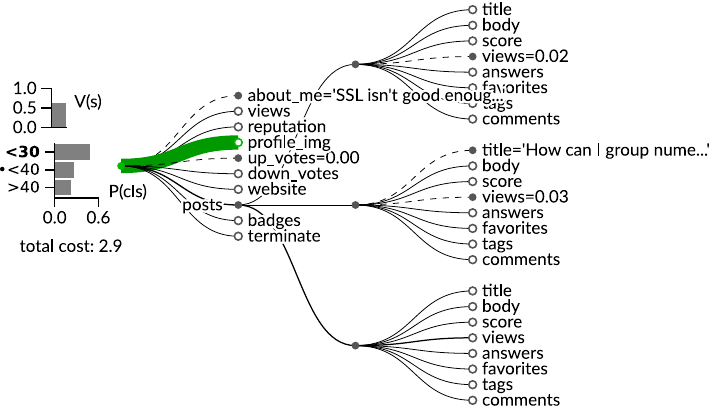}
  7:~\includegraphics[align=c,scale=0.5]{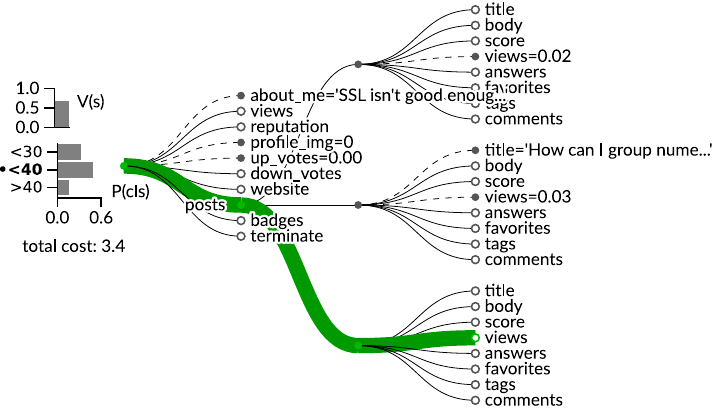}
  8:~\includegraphics[align=c,scale=0.5]{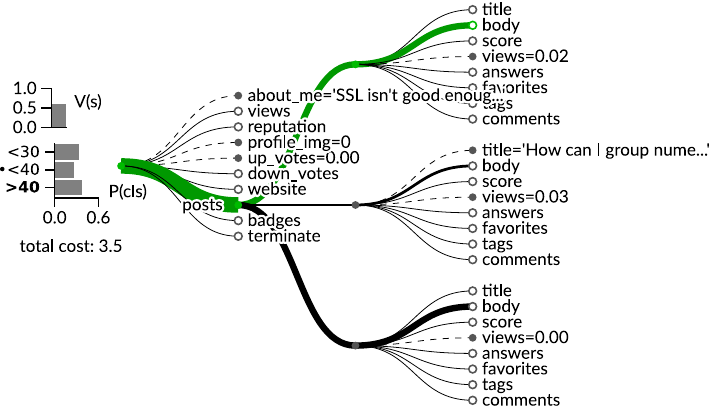}
  9:~\includegraphics[align=c,scale=0.5]{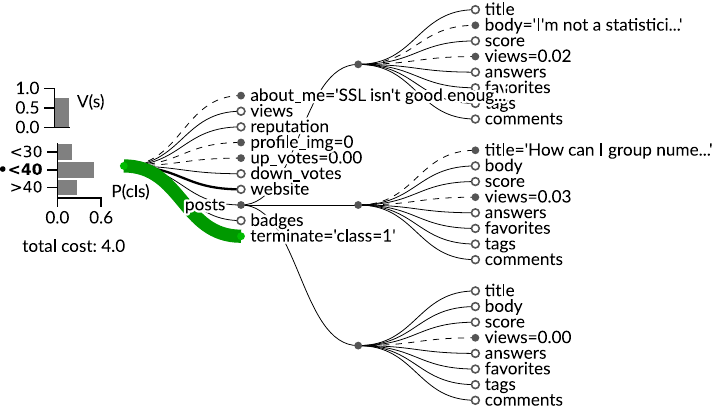}
  \caption{Sample from dataset \emph{stats}}
\end{figure}

\clearpage
\section{Convergence Graphs} \label{apx:convgraphs}
For each dataset, we present four different runs with different $\lambda$ in the following figures. Each subfigure, corresponding to a single run, is divided into three diagrams, showing the progression of average \emph{reward}, \emph{cost} and \emph{accuracy} during the training. The reader can use the plots to examine behavior of the algorithm (mainly the importance of cost vs. accuracy) with different parameters $\lambda$. Evaluation on training, validation and testing sets is shown. The best iteration (based on the validation reward) is displayed by the dashed vertical line.

\begin{figure}[ht]
  \centering

  \includegraphics[width=0.49\linewidth]{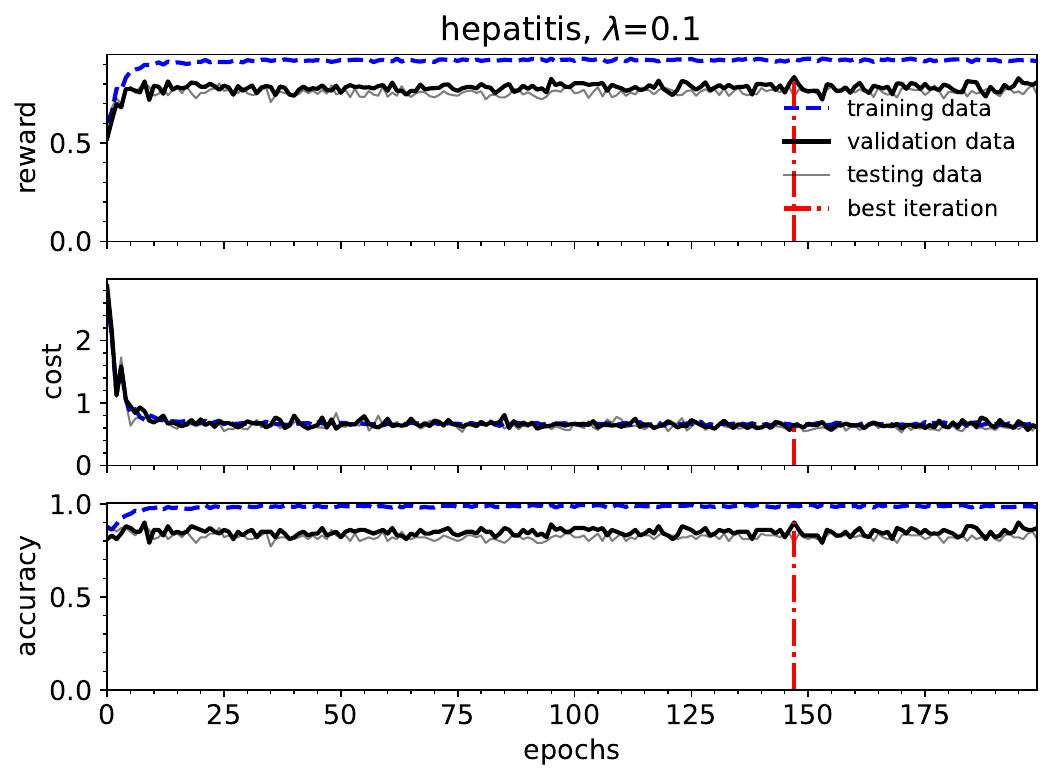}
  \includegraphics[width=0.49\linewidth]{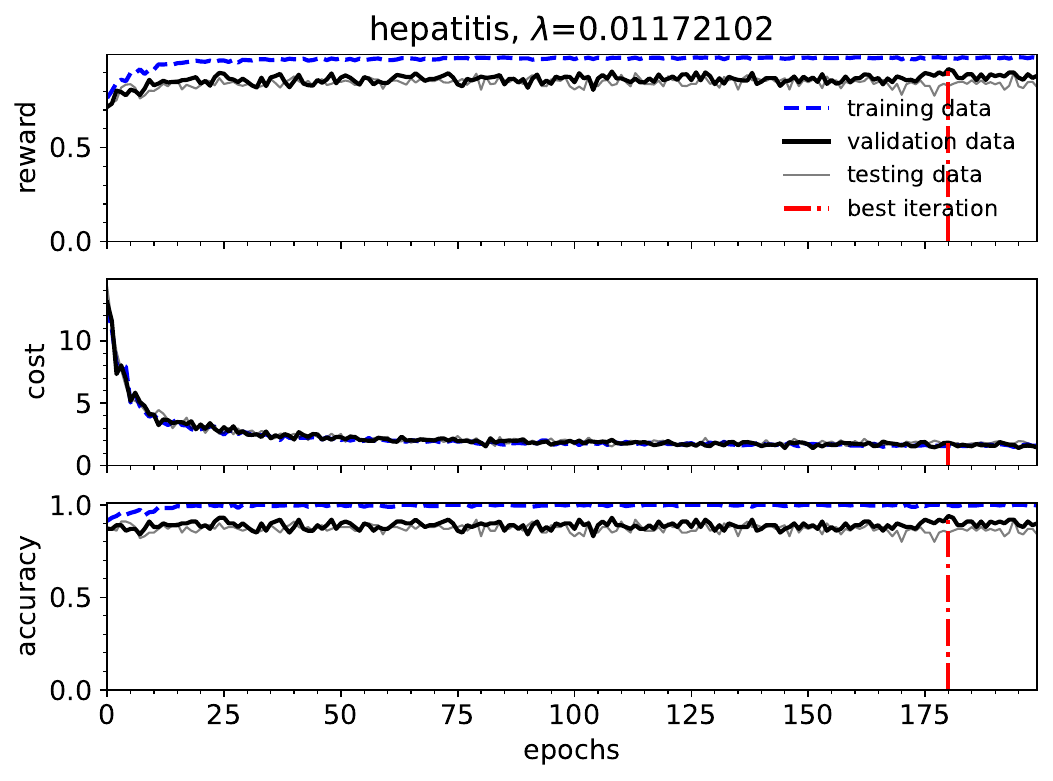}
  \includegraphics[width=0.49\linewidth]{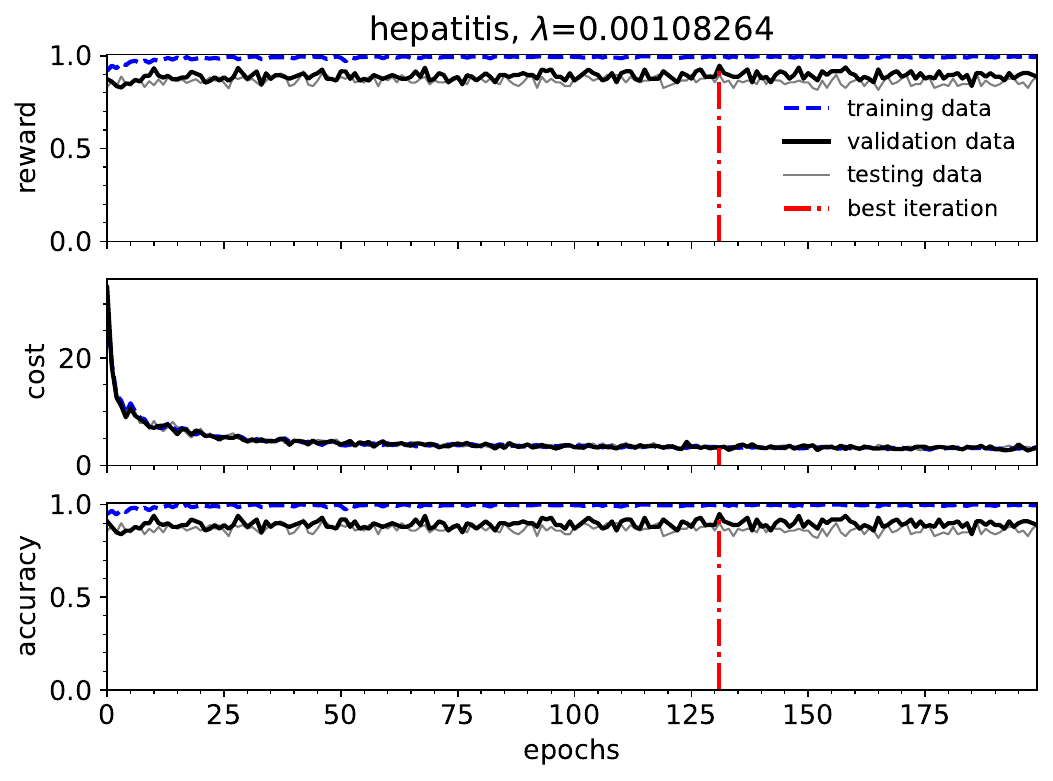}
  \includegraphics[width=0.49\linewidth]{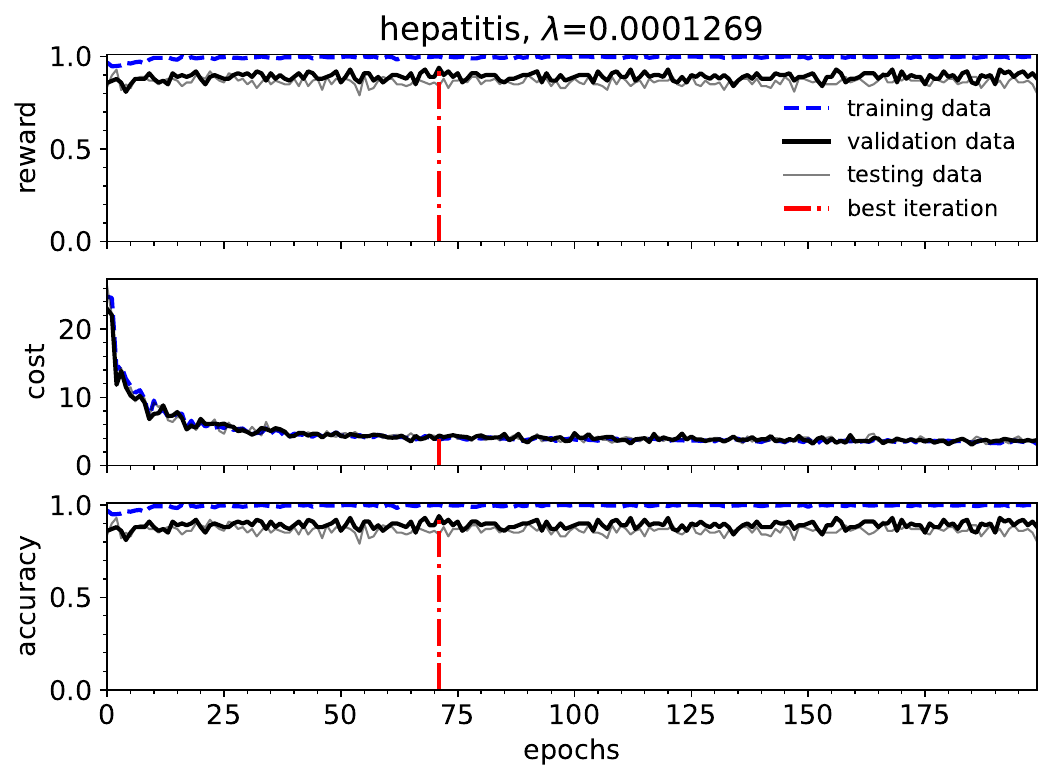}
  
  \caption{Convergence graphs for \emph{hepatitis} dataset.}  
\end{figure}

\begin{figure}[ht]
  \centering

  \includegraphics[width=0.49\linewidth]{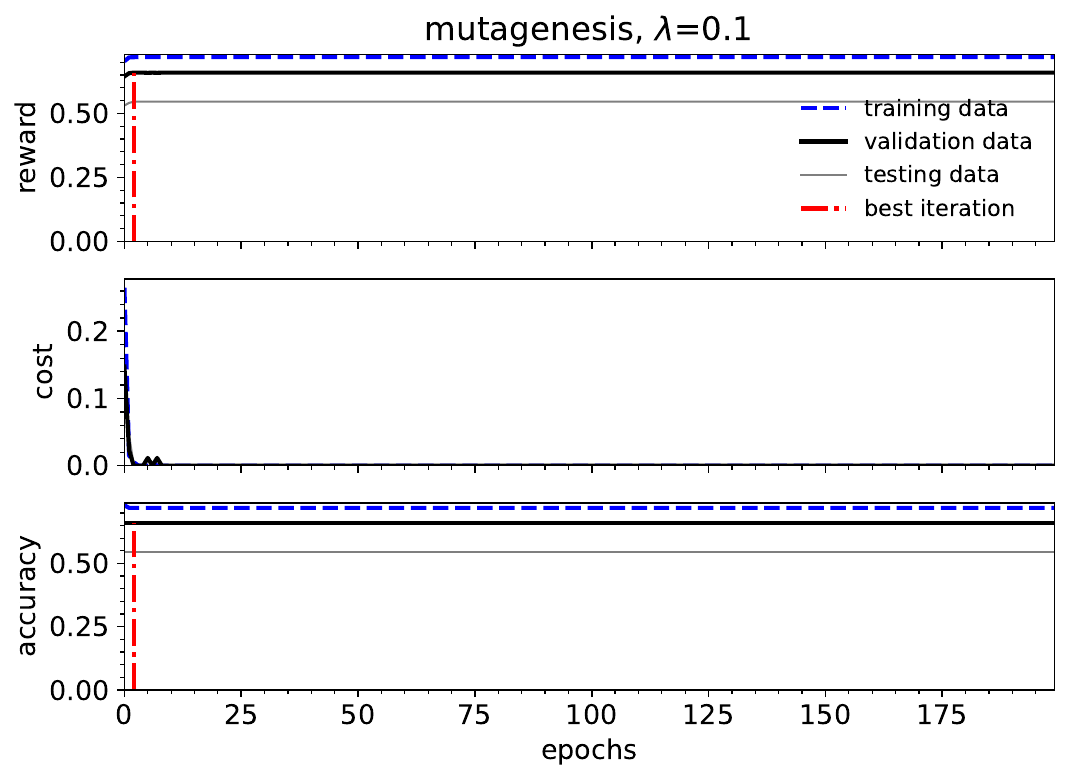}
  \includegraphics[width=0.49\linewidth]{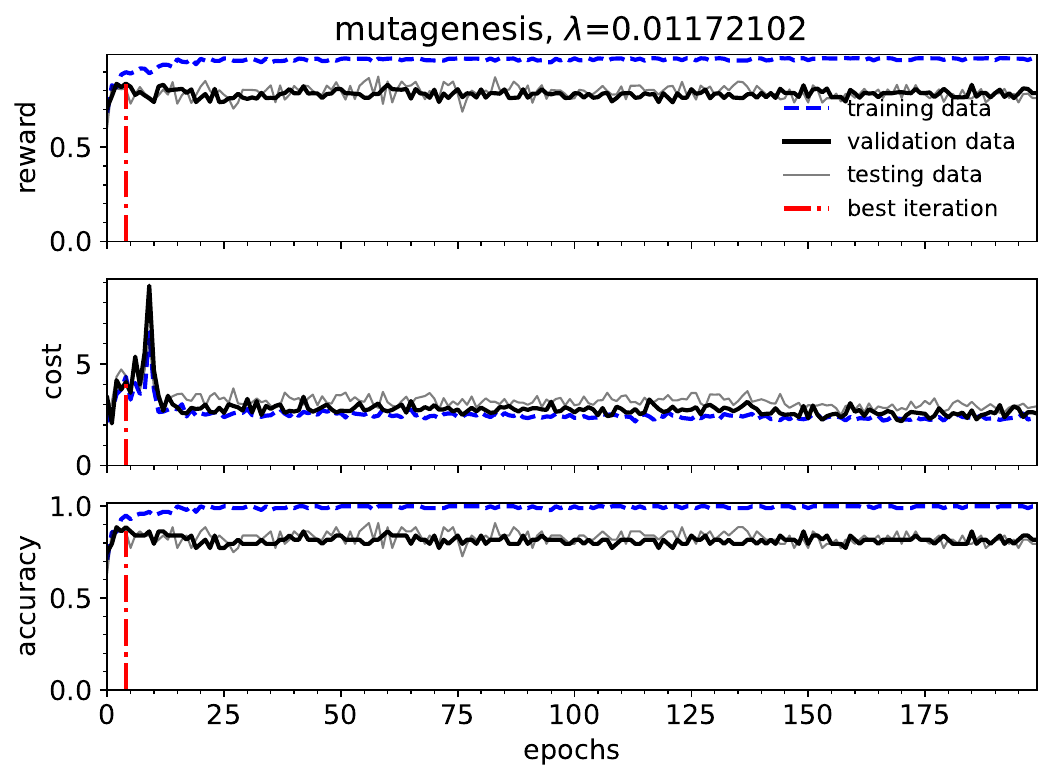}
  \includegraphics[width=0.49\linewidth]{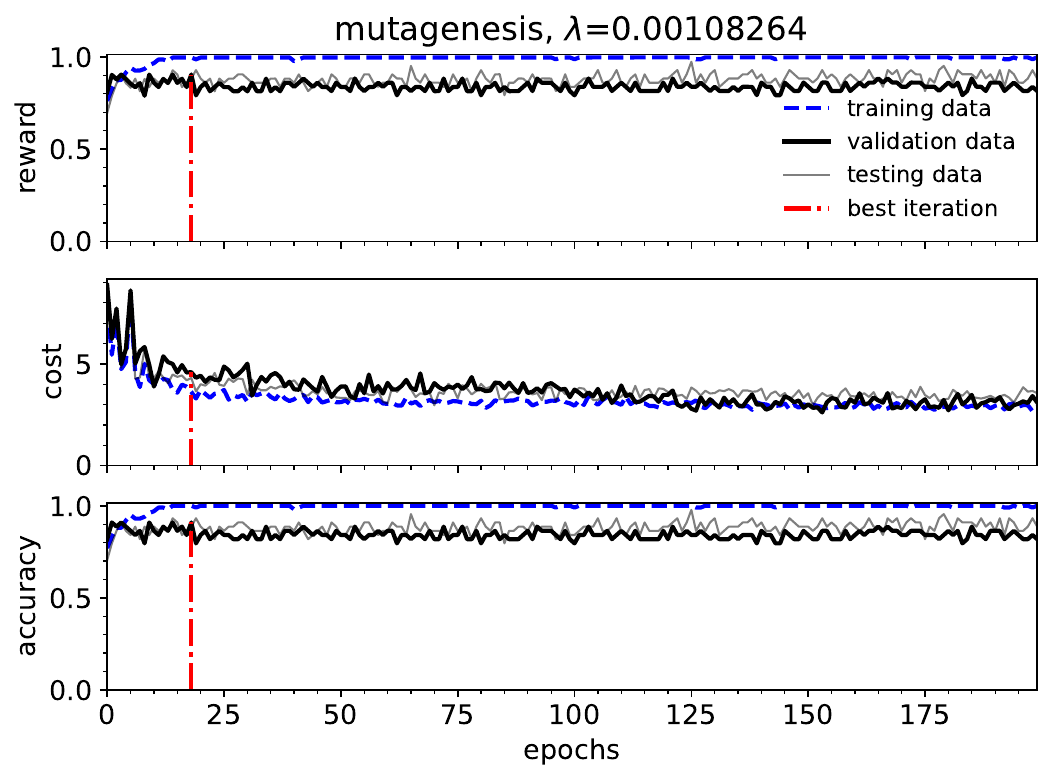}
  \includegraphics[width=0.49\linewidth]{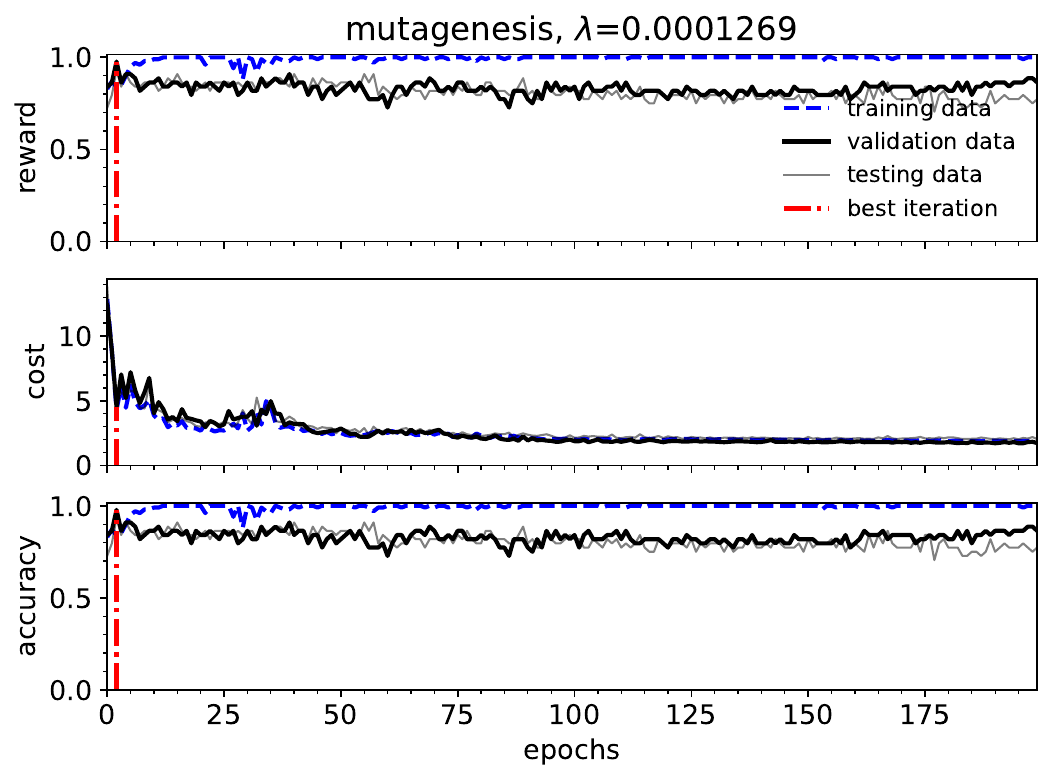}
  
  \caption{Convergence graphs for \emph{mutagenesis} dataset.}  
\end{figure}

\begin{figure}[ht]
  \centering

  \includegraphics[width=0.49\linewidth]{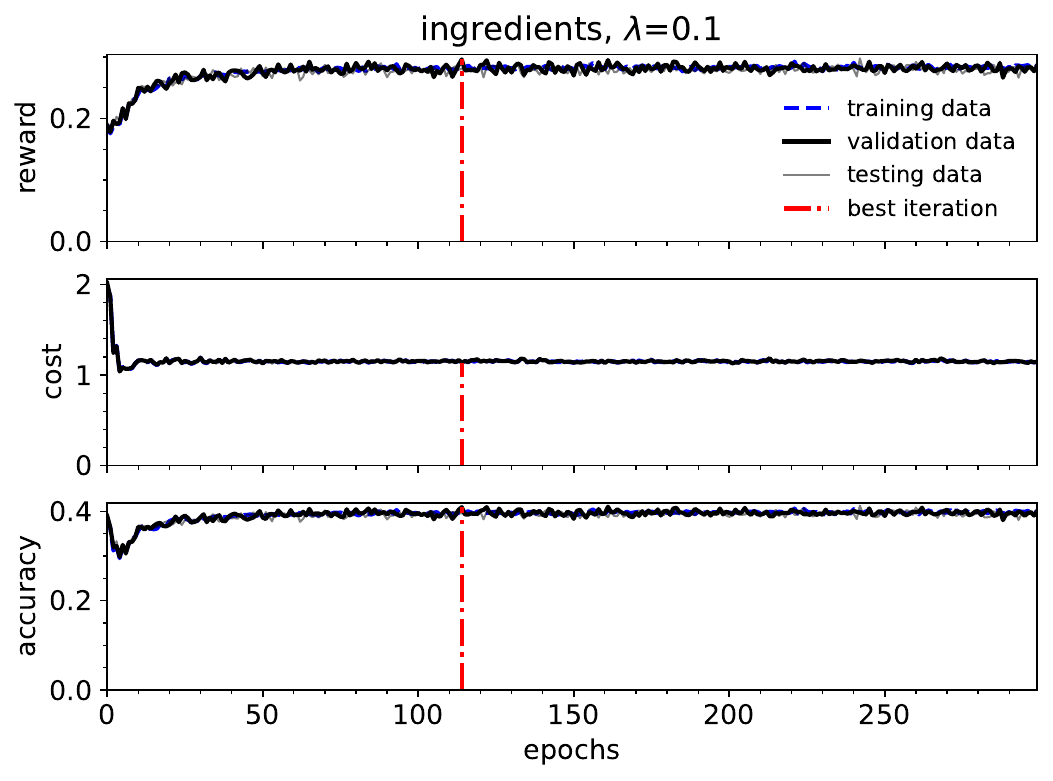}
  \includegraphics[width=0.49\linewidth]{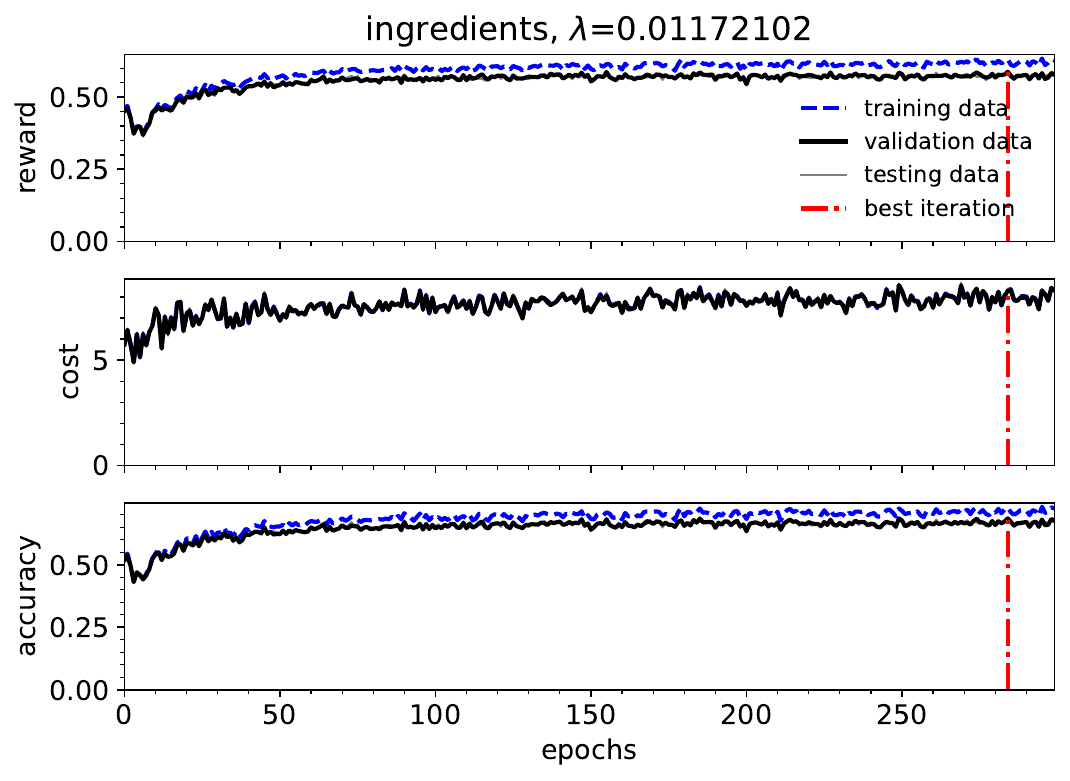}
  \includegraphics[width=0.49\linewidth]{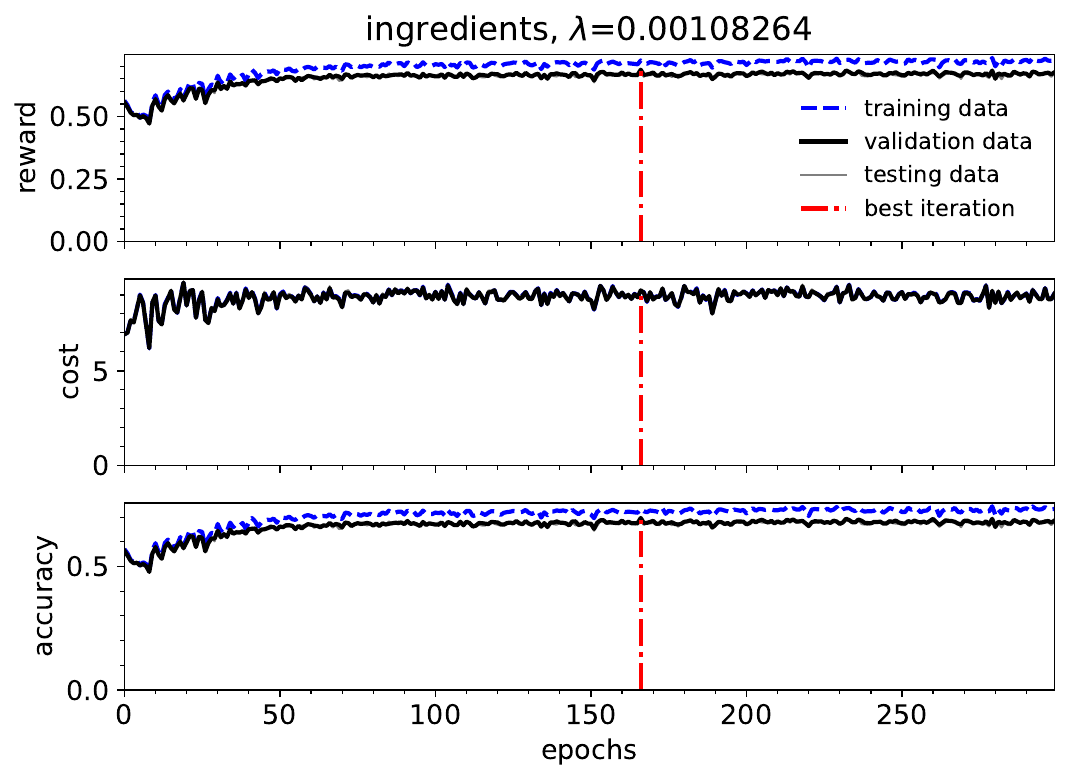}
  \includegraphics[width=0.49\linewidth]{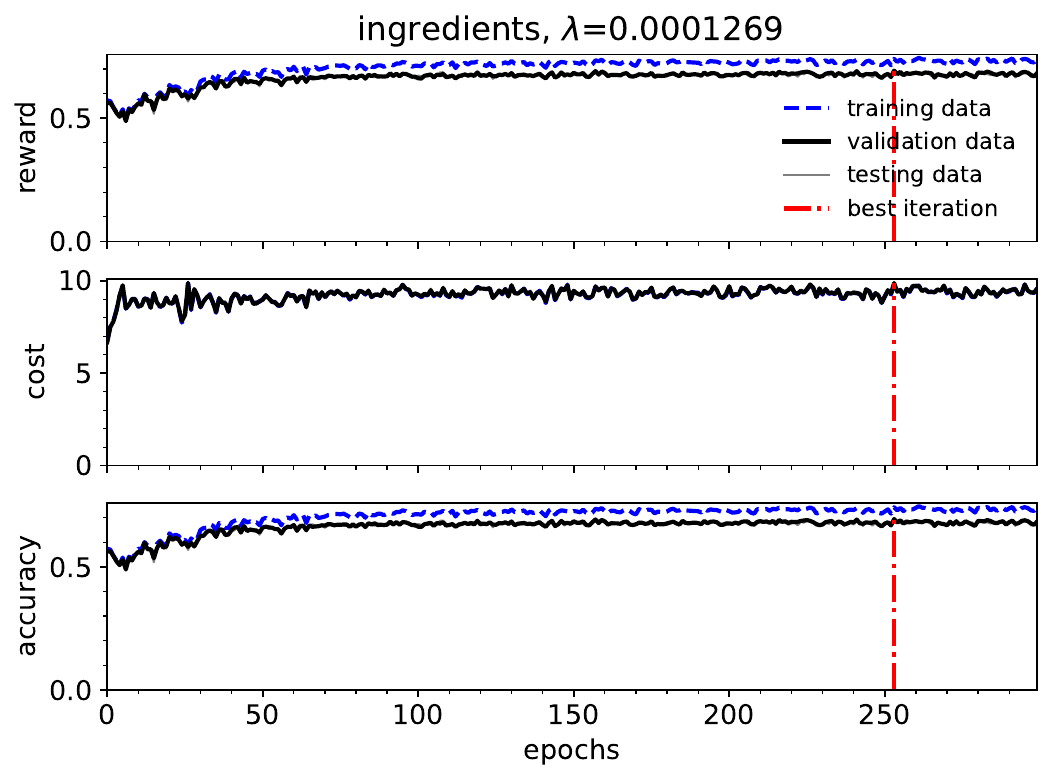}
  
  \caption{Convergence graphs for \emph{ingredients} dataset.}  
\end{figure}

\begin{figure}[ht]
  \centering

  \includegraphics[width=0.49\linewidth]{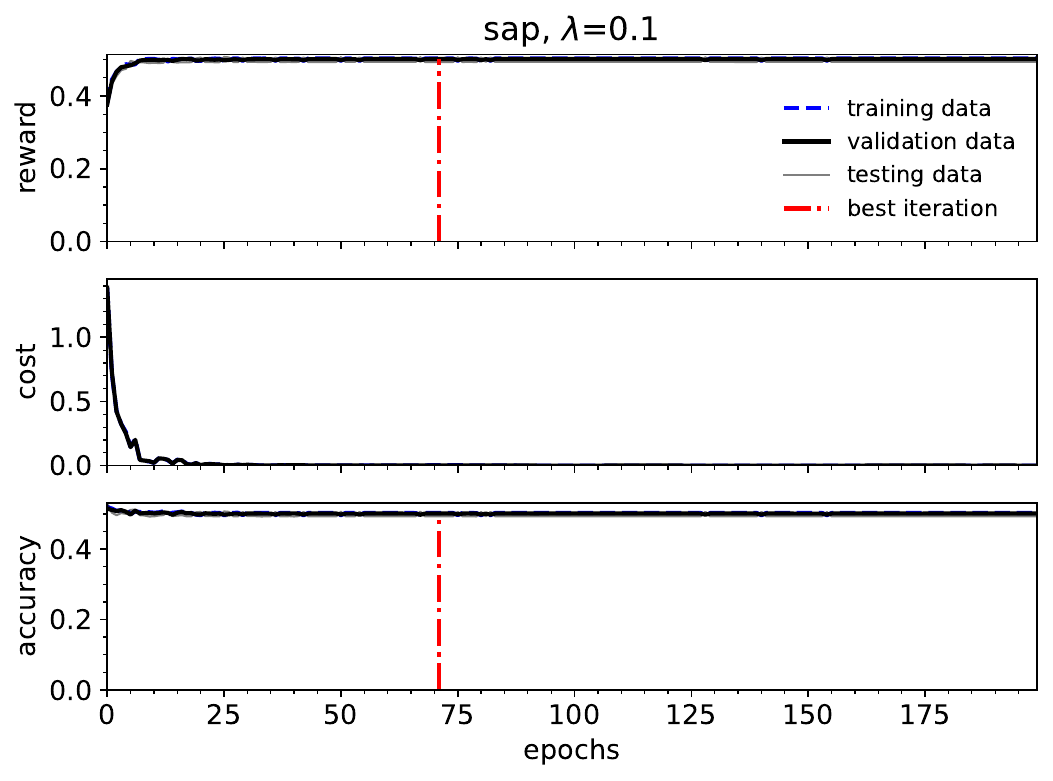}
  \includegraphics[width=0.49\linewidth]{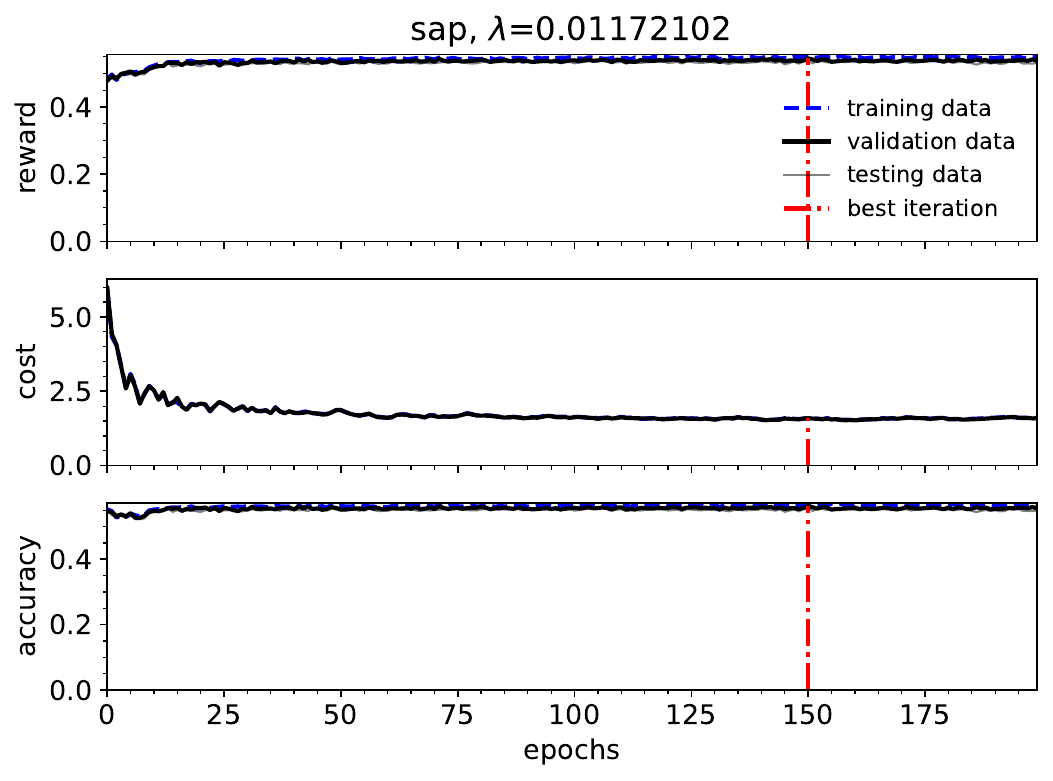}
  \includegraphics[width=0.49\linewidth]{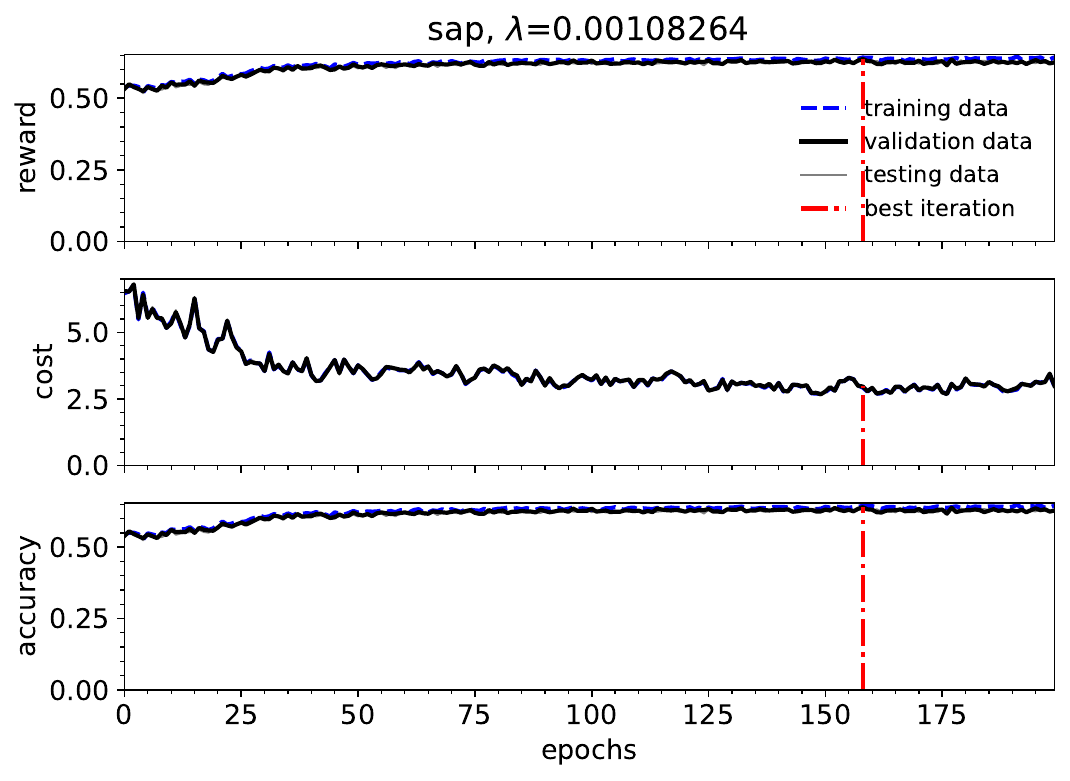}
  \includegraphics[width=0.49\linewidth]{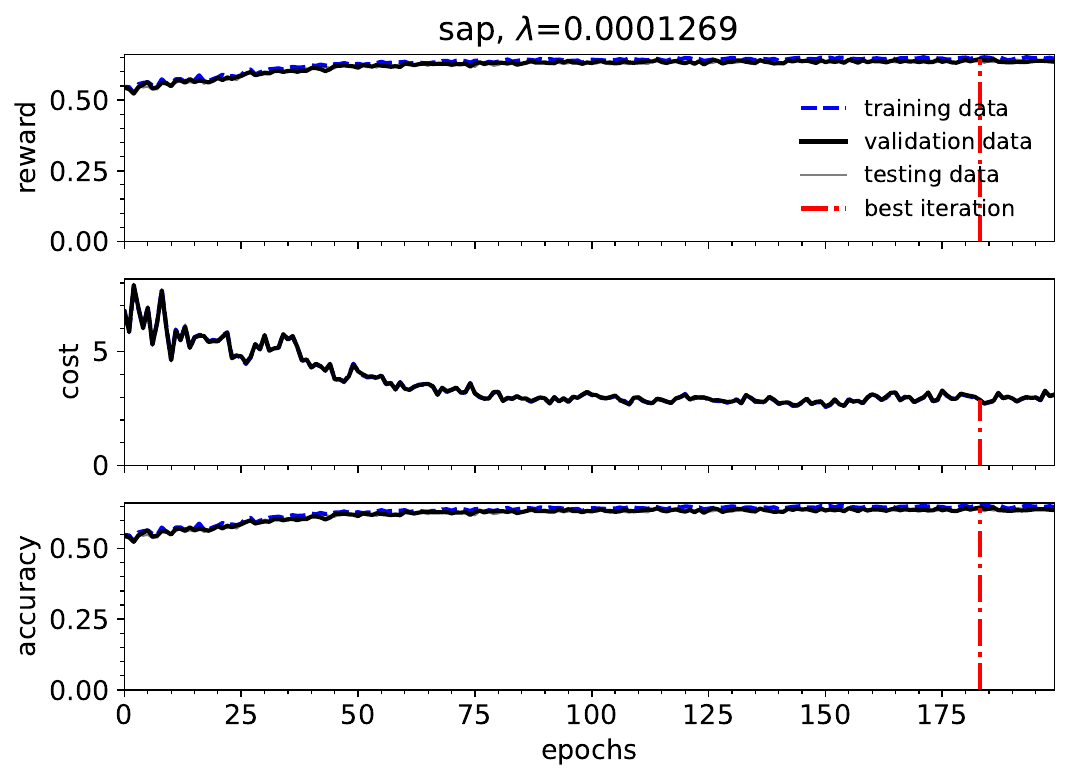}
  
  \caption{Convergence graphs for \emph{sap} dataset.}  
\end{figure}

\begin{figure}[ht]
  \centering

  \includegraphics[width=0.49\linewidth]{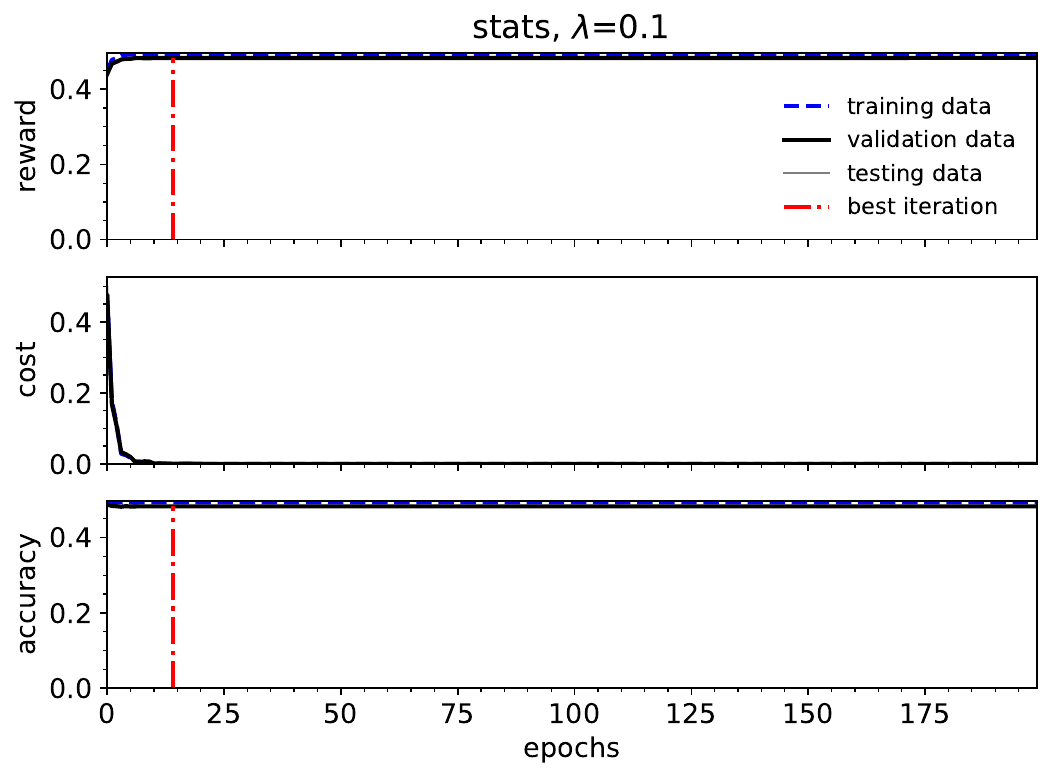}
  \includegraphics[width=0.49\linewidth]{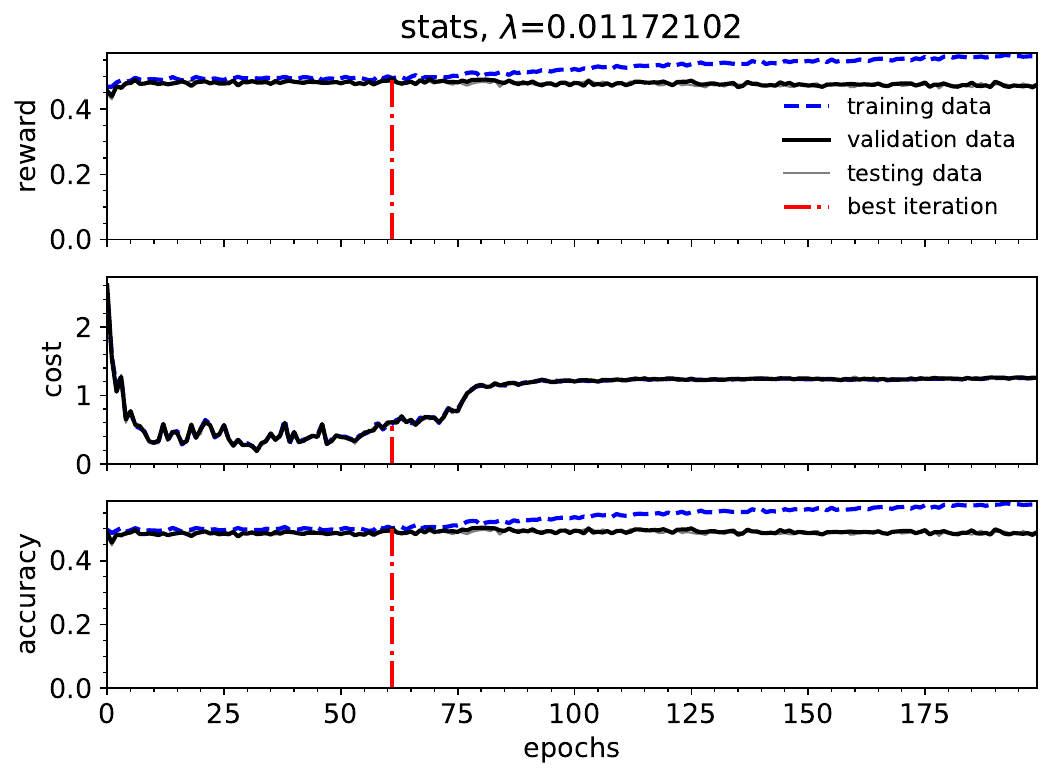}
  \includegraphics[width=0.49\linewidth]{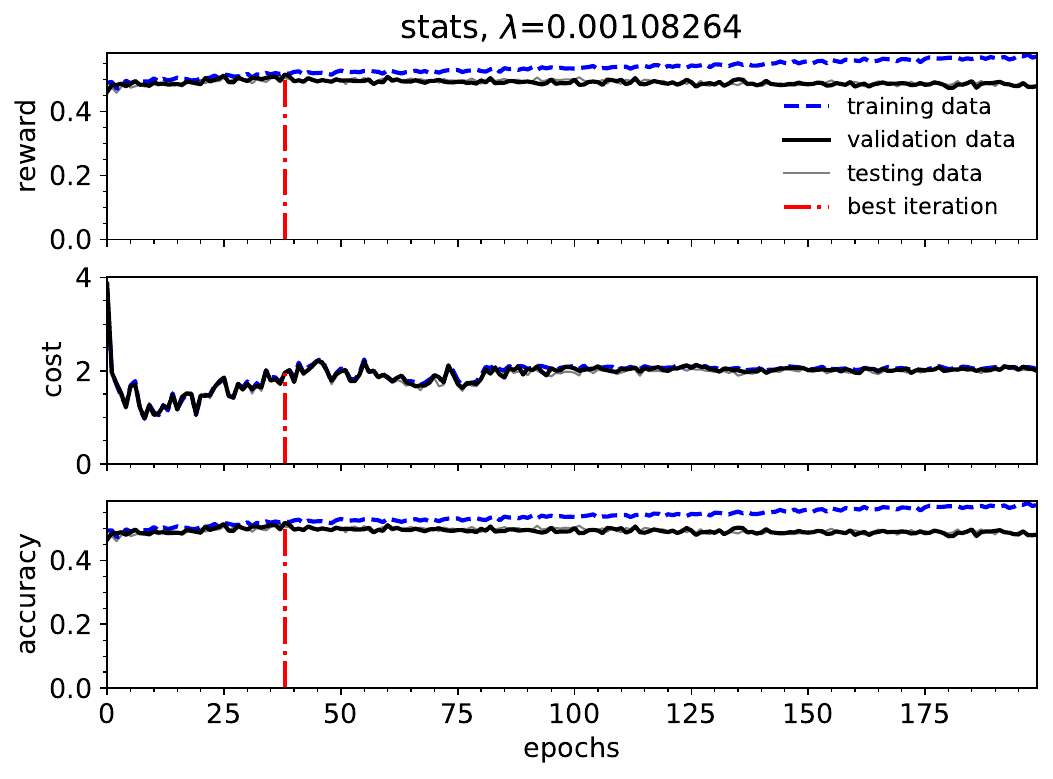}
  \includegraphics[width=0.49\linewidth]{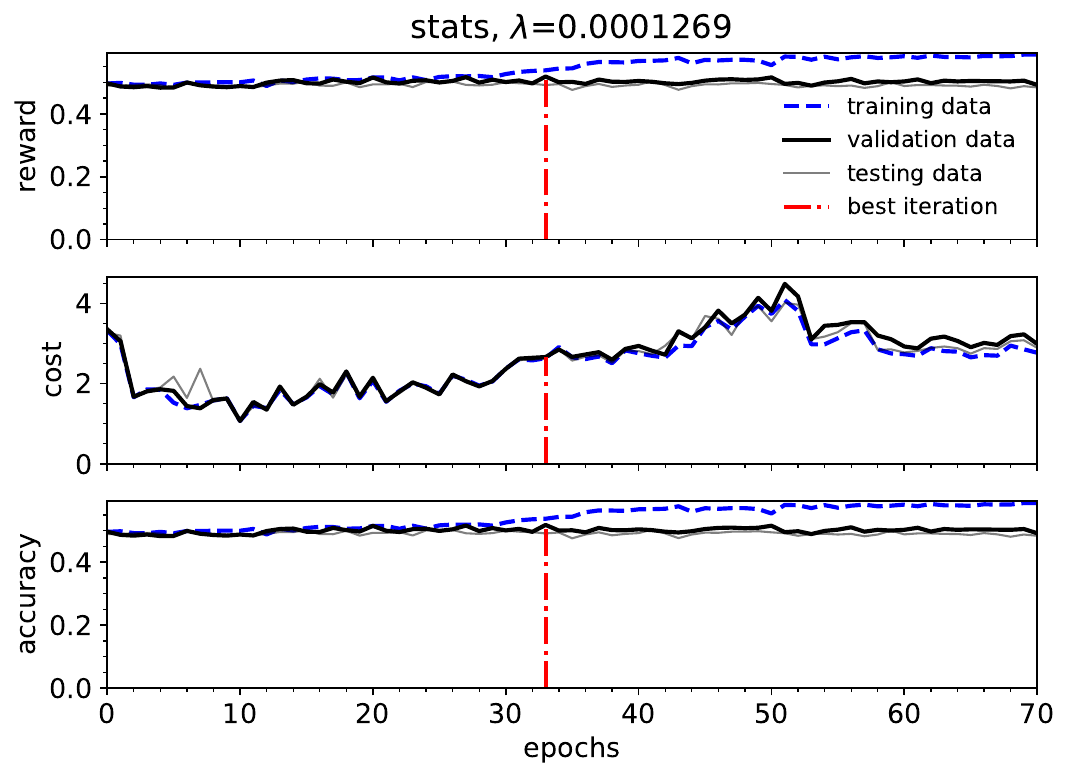}
  
  \caption{Convergence graphs for \emph{stats} dataset.}  
\end{figure}

\begin{figure}[ht]
  \centering

  \includegraphics[width=0.49\linewidth]{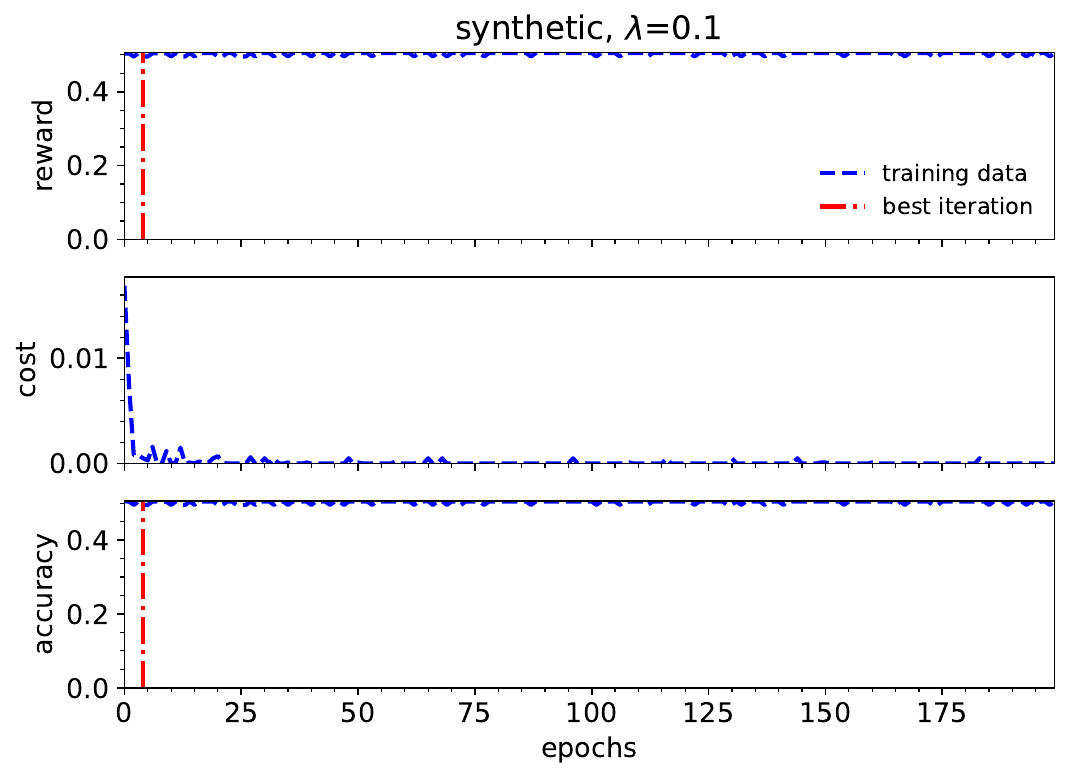}
  \includegraphics[width=0.49\linewidth]{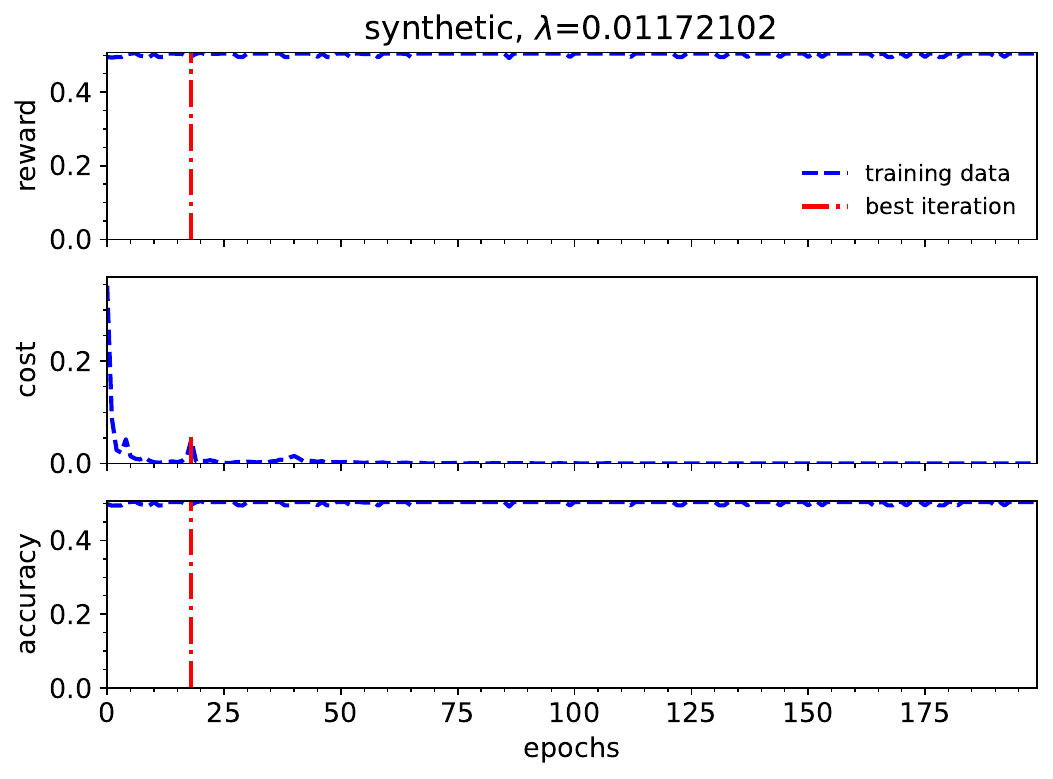}
  \includegraphics[width=0.49\linewidth]{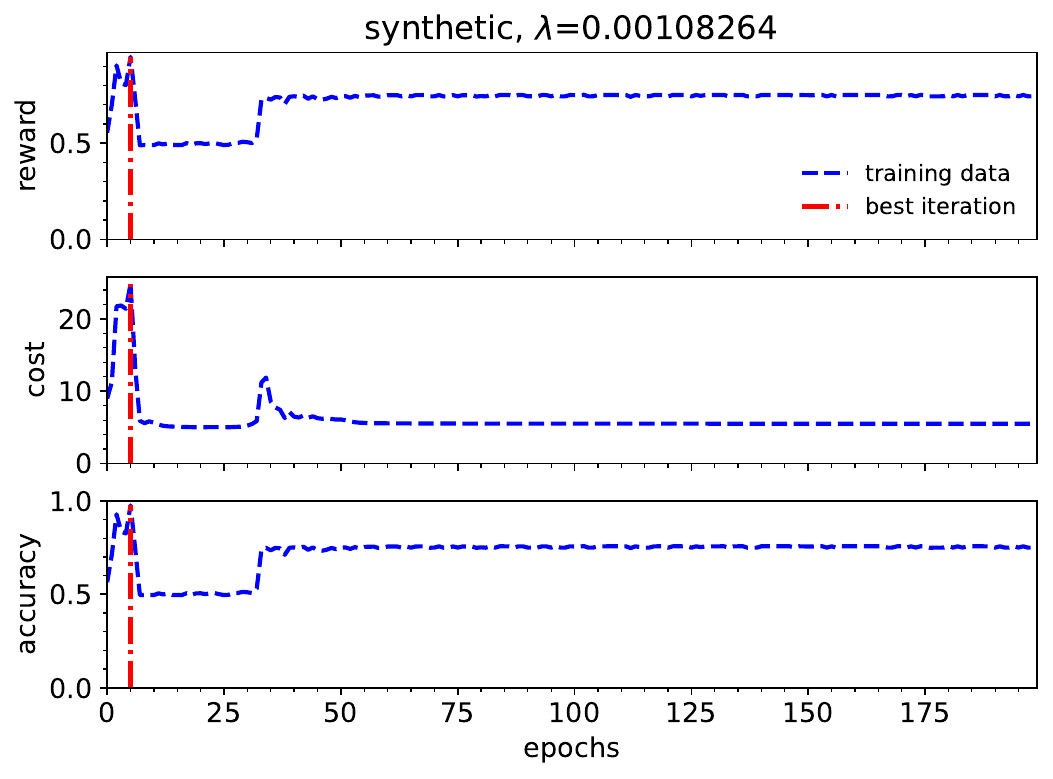}
  \includegraphics[width=0.49\linewidth]{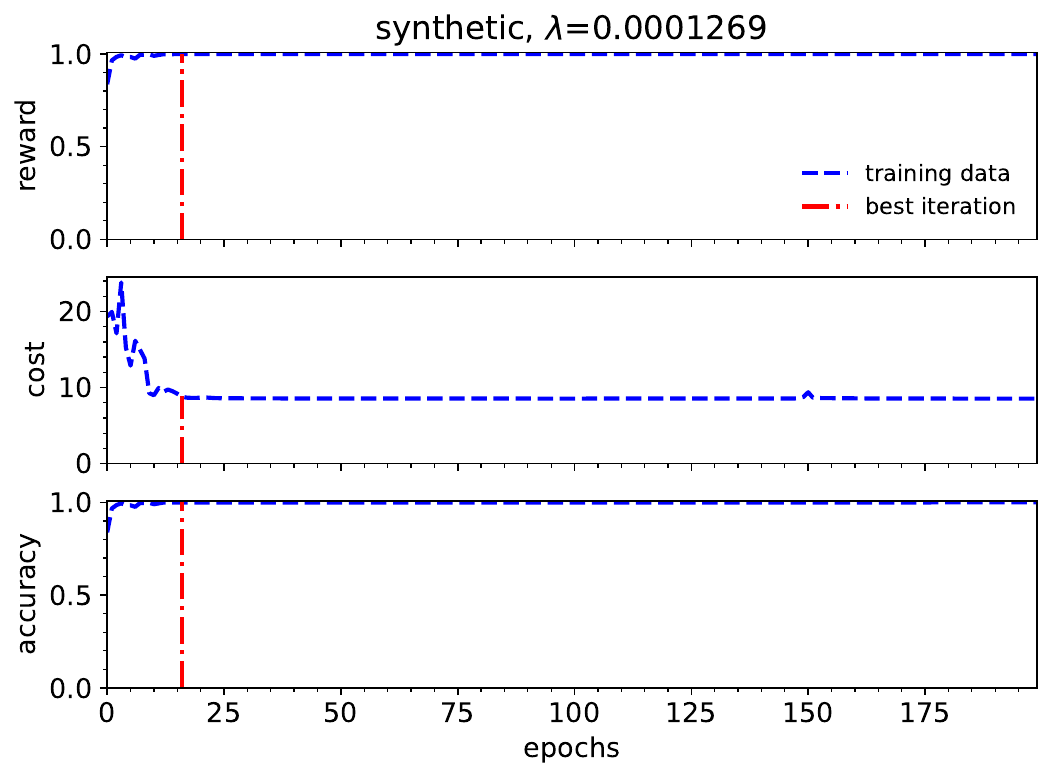}
  
  \caption{Convergence graphs for \emph{synthetic} dataset. Note that, for this dataset, the train, validation and test sets are the same, therefore we show only the train set.}  
\end{figure}

\begin{figure}[ht]
  \centering

  \includegraphics[width=0.49\linewidth]{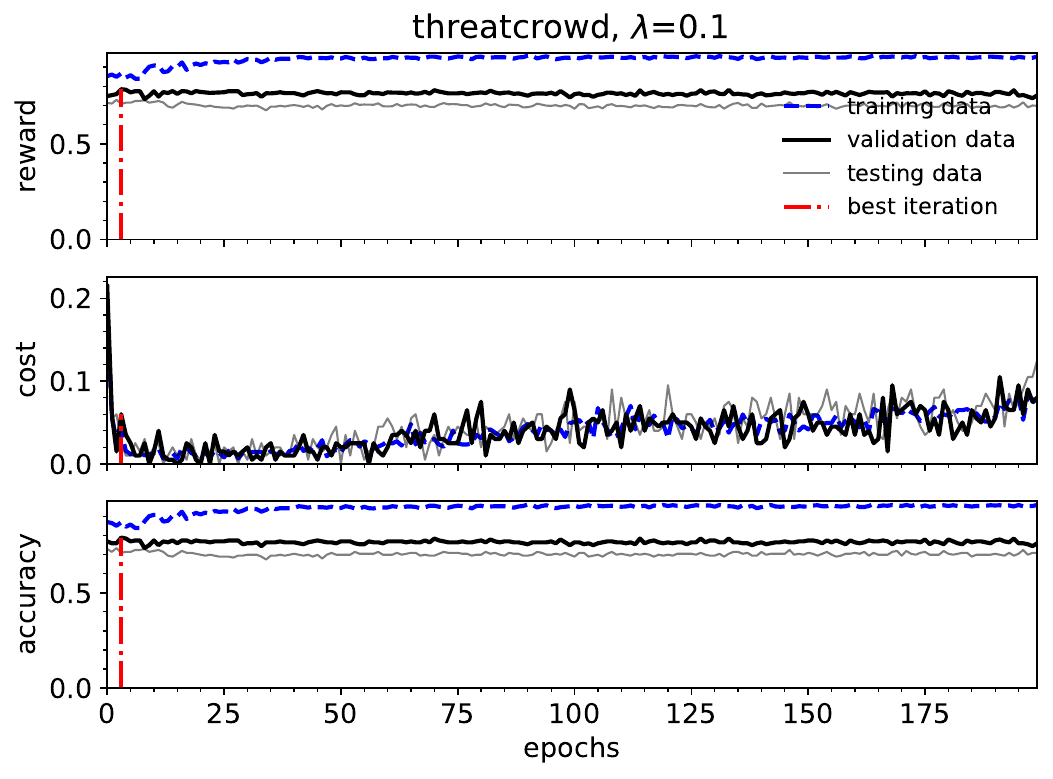}
  \includegraphics[width=0.49\linewidth]{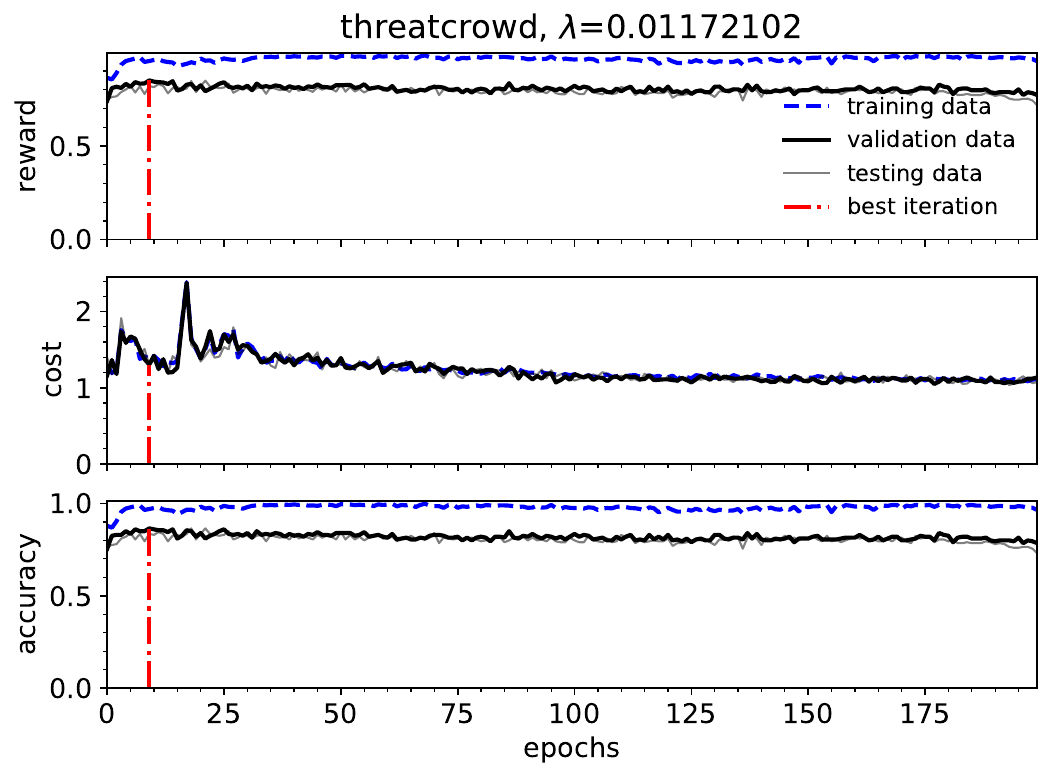}
  \includegraphics[width=0.49\linewidth]{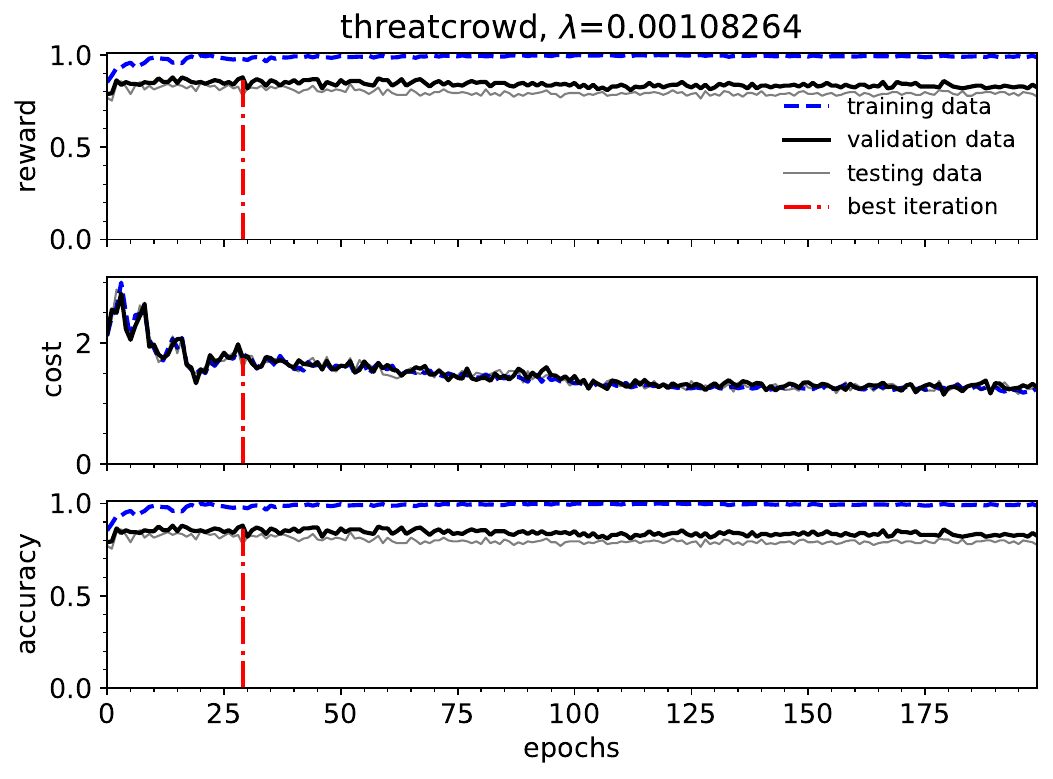}
  \includegraphics[width=0.49\linewidth]{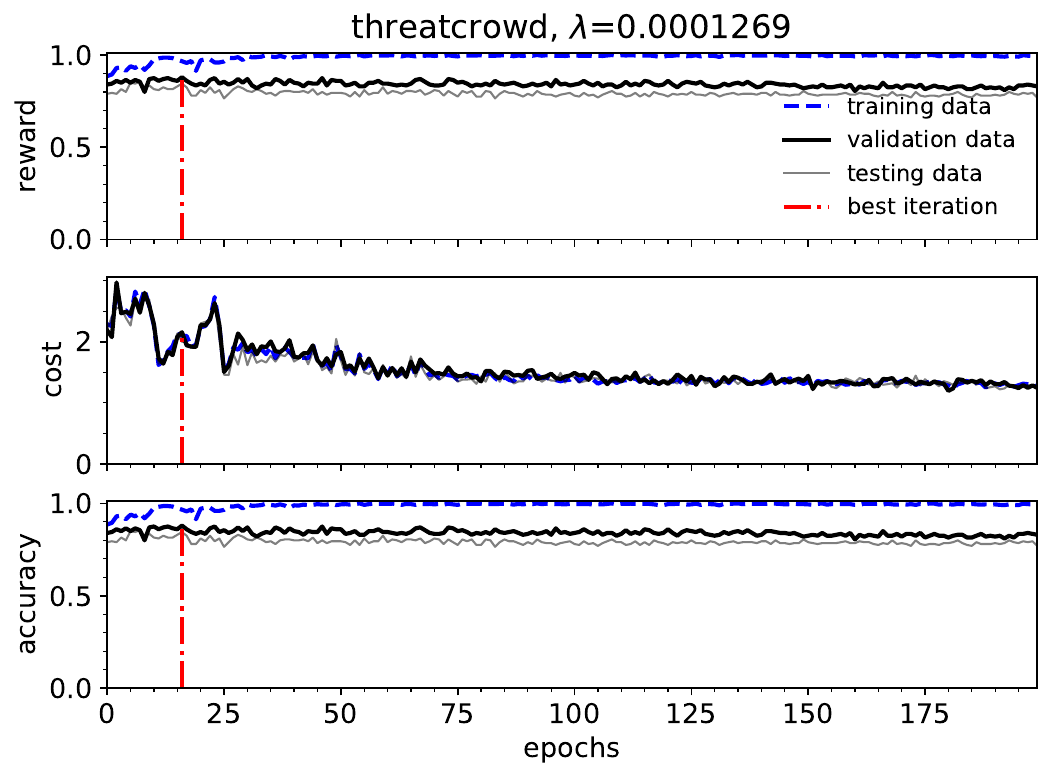}
  
  \caption{Convergence graphs for \emph{threatcrowd} dataset.}  
\end{figure}
\end{appendices}

\end{document}